\documentclass[acmsmall,screen]{acmart}

\AtBeginDocument{%
  }

\setcopyright{acmcopyright}
\copyrightyear{2024}
\acmYear{2024}
\makeatletter
\@printpermissionfalse
\@printcopyrightfalse
\@acmownedfalse
\@ACM@nonacmtrue
\makeatother

\usepackage{graphics}
\usepackage{multirow, makecell}
\usepackage{algorithmic}
\usepackage{algorithm}
\begin{document}

\title{Deepfake Detection: A Comprehensive Survey from the Reliability Perspective}

\author{Tianyi Wang}
\email{terry.ai.wang@gmail.com}
\orcid{0000-0003-2920-6099}
\affiliation{
  \institution{Department of Computer Science, The University of Hong Kong}
  \streetaddress{Pok Fu Lam}
  \city{Hong Kong}
  \country{China}
}

\author{Xin Liao}
\email{xinliao@hnu.edu.cn}
\orcid{0000-0002-9131-0578}
\affiliation{%
 \institution{College of Computer Science and Electronic Engineering, Hunan University}
 \streetaddress{Address}
 \city{Changsha}
 \state{Hunan}
 \country{China}
 \postcode{410082}
}

\author{Kam Pui Chow}
\email{chow@cs.hku.hk}
\orcid{0000-0003-4552-9744}
\affiliation{
  \institution{Department of Computer Science, The University of Hong Kong}
  \streetaddress{Pok Fu Lam}
  \city{Hong Kong}
  \country{China}
}

\author{Xiaodong Lin}
\email{xlin08@uoguelph.ca}
\orcid{0000-0001-8916-6645}
\affiliation{
  \institution{School of Computer Science, University of
Guelph}
  \streetaddress{50 Stone Road East}
  \city{Guelph}
  \state{Ontario}
  \country{Canada}
  \postcode{N1G 2W1}
}

\author{Yinglong Wang}
\email{wangyl@sdas.org}
\orcid{0000-0002-8350-7186}
\affiliation{%
  \institution{Key Laboratory of Computing Power Network and Information Security, Ministry of Education, Qilu University of Technology (Shandong Academy of Sciences)}
  \city{Jinan}
  \state{Shandong}
  \country{China}
  \postcode{250014}
  \thanks{Corresponding authors: Yinglong Wang and Xin Liao.}
}

\renewcommand{\shortauthors}{Wang et al.}

\begin{abstract}
  The mushroomed Deepfake synthetic materials circulated on the internet have raised a profound social impact on politicians, celebrities, and individuals worldwide. In this survey, we provide a thorough review of the existing Deepfake detection studies from the reliability perspective. We identify three reliability-oriented research challenges in the current Deepfake detection domain: transferability, interpretability, and robustness. Moreover, while solutions have been frequently addressed regarding the three challenges, the general reliability of a detection model has been barely considered, leading to the lack of reliable evidence in real-life usages and even for prosecutions on Deepfake-related cases in court. We, therefore, introduce a model reliability study metric using statistical random sampling knowledge and the publicly available benchmark datasets to review the reliability of the existing detection models on arbitrary Deepfake candidate suspects. Case studies are further executed to justify the real-life Deepfake cases including different groups of victims with the help of the reliably qualified detection models as reviewed in this survey. Reviews and experiments on the existing approaches provide informative discussions and future research directions for Deepfake detection.
\end{abstract}





\maketitle

\section{Introduction}
In June 2022, a mother from Pennsylvania, known as a `Deepfake mom', is sentenced to probation for three years because of her harassment of the rivals on her daughter’s cheerleader team~\cite{Katro2022BucksProb}. She is originally accused of using the so-called Deepfake technology to generate and spread fake videos of her daughter’s opponents depicting indelicate behaviors in March 2021. However, as admitted by the prosecutors, the justification for generating Deepfake videos is unable to be confirmed without accurate evidence and tools~\cite{Harwell2021CheerProve}. The term Deepfake refers to a deep learning technique raised by the Reddit user `deepfakes'~\cite{deepfakes} in 2017~\cite{Deepfakes2019Learn} that could automatically execute face-swapping from a source person to a target one while maintaining all other contents of the target image unchanged including expression, movement, and background scene. Later, the face reenactment technology~\cite{wayne2018reenactgan, ICface2019Tripathy, OneShotFace2019, one2022kang}, which transfers attributes of a source face to a target one while maintaining the target’s facial identity, is also classified as Deepfake from a comprehensive standpoint.

Deepfake can bring benefits and convenience to people's daily lives, especially from the perspective of human entertainment. Specifically, movie lovers may swap their faces onto movie clips and perform as their favorite superheroes and superheroines~\cite{Kietzmann2020TrickOrTreat}. On the other hand, a tainted celebrity who is no longer allowed to appear on TV shows~\cite{zhengshuang_scandal, zhaowei_scandal, zhangzhehan_scandal} can be face-swapped within completed TV productions instead of reshooting or removing the episode. Moreover, Deepfake is available for bringing a deceased person digitally back to life~\cite{Rima2019Museum} and greeting families and friends with desired conversations. Besides, e-commerce is another scenario to exert positive effects of Deepfake by trying on clothes in an online fitting room~\cite{Westerlund2019Emergence}. 

Benefiting from the publicly available source code implementations on the internet, various Deepfake mobile applications~\cite{wombo, faceswaplive, faceapp} have been released. The early product in 2018, FakeApp~\cite{Guilloux2018FakeApp}, requires a large number of input images to achieve satisfactory synthetic results. Later in 2019, the popular Chinese application, ZAO~\cite{ZAO2019}, can generate face-swapping outputs by simply inputting a series of selfies. Recently, a more powerful facial synthetic tool, Reface~\cite{Shvets2022Reface}, is built with further functionalities such as moving and singing animations based on hyper-realistic synthetic results. Although entertaining human lives, the free-access Deepfake applications require practically no experience in the field when generating fake faces, which therefore poses crucial potential threats to society. Because of the hyper-realistic quality that is indistinguishable by human eyes~\cite{sophie2022synthetic, sophie2022synthesized, hany2022ai-synthesized}, Deepfake has already been ranked as the most serious artificial intelligence crime threat since 2020~\cite{ScienceDaily2020}, and the current and potential victims include politicians, celebrities, and even every human being on earth. Besides the experimental fake Obama\footnote[1]{https://www.youtube.com/watch?v=cQ54GDm1eL0}, a fake president Zelensky~\cite{Miller2022Zelensky} has caused panic in Ukraine during the Russia-Ukraine war. As for celebrities, fake porn videos have frequently targeted female actresses and representative victims include Emma Watson, Natalie Portman, and Ariana Grande~\cite{Kelion2018Emma, Lee2018Portman}. Additionally, remember the `Deepfake mom' as discussed at the beginning? It is one of the best proofs that anyone can become a victim of Deepfake in modern society. 

To protect individuals and society from the negative impacts of misusing Deepfake, Deepfake detection approaches have been frequently designed and they mostly conduct a binary classification task to identify real and fake faces with the help of deep neural networks (DNNs). In this survey, we provide an in-depth review of the developing Deepfake detection approaches from the reliability perspective. In other words, we focus on studies and topics that devote to the ultimate success of Deepfake detection in real-life usages, and more importantly, for criminal investigations in court-case judgments. Early studies mainly focus on the in-dataset model performance such that the detection models are trained and validated the performance on the same dataset. While most recent work has achieved promising detection performance for the in-dataset test, the research challenges at the current stage for Deepfake detection can be concluded in three aspects, namely, \textit{transferability}, \textit{interpretability}, and \textit{robustness}. The transferability topic refers to the progress of improving the cross-dataset ability of models when evaluated on unseen data. As the detection performance keeps advancing by various approaches, interpretability is another research goal to explain the reason that the detection model determines the falsification. Moreover, when applying well-trained and well-performed detection models for real-life scenarios, robustness is considered a main topic in dealing with various real-life conditions. 

While research has been incrementally attempted regarding the three challenging topics, a reliable Deepfake detection model is expected to benefit people's daily lives with good transferability on unseen data, have convincing interpretability of the detection decision, and show robust performance against practical application scenarios and conditions in real life. However, there lacks further discussion on model reliability in existing research papers and surveys. In particular, without an authenticated scheme to nominate the detection models as reliable evidence to assist prosecutions and judgments in court, similar failures as the `Deepfake mom’ case will happen again due to the lack of reliable detection tools to support the accusation of Deepfake even though detection performance on each benchmark dataset is reported in current work. In other words, the trustworthiness of a model-derived falsification needs to be proved before it can convince people in real-life usages and for court-case judgments. To fulfill the research gap of the model reliability study, beyond the comprehensive review of Deepfake detection, we devise a scheme to scientifically validate the reliability of the well-developed Deepfake detection models using statistical random sampling knowledge~\cite{randsample2018Martino}. To guarantee the credibility of the reliability study, we concurrently introduce a systematic workflow of data pre-processing including image frame selection and extraction from videos and face detection and cropping, which has been barely mentioned with concrete details in past work. Thereafter, we quantitatively evaluate and record the selected state-of-the-art Deepfake detection approaches by training and testing with their reported optimal settings on the same group of pre-processed datasets in a completely fair game. Thenceforth, we validate the detection model reliability following the designed evaluation scheme. In the end, a case study is enforced to justify the results derived by the detection models on four well-known real-life synthetic videos concerning the reliable detection accuracies statistically at 90\% and 95\% confidence levels based on the research outcomes from the model reliability study. Interesting findings and future research topics that have been scarcely concluded in previous studies~\cite{TOLOSANA2020Deepfakes, Kietzmann2020TrickOrTreat, Westerlund2019Emergence, Tolosana2021DeepFakes, Creation2021Mirsky, Farid2022Creating} are analyzed and discussed. Furthermore, we believe that the proposed Deepfake detection model reliability study scheme is informative and can be adopted as evidence to assist prosecutions in court once granted approval by authentication experts or institutions following the legislation.

The rest of the paper is organized as follows. We provide a brief review of the popular synthetic techniques and publicly available benchmark datasets in Section~\ref{sec:generation_evolution}. Then, we define the challenges of the Deepfake detection research and provide a thorough review of the development history of the Deepfake detection approaches in Section~\ref{deepfake_detection}. In Section~\ref{reliability_study}, we illustrate the model reliability study scheme and demonstrate the algorithm details. In Section~\ref{deepfake_preparation}, we detailedly introduce a standardized data pre-processing workflow and list the participating datasets in the experiments of this paper. In Section~\ref{model_evaluation}, we conduct detection performance evaluation and reliability justification using selected state-of-the-art models on the benchmark datasets and following the reliability study scheme, respectively. Section~\ref{case_study} exhibits Deepfake detection results of the selected models when applying to the real-life videos in a case study, and discussions along with experiment results from early sections are presented in Section~\ref{discussion}. Section~\ref{conclusion} concludes the remarks and highlights the potential future directions in the research domain.

\section{The Evolution of Deepfake Generation}
\label{sec:generation_evolution}

\subsection{Deepfake Generation}
\label{deepfake_generation}

Deepfake is initially raised in the Reddit community when the open-source implementation was first published by the user `deepfakes' simultaneously in 2017. Early research mainly focuses on subject-specific approaches, which can only swap facial identities that the models have seen during training. The most popular framework of the existing Deepfake synthesis studies~\cite{deepfakes, perov2021deepfacelab} for the subject-specific identity swap is an autoencoder~\cite{Kingma2014Autoencoder}. In a nutshell, the autoencoder contains a shared encoder that extracts identity-independent features from the source and target faces and two unique decoders each is in charge of reconstructing synthetic faces of a corresponding facial identity. Specifically, in the training phase, faces of the source identity are fed to the encoder for identity-independent feature extraction. The extracted context vectors are then passed through the decoder that corresponds to reconstructing the faces of the source identity. Similarly, the target face reconstruction is trained following the same workflow. When using a well-trained model to operate face-swapping, a target face after context vector extraction is fed to the decoder that reconstructs the source identity. Thenceforth, the decoder generates a look maintaining the facial expression and movement of the target face while having the identity of the desired source face. If the face-swapping model is trained for both directions, a target face may be face-swapped onto a source face following the same workflow.

Recent studies gradually focus on subject-agnostic methods to enable face-swapping for arbitrary identities with higher resolutions and have exploited Generative Adversarial Networks (GAN)~\cite{Goodfellow2014Generative} for better synthesis authenticity~\cite{cyclegan2017zhu, Ding2018ExprGANFE, Natsume2018RSGAN, karras2018progressive, brock2018large, stargan2018_choi, nirkin2019fsgan, gaugan2019park, gao2021info, zhang2021facial, stylegan2021karras}. In other words, they aim to consistently produce high-quality face-swapping results even on facial identities that are unseen during model training. GAN is a generator-discriminator architecture that is trained by having two components battle against each other to advance the output quality. In practice, the generator is periodically trained to fool the discriminator with synthetic faces. For instance, FaceShifter~\cite{Li2019FaceShifter} and SimSwap~\cite{Chen2020SimSwap} each devise particular modules to preserve facial attributes that are hard to reconstruct and maintain fidelity for arbitrary facial identities. MegaFS~\cite{zhu2021megafs} and HifiFace~\cite{Wang2021HifiFace} accomplish face-swapping at high resolutions of 512 and 1,024 for arbitrary facial identities, respectively, relying on the reconstruction ability of GAN. In the last step, the generated fake face is usually blended back to the pristine target image with tuning techniques such as blurring and smoothing~\cite{blur2018zhang} to reduce the visible Deepfake traces.

\subsection{Deepfake Benchmark Datasets}
\label{dataset_evolution}

Benchmark datasets are vital in the development history of Deepfake detection models. Dolhansky et al.~\cite{WildDeepfake2020} raised the idea to break down previous datasets into three generations based on the two-generation categorization in the early work~\cite{jiang2020deeperforensics1}. As listed in Table~\ref{tab1}, UADFV~\cite{ExposingDeep2019Li}, DeepfakeTIMIT~\cite{DeepfakeTIMIT2018Korshunov}, and FaceForensics++ (FF++)~\cite{Rossler2019FaceForensics} are categorized to the first generation; Deepfake Detection Dataset~\cite{DFD2019Nick}, DeepFake Detection Challenge (DFDC) Preview~\cite{DFDC_Preview}, and Celeb-DF~\cite{Li2020CelebDF} are in the second generation; DeeperForensics-1.0 (DF1.0)~\cite{jiang2020deeperforensics1} and DeepFake Detection Challenge (DFDC)~\cite{DFDC} are in the third generation. In summary, later generations contain general improvements over the previous ones in terms of dataset magnitude or synthetic method diversity. 

Unlike the summary by Dolhansky et al.~\cite{WildDeepfake2020}, the agreement from individuals appearing is not considered in this study, and we instead re-define the third generation such that the datasets are of better quality, broader diversity and magnitude, higher difficulty than the early generations, or challenging detailed discrepancies in early synthetic videos are resolved in the datasets. Specifically, besides DFDC with large manipulation diversity and dataset magnitude and DF1.0 with large magnitude and considerable difficulty by adding deliberate perturbations, we further classify the following datasets in the third generation, namely, FaceShifter~\cite{Li2019FaceShifter}, WildDeepfake~\cite{WildDeepfake2020}, and KoDF~\cite{KoDF2021Kwon}.

FaceShifter~\cite{Li2019FaceShifter}, although synthesized based on real videos of FF++, has specifically solved the so-called facial occlusion challenge that appears in previous datasets. In other words, the synthetic results are better handled even in difficult cases where parts of the face are blocked or obscured by objects such as accessories or other body parts such as hair tippings. WildDeepfake (WDF)~\cite{WildDeepfake2020} appears to be a special one in the third generation because videos are totally collected from the internet, which matches the real-life Deepfake circumstance the best. The most recent KoDF~\cite{KoDF2021Kwon} dataset is so far the largest Deepfake benchmark dataset that is publicly available with reasonable diversity and contains synthetic videos at high resolutions.

\begin{table}
\centering
\caption{Information of the existing Deepfake datasets categorized into three generations based on quality, diversity, and difficulty. Publication year, the number of real and fake sequences, and the source of real and fake materials are listed.}
\resizebox{\textwidth}{!}{
\begin{tabular}{lcccc}
\toprule\noalign{\smallskip}
Dataset & Year & \# Real / Fake  & Real / Fake Source & Generation \\
\noalign{\smallskip}\midrule\noalign{\smallskip}
UADFV~\cite{ExposingDeep2019Li} & 2018 & 49 / 49 & YouTube / FakeApp~\cite{Guilloux2018FakeApp} & \multirow{3}{*}{1$^{\textrm{st}}$ Generation} \\
DeepfakeTIMIT~\cite{DeepfakeTIMIT2018Korshunov} & 2018 & -- / 620 & faceswap-GAN~\cite{faceswap-GAN} \\
FF++~\cite{Rossler2019FaceForensics} & 2019 & 1,000 / 4,000 & YouTube / 4 methods\footnotemark[2] \\
\noalign{\smallskip}\midrule\noalign{\smallskip}
DFD~\cite{DFD2019Nick} & 2019 & 363 / 3,068 & consenting actors / unknown methods & \multirow{3}{*}{2$^{\textrm{nd}}$ Generation} \\
DFDC Preview~\cite{DFDC_Preview} & 2019 & 1,131 / 4,119 & crowdsourcing / 2 unknown methods \\
Celeb-DF~\cite{Li2020CelebDF} & 2019 & 590 / 5,639 & YouTube / improved Deepfake \\
\noalign{\smallskip}\midrule\noalign{\smallskip}
DF1.0~\cite{jiang2020deeperforensics1} & 2020 & -- / 10,000 & FF++ real / DF-VAE & \multirow{5}{*}{3$^{\textrm{rd}}$ Generation} \\
FaceShifter~\cite{Li2019FaceShifter} & 2020 & -- / 1,000 & FF++ real / GAN-based \\
DFDC~\cite{DFDC} & 2020 & 23,654 / 104,500 & crowdsourcing / 8 methods\footnotemark[3] \\
WDF~\cite{WildDeepfake2020} & 2020 & 3,805 / 3,509 & video-sharing websites \\
KoDF~\cite{KoDF2021Kwon} & 2021 & 62,166 / 175,776 & lab-controlled / 6 manipulations\footnotemark[4] \\
\noalign{\smallskip}\bottomrule
\end{tabular}
}
\label{tab1}
\end{table}

\section{Reliability-Oriented Challenges of Deepfake Detection}
\label{deepfake_detection}
Detection work on Deepfake has been proposed since the first occurrence of Deepfake contents. Classical forgery detection approaches~\cite{Ferrara2012ImageFL, Fridrich2012Rich, Pan2012Exposing, Cozzolino2014Image, Peng2017Optimized3L, Deng2019ArcFace} mainly focus on the intrinsic statistics and hand-crafted traces such as eye blinking~\cite{InIctuOculi2018Li, DeepVision2020Jung}, head pose~\cite{ExposingDeep2019Li}, and visual artifacts~\cite{VisualArtif2019Matern} to analyze the spatial feature manipulation patterns. Besides, there are papers that have derived high accuracies and AUC scores by training and testing on the same dataset of a synthetic method. Several studies~\cite{deepfake2018david, Sabir2019RecurrentCS} integrated CNN and Long Short-term Memory (LSTM)~\cite{lstm} for spatial and temporal features analyses, respectively, and accomplished detection for in-dataset evaluations on self-collected data and the FF++ dataset. Hsu, Zhang, and Lee~\cite{hsu2020deep} utilized GAN-generated fake samples for real-fake pairwise training using DenseNet~\cite{dense_net}. Agarwal et al.~\cite{Agarwal2020Detecting_WIFS} accomplished detection on face-swap Deepfake using CNNs with biometric information including facial expressions and head movements. However, although accomplished well-pleasing in-dataset detection performance on some early or self-collected datasets, they are mostly easily fooled by the hyper-realistic Deepfake contents in the current research domain because of limitations in dataset quality, dataset diversity, and method or model ability. 

\begin{figure}
\centering
\includegraphics[width=\textwidth]{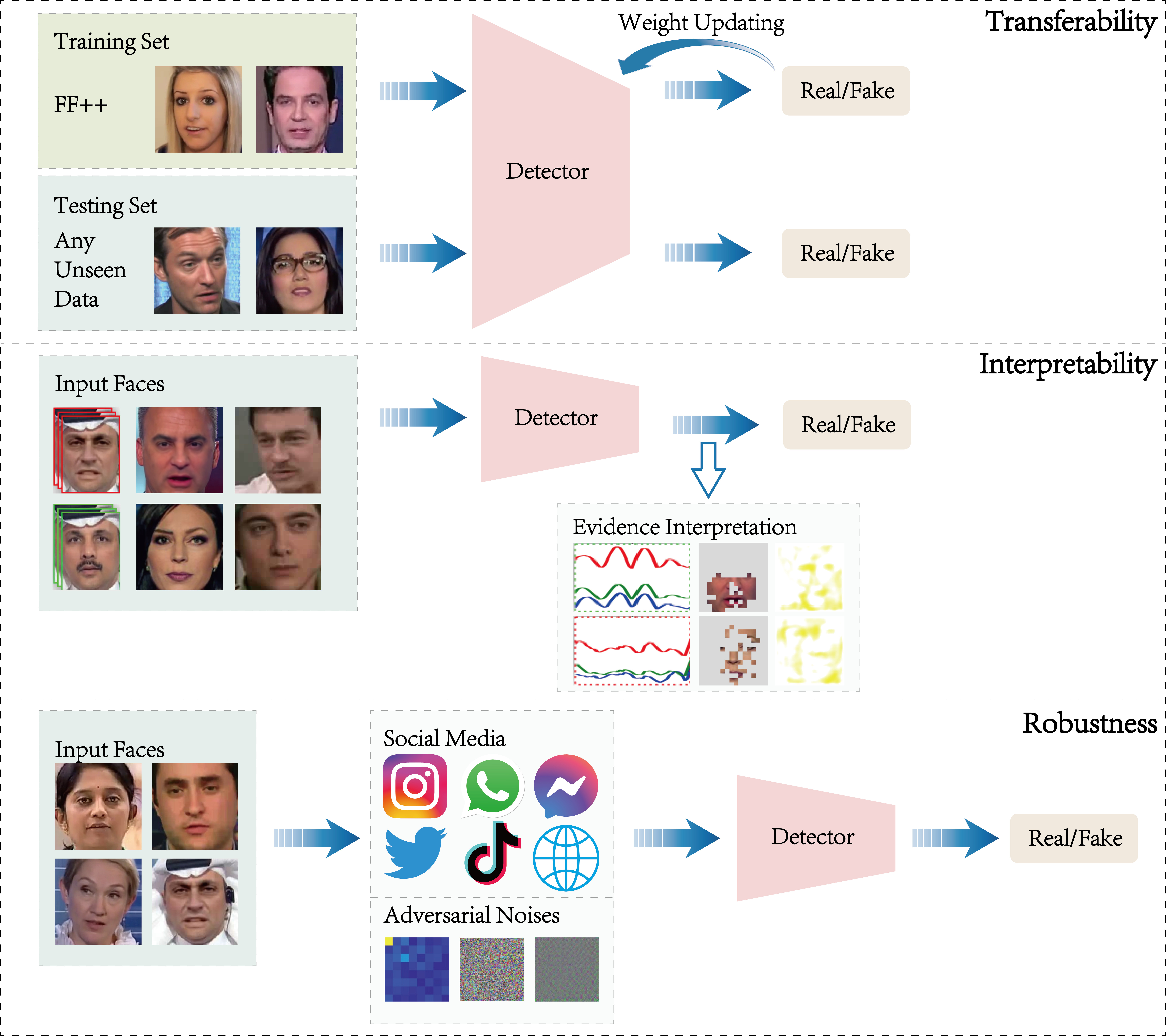}
\caption{Demonstrations of the three challenges from top to bottom. Transferability (top) refers to models that focus on stable detection ability on unseen benchmark datasets; interpretability (middle) refers to efforts on explaining the model detected falsification; robustness (bottom) refers to models that handle Deepfake suspects under different real-life conditions and scenarios. }
\label{fig_challenges}
\end{figure}

Later studies gradually consider Deepfake detection as a binary classification task using DNNs. As malicious Deepfake contents have started to jeopardize human society and the cases are even discussed in court, reliably trusted detection methods are eagerly desired by the public. In particular, three challenges (Fig.~\ref{fig_challenges}) of the current Deepfake detection research domain can be summarized regarding the reliability goal, namely, transferability, interpretability, and robustness. 

\footnotetext[2]{FaceSwap, Deepfakes, Face2Face, and NeuralTextures. }
\footnotetext[3]{DF-128, DF-256, MM/NN, NTH, FSGAN, StyleGan, refinement, and audio swaps.}
\footnotetext[4]{FaceSwap, DeepFaceLab, FSGAN, FOMM, Audio-driven.}

\subsection{Transferability}
\label{transferability}
Deep learning models usually exhibit satisfactory performance on the same type of data that are seen in the training process but perform poorly on unseen data. In real life, Deepfake materials can be generated via various synthetic techniques~\cite{faceswap,faceswap-GAN,Chen2020SimSwap,gao2021info,zhu2021megafs} as abundant on-the-shelf easily accessible face-swapping implementations are publicly available, and a reliable Deepfake detection model is expected to perform well on unseen data to imitate real-life Deepfake cases. Therefore, guaranteeing the transferability of the detection models for cross-dataset performance is necessary and frequently discussed. 

With the fast development of deep learning techniques, various methods are devised using simple Convolutional Neural Network (CNN) based models. Zhou et al.~\cite{zhou2017twostream} fused a CNN stream with a support vector machine (SVM)~\cite{SVM1998_Hearst} stream to analyze face features with the assistance of local noise residuals. Afchar et al.~\cite{Afchar2018MesoNet} studied the mesoscopic features of images with successive convolutional and pooling layers. Nguyen et al.~\cite{Nguyen2019MultitaskLF} designed a multi-task learning scheme to simultaneously perform detection and localization using CNNs. DFT-MF~\cite{Jafar2020forensics} thinks that the mouth features can be important for detection and utilizes a convolutional model to detect Deepfake by verifying, analyzing, and isolating mouth and lip movements. Face X-ray~\cite{Li2020face} performs a medical x-ray on the candidate Deepfake faces by revealing whether the blending of two different source images can be decomposed.

Rather than the basic CNN architectures, well-designed and pre-trained CNN backbones are frequently exploited to improve model performance on Deepfake detection, especially for the cross-dataset performance on unseen data as more benchmark datasets have been released. An early approach~\cite{Amerini2019Deepfake} proposes optical flow analysis with pre-trained VGG16~\cite{Simonyan2015very} and ResNet50~\cite{He2016Deep} CNN backbones and achieves preliminary in-dataset test performance on the FaceForensics++ (FF++) dataset~\cite{Rossler2019FaceForensics}. Capsule~\cite{nguyen2019use} employs capsule architectures~\cite{Sabour2017dynamic} with light VGG19-based network parameters but achieves similar detection performance to the traditional approaches leveraging CNNs. Li et al.~\cite{li2019exposing} conducted a strengthened model, DSP-FWA, with the help of the spatial pyramid pooling~\cite{He2015spatial} with ResNet50 as the backbone. This method is shown to be applicable to Deepfake materials at different resolution levels. FFD~\cite{Dang2020on} leverages the popular attention mechanism by element-wise multiplication to study the feature maps and utilizes the XceptionNet CNN backbone, achieving marginally more promising performance than the work by Rossler et al.~\cite{Rossler2019FaceForensics}. SSTNet~\cite{Wu2020SSTNet} exploits Spatial, Steganalysis, and Temporal features using XceptionNet~\cite{Chollet2017Xception} and exhibits reasonable intra-cross-dataset performance on FF++. Given the assumption that Deepfake only modifies the internal part of the face, Nirkin et al.~\cite{Nirkin2022DeepFake} adopted two streams of XceptionNet~\cite{Chollet2017Xception} for face and context (hair, ears, neck) identification and another XceptionNet to classify real and fake based on the learned discrepancies between the two. Later, Bonettini et al.~\cite{Bonettini2021ensemble} and Tariq et al.~\cite{tariq2018detecting} studied the ensemble of various pre-trained convolutional models including EfficientNetB4~\cite{Tan2019EfficientNet}, XceptionNet~\cite{Chollet2017Xception}, DenseNet~\cite{dense_net}, VGGNet~\cite{Simonyan2015very}, ResNet~\cite{He2016Deep}, and NASNet~\cite{nas_net} backbones to detect Deepfake. Rossler et al.~\cite{Rossler2019FaceForensics} employed the pre-trained well-designed XceptionNet~\cite{Chollet2017Xception} network and achieved state-of-the-art detection performance at the time on FF++. Besides, a special study introduced by Wang et al.~\cite{wang2020cnngen} proves the transferability of the model trained on one CNN-generated dataset to the rest ten using pre-trained ResNet50~\cite{He2016Deep}.

Meanwhile, frequency cues are also noticed and analyzed by researchers. While early image forgery detection work~\cite{bai2020fake} focus on all high, medium, and low frequencies via Fourier transform, Frank et al.~\cite{LeveFreq2020Frank} were the first that raised the idea of finding frequency inconsistency between real and fake by employing low-frequency features to assist Deepfake detection using RGB information. Since low-frequency traces are mostly hidden by blurry facial features, later studies mainly analyze high-level features. F3-Net~\cite{Qian2020Thinking} exploits both low- and high-frequency features without RGB feature extraction operation. DFT~\cite{durall2020unmasking}, Two-Branch~\cite{Masi2020Two-Branch}, SPSL~\cite{SpatialPhase2021Liu}, and MPSM-RFAM~\cite{LocalRel2021Chen} accomplish promising detection results by analyzing high-frequency spectrum along with the on-the-shelf RGB features. Li et al.~\cite{FreqAware2021Li} extracted both middle- and high-frequency features and correlated frequency and RGB features for detection. By pointing out the drawbacks of using coarse-grained frequency information, Gu et al.~\cite{Gu2022Exploiting} combined fine-grained frequency information with RGB features for better feature richness in a latest work. Moreover, Jeong et al.~\cite{Jeon2022FrePGAN} designed a new training scheme with frequency-level perturbation maps added, which further enhanced the generalization ability of the detection model regarding all GAN-based generators.

Since the CNN architecture and backbones lack generalization ability and mainly focus on local features, even XceptionNet is restricted in learning the global features for further performance improvements. Therefore, developed solutions have introduced convolutional spatial attention to enlarge the local feature area and learn the corresponding relations, and the detection AUC scores have been gradually raised above 70\% on average upon unseen datasets accordingly. The SRM~\cite{Luo2021Generalizing} approach makes two streams of network using XceptionNet as the CNN backbone and focuses on the high-frequency information and RGB frames with a spatial attention module in each stream. The MAT model~\cite{Zhao2021Multi-Attentional} proposes the CNN backbone EfficientNetB4 and borrows the convolutional attention idea to study different local patches of the input image frame. Specifically, the artifacts in shallow features are zoomed in for fine-grained enhancement in the detection performance. With the success of the transformer architecture~\cite{Vaswani2017Attention} in the natural language processing (NLP) domain, different versions of vision transformer~\cite{Alexey2021An, liu2021Swin, Peng2021Conformer, Hugo2021Training, Wang2021Pyramid} have derived reasonable performance in the computer vision domain due to its ability on global image feature learning. The architectures of vision transformers (ViT) have also been employed for Deepfake detection with promising results in recent work~\cite{Jeon2020FDFtNet, Heo2022DeepFake, Deressa2021Deepfake, DCPT2022Wang, VFD2023Cheng}.

While reaching the bottleneck specially for the cross-dataset performance even resorting to the advanced powerful but potentially time-consuming neural networks, most recent work gradually focuses on strategies to enrich diversity in training data. Such attempts have improved the detection AUC scores even up to 80\% on some unseen datasets. PCL~\cite{SelfConsis2021Zhao} is introduced with an inconsistency image generator to add synthetic diversity and provide richly annotated training data. Sun et al.~\cite{Sun2022Dual} proposed Dual Contrastive Learning (DCL) to study positive and negative paired data and enrich data views for Deepfake detection with better transferability. FInfer~\cite{Hu2022FInfer} inferences future image frames within a video and is trained based on the representation-prediction loss. Shiohara and Yamasaki~\cite{SBIs2022Shiohara} presented a novel synthetic training data called self-blended images (SBIs) by blending solely pristine images from the original training dataset, and classifiers are thus boosted to learn more generic representations. Chen et al.~\cite{SLADD2022Chen} proposed the SLADD model to further synthesize the original forgery data for model training to enrich model transferability on unseen data using pairs of pristine images and randomly selected forgery references with forgery configurations including forgery region, blending type, and mix-up blending ratio. Cao et al.~\cite{RECCE2022Cao} introduced the RECCE model to emphasize the common compact representations of genuine faces based on reconstruction-classification learning on real faces. The reconstruction learning over real images enhances the learning representations to be aware of unknown forgery patterns. Liang et al.~\cite{Exploring2022Liang_ECCV} proposed an easily embeddable disentanglement framework to remove content information while maintaining artifact information for training and Deepfake detection using reconstructed data with various combinations of content and artifact features of real and fake samples. The OST~\cite{chen2022ost_nips} method improves the detection performance by preparing pseudo-training samples based on testing images to update the model weights before applying to the true samples.

\subsection{Interpretability}
\label{interpretability}

Despite the promising ability of deep learning models, they suffer the weak interpretability problems due to the black-box characteristic~\cite{Black-Box2019Loyola}. In other words, it is hard to explain how and why a model comes up with a particular result. For the Deepfake detection task, while achieving promising detection performance statistically for in- and cross-dataset evaluations, the interpretability issue is still maintained to be fully resolved at the current stage. In other words, people tend to trust methods that are easily understandable via common sense rather than those with satisfactory accuracies but derived based on features that are hard to explain. Consequently, forensic evidence is critical to be probed to interpret and support the detection model performance by highlighting the reasons for classifying fake samples. In a nutshell, the interpretability challenge is to answer the following questions regarding a Deepfake suspect in order to be reliable:

\begin{itemize}
\item Why is the content classified as fake? 
\item Based on which part does the detection model determine the content as fake?
\end{itemize}

As confirmed by Baldassarre et al.~\cite{Quantitative2022Baldassarre_BMVC} using a series of quantitative metrics for evaluating the interpretability of the detection models, heatmaps are generally futile and unacceptable when being adopted to explain the detected artifacts~\cite{Supervised2022Xu_WACV}. Moreover, heatmaps of the real faces are displayed in like manners, which are indistinguishable from that of the fake ones. Therefore, to answer the above questions, a group of noise-based research has been attempted to probe the Deepfake forensic traces for explanations upon the detection results. Early work mostly relies on the Photo Response Non-Uniformity (PRNU), a small factory-defects-generated noise pattern in the light-sensitive sensors of a digital camera~\cite{Lukas2006Digital}. PRNU has shown strong abilities in source anonymization~\cite{picetti2020dippas} and source device identification~\cite{Saito2017ATheoretical, Marra2017blind}. Unfortunately, most of the PRNU-based Deepfake detection studies~\cite{Koopman2018detection, Weever2020DeepfakeDT} have failed to show strong detection performance statistically. Therefore, the PRNU noise pattern can be a useful instrument for source device identification tasks, but it may not be a meaningful forensic noise tracing tool to satisfy the purpose of the Deepfake detection task with respect to the interpretability goal.
%~\cite{Improving2022Yu_TIFS}

Later approaches~\cite{Amerini2019Deepfake, Jafar2020forensics, Li2020face, Chen2022SNIS} step up to utilize CNNs for noise extraction and analyze the trace differences between real and fake. A milestone denoiser, DnCNN, proposed by Zhang et al.~\cite{zhang2017beyond} is able to perform blind Gaussian denoising with promising performance, and has been later extended to study the camera model fingerprint for various downstream tasks including forgery detection and localization for face-swapping images based on the underlying noises~\cite{Cozzolino2020noiseprint, Verdoliva2019Extracting}. Recently, studies~\cite{Guarnera2020Fighting, Guarnera2020DeepFake} aim to extract the manipulation traces for detection and interpretation. Guo et al.~\cite{GUO2021fake} proposed the AMTEN method to suppress image content and highlight manipulation traces as more filter iterations are applied. Wang et al.~\cite{DeepfakeNoise2022Wang, Wang2023NoiseDF_AAAI} utilized the pre-trained denoisers for Deepfake forensic noise extraction and investigated the underlying Deepfake noise pattern consistency between face and background squares with the help of the siamese architecture~\cite{Bromley1993}. Guo et al.~\cite{Guo2022Exposing} proposed a guided residual network to maintain the manipulation traces within the guided residual image and analyzed the difference between real and fake. 

Besides studies~\cite{Qi2020DeepRhythm, Ciftci2020FakeCatcher} that rely on biological signals by analyzing minuscule periodic changes through the faces visualizing distinguishable sequential signal patterns as indicators, other studies mostly attempt to explore universal and representative artifacts directly from the visual concepts of the Deepfake materials. The PRRNet~\cite{PRRNet2021SHANG} approach studies region-wise relation and pixel-wise relation within the candidate image for detection and can roughly locate the manipulated region using pixel-wise values. Trinh et al.~\cite{Interpretable2021Trinh_WACV} fed the dynamic prototype to the detection model and successfully visualized obvious artifact fluctuation via the prototype. Yang et al.~\cite{yang2021beyond_ijcai} proposed to re-synthesis testing images by incorporating a series of visual tasks using GAN models, and they finally extracted visual cues to help perform detection. Obvious artifact differences can be observed by their Stage5 model on real and fake samples. Most recently, Dong et al.~\cite{Explaining2022Dong_ECCV} introduced FST-matching to disentangle source-, target-, and artifact-relevant features from the input image and improved the detection performance by utilizing artifact-relevant features solely. The adopted features within the fake samples are visualized to explain the detection results. 

\subsection{Robustness}
\label{robustness}

As the transferability and interpretability challenges have been frequently undertaken and reasonable or even promising results have been achieved accordingly, there lacks a stage to be fulfilled in order for the detection approaches to be useful in real-life cases. The challenge of this stage can be summarized as robustness. To be specific, the quality of the candidate Deepfake material in real life is not as ideal as the benchmark datasets in experimental conditions most of the time. Consequently, they may experience different levels of compression operation in multiple scenarios due to objectively limited conditions~\cite{Marcon2021Detection}. Moreover, post-processing strategies and artificially added perturbations can cause further challenges and even disable the well-trained and well-performed detection models. Therefore, the robustness of the detection approaches is necessary to be significantly considered when facing various real-life application scenarios under restricted and special conditions.

Multiple studies have been conducted mainly regarding two types of real-life conditions, namely, the passive condition and active condition. The passive condition refers to scenarios with objective limitations such as video compression due to network flow settings. In an early study, Kumar, Vatsa, and Singh~\cite{kumar2020detecting} designed multiple streams of ResNet18~\cite{He2016Deep} to specifically deal with face reenactment manipulation at various compression levels. Hu et al.~\cite{Hu2021Detecting} proposed a two-stream method that specifically analyzes the features of compressed videos that are widely spread on social networks. The LRNet~\cite{sun2021improving} model is developed to stay robust when detecting highly compressed or noise-corrupted videos. Cao et al.~\cite{cao2021metric} solved the difficulty of detecting against compressed videos by feeding compression-insensitive embedding feature spaces that utilize raw and compressed forgeries to the detection model. Wu et al.~\cite{Robust2022Wu_TIFS, Robust2022Wu_CVPR} analyzed noise patterns generated via online social networks before feeding image data into the detection model for training. The method has won top ranking against existing approaches especially when facing forgeries after being transmitted through social networks. Le and Woo~\cite{Binh2022ADD} employed attention distillation in the frequency perspective and have successfully raised the detection ability on highly compressed data using ResNet50. RealForensics~\cite{Leveraging2022Haliassos_CVPR} aims to solve Deepfake contents with real-life quality by training the detection model with an auxiliary dataset containing real talking faces before utilizing the benchmark datasets. The method has significantly advanced the detection performance against multiple objective scenarios and adversarial attacks such as Gaussian noise and video compression. Besides, a recent work~\cite{ForgeryNIR2022Wang_TIFS} has constructed a new dataset containing Deepfake contents under the near-infrared condition to prevent potential future Deepfake attacks in the corresponding scenarios.

On the other hand, the active condition summarizes deliberate adversarial attacks such as distortions and perturbations. Gandhi and Jain~\cite{adversarial2020gandhi} explored Lipschitz regularization~\cite{Woods2019Adversarial} and Deep Image Prior (DIP)~\cite{Lempitsky2018Deep} to remove the artificial perturbations they generated and maintain model robustness. Yang et al.~\cite{Practical2021Yang} simulated commonly seen data corruption techniques on the benchmark datasets to increase data diversity in model training. Operations such as resolution down-scaling, bit-rate adjustment, and artifacting have effectively boosted the detection ability against in-the-wild Deepfake contents. Hooda et al.~\cite{hooda2022towards} designed a Disjoint Deepfake Detection (D3) detector that improves adversarial robustness against artificial perturbations using an ensemble of models. The LTTD~\cite{Delving2022Guan_NIPS} framework is enforced to conquer the challenges brought by post-processing procedures of Deepfake generation like visual compression to models that rely on low-level image feature patterns. Lin et al.~\cite{Lin2022Exploiting} proved the fact that having temporal information to participate in detection makes the detector less prone to black-box attacks.

Moreover, in the latest studies, researchers have frequently illustrated the necessity of robustness by fooling the well-trained detection models with stronger adversarial attacks~\cite{Hussain2022Exposing, goodman2020advbox}. Jia et al.~\cite{Exploring2022Jia_CVPR} even emphasized the robustness of the detection models against potential adversarial attacks by injecting frequency perturbations to fool the state-of-the-art approaches. Also, the robustness of Deepfake detectors is evaluated via the Fast Gradient Sign Method (FGSM) and the Carlini-Wagner L2-norm attack in several studies~\cite{adversarial2020gandhi, Evaluating2022Shahriyar}. Moreover, Carlini and Farid~\cite{Carlini2020Evading_CVPR_Workshops} employed white- and black-box attacks on a well-trained detector in five case studies. Obvious performance damping can be observed from the reported results. 
Faces with imperceptible visual variation even after perturbations and noises are added in have highlighted the importance of studying model robustness in the current Deepfake detection research domain. 

\section{Detection Model Reliability Study}
\label{reliability_study}

\subsection{Overview}
\label{overview_reliability}

In the ideal scenario, a reliable Deepfake detection model should retain promising transferability on unseen data with unknown synthetic techniques, pellucid explanation upon the falsification, and robust resistance against different real-life application conditions. As reviewed in Section~\ref{deepfake_detection}, despite a method favorably satisfying all three challenges simultaneously is not yet accomplished, a metric that nominates the reliability of a method for real-life usages and court-case judgments is needed.

Regardless of the largely improved but still unsatisfied cross-dataset performance in the evolution of Deepfake detection, existing studies have only evaluated the model performance on each testing dataset to show the model detection abilities while the values for each evaluation metric (accuracy and AUC score) vary depending on different testing sets adopted in experiments. On the contrary, in real-life cases, people have no clue about fake content regarding its source dataset or the corresponding facial manipulation techniques, and the malicious attacker is unlikely to reveal such crucial information. Consequently, for a victim of Deepfake to defend his or her innocence or accuse the attacker~\cite{Kelion2018Emma, Lee2018Portman, Rudy2021Cheerleader, Miller2022Zelensky}, simply presenting a model detection decision and listing the numerical model performance on each benchmark dataset may not be convincing and reliable. Specifically, a unique statistical claim is necessary regarding the detection performance to discuss the model trustworthiness on any arbitrary candidate suspect instead of varying on each testing dataset when adopting the detection model as forensic evidence for criminal investigation and court-case judgments. To conclude, the following questions are to be solved: 

\begin{itemize}
\item Can a detection model assist or act as evidence in forensic investigations in court?
\item How reliable is the detection model when performing as forensic evidence in real-life scenarios?
\item How accurate is the detection model regarding an arbitrary falsification?
\end{itemize}

Unfortunately, to the best of our knowledge, no existing work has studied the model reliability or come up with a reliable claim for the model to play the role of forensic evidence. Therefore, in this study, inspired by the reliability study on antigen diagnostic tests~\cite{FDA2022antigen, EU2022antigen, HK2022antigen} for the recent COVID-19 pandemic~\cite{Osterman2021Antigen}, we conduct a quantitative study by investigating the detection model reliability with a new evaluation metric with statistical techniques. Unlike the studies for antigen diagnostic tests that prefer to achieve a perfect specificity rather than sensitivity as the goal is to avoid missing any positive case~\cite{WHO2022}, we wish to correctly identify both real and fake materials with no priority. In particular, we construct a population to imitate real-life Deepfake distribution and design a scientific random sampling scheme to analyze and compute the confidence intervals for the values of accuracy and AUC score metrics regarding the Deepfake detection models. As a result, numerical ranges indicating the reliable model performance at 90\% and 95\% confidence levels can be derived for both accuracy and AUC score.

\subsection{Deepfake Population}
\label{population}

In reality, a candidate Deepfake suspect can only have two possible categories, namely, real and fake. Admittedly, most public images and videos circulating on the internet are pristine without artificial changes. However, whenever a real-life Deepfake case is raised such that the authenticity of the candidate material needs to be justified, we could not consider all images and videos in the world as the target population~\cite{target_population} because most of the real ones are not likely to be disputed in the discussion of Deepfake. At the same time, the probability that the candidate material is fake does not necessarily equal to the proportion of fake ones regarding all images and videos in the world. 

Despite the uncertainty of the real-life Deepfake population and distribution, in this study, we construct a sampling frame~\cite{sampling_frame} with the accessible high-quality Deepfake benchmark datasets to imitate the target population of Deepfake in real life for detection model reliability analysis. Details of the participating datasets are introduced in Section~\ref{datasets_involved}. 

\subsection{Random Sampling}
\label{sampling algorithm}

We perform random sampling~\cite{randsample2018Martino} from the constructed sampling frame with a sample size of $s$ for $t$ trials. Two sampling options are considered in this study: balanced and imbalanced. For a balanced sampling setting, we maintain the condition that the same amount of real and fake samples are randomly drawn. On the contrary, an imbalanced setting allows a completely random stochastic rule with respect to real and fake samples. 

For an arbitrary Deepfake detection model $M$ after sufficient training, we randomly draw $s$ samples from the sampling frame following the sampling option. Then the $s$ samples are fed to model $M$ for authenticity prediction, deriving predicted labels and prediction scores. After that, the accuracy and AUC score metrics are computed accordingly based on the ground-truth authenticities of the sampled faces. Such a sampling process is repeated for $t$ trials and a total of $t$ accuracies and $t$ AUC scores are derived in the end. Take the $t$ accuracies as an example, we first compute the mean value $\Bar{x}$ and standard deviation $\sigma$ following

\begin{equation}
    \Bar{x}=\frac{\Sigma^t_{i=1}x_i}{t},
\label{eq3}
\end{equation}
and

\begin{equation}
    \sigma=\sqrt{\frac{\Sigma^t_{i=1}(x_i-\Bar{x})^2}{t-1}},
\label{eq4}
\end{equation}
where $x_i$ refers to the accuracy value of the $i$-th trial. 

According to the central limit theorem (CLT)~\cite{CLT2011Fischer}, the distribution of sample means tends toward a normal distribution as the sample size gets larger. Therefore, the normal distribution confidence interval $CI$ can be calculated by

\begin{equation}
    CI=\Bar{x}\pm z\frac{\sigma}{\sqrt{s}},
\label{eq5}
\end{equation}
where parameter $z$ represents the z-score, an indicator of the confidence level following the instruction of the z-table [16]. When deducing statistical results for the $t$ AUC scores, the above workflow applies identically.

Since the target population is of unknown distribution, different values of the sample size $s$ are adopted in order to settle the confidence intervals at different confidence levels. Meanwhile, considering that an insufficient number of trials per sample size may cause bias when locating the sample mean, different values of trails for $t$ are chosen to eliminate the potential bias. Detailed steps of the model reliability study can be summarized as Algorithm~\ref{alg1}.

\renewcommand{\algorithmicrequire}{\textbf{Input:}}
\renewcommand{\algorithmicensure}{\textbf{Output:}}
\begin{algorithm}
\caption{Deepfake detection model reliability study.}
\label{alg1}
\begin{algorithmic}[1]
\REQUIRE $M=\textrm{well-trained detection model}$
\ENSURE 90\% and 95\% $CI$ for each sample size
\STATE$f_r \leftarrow \textrm{list of real samples}$
\STATE $f_f \leftarrow \textrm{list of fake samples}$
\STATE $o \leftarrow \textrm{balance sampling option}$
\STATE $t \leftarrow \textrm{number of trials}$
\STATE $S \leftarrow [s_1, s_2, ..., \textrm{len}(S)]$
\STATE $\textrm{shuffle}(\cdot)$ $\leftarrow$ function to shuffle the list
\STATE $\textrm{acc}(\cdot)$ $\leftarrow$ function to calculate accuracy
\STATE $\textrm{auc}(\cdot)$ $\leftarrow$ function to calculate AUC score
\FOR{$i \leftarrow 0$ \TO $\textrm{len}(S)-1$}
\STATE acc\_lst $\leftarrow$ []
\STATE auc\_lst $\leftarrow$ []
\FOR{$j \leftarrow 0$ \TO $t$}
\IF{$o==\textrm{True}$}
\STATE $f_r \leftarrow$ shuffle($f_r$)
\STATE $f_f \leftarrow$ shuffle($f_f$)
\STATE samples $\leftarrow f_r[0:\frac{S[i]}{2}] + f_f[0:\frac{S[i]}{2}]$
\ELSE
\STATE $f_a \leftarrow$ shuffle($f_r + f_f$)
\STATE samples $\leftarrow f_a[0:S[i]]$
\ENDIF
\STATE preds, pred\_scores, labels $\leftarrow M(\textrm{samples})$
\STATE acc\_lst.append(acc(preds, labels))
\STATE auc\_lst.append(auc(pred\_scores, labels))
\ENDFOR
\STATE compute $\Bar{x}$ and $\sigma$ for acc\_lst and auc\_lst
\STATE compute and record 90\% and 95\% $CI$
\ENDFOR
\end{algorithmic}
\end{algorithm}

\section{Dataset Preparation}
\label{deepfake_preparation}

The choice of training dataset and data pre-processing scheme can significantly affect the performance of a deep learning model. Among the evolution of Deepfake datasets, as introduced in Section~\ref{dataset_evolution}, various benchmark datasets are frequently adopted for training and testing to boost the detection model performance, but the workflow of data pre-processing operations has been barely discussed in detail in existing studies. Moreover, there lacks a standard pre-processing scheme in the current domain, causing difficulty in model comparison due to non-uniform training datasets after pre-processing by different detection work. On the other hand, using heedlessly prepared datasets for sampling frame construction can lead to improbable results towards the reliability study. Therefore, in this paper, we expound a standard and systematic workflow of data preparation and pre-processing to resolve the inconsistency and benefit both veterans and new starters in the research domain, ensuring a fair game for other work to compare with the model performance as exhibited in this paper following the same settings.

\subsection{Dataset Pre-processing}
\label{pre-processing}

While a video-level detector may rely on special data processing arrangement directly upon videos, for the frame-level detectors, since the detection results are evaluated based on all selected frames, to avoid potential biases toward particular videos, it is meaningful to keep the amount of extracted faces from each video equivalent during model training and testing. Therefore, we firstly obtain $c$ image frames using FFmpeg~\cite{ffmpeg} for each candidate video with an equal frame interval between every two adjacent extracted frames following

\begin{equation}
    P=\frac{iN}{c} \textrm{ for } 0 \leq i<c \textrm{ and } i\in\mathbb{Z},
\label{eq1}
\end{equation}
where $N$ refers to the number of frames that contain faces in the video and $P=\{p_0, p_1, ..., p_{c-1}\}$ contains the sequentially ordered indices for which frames to be extracted from the video. In other words, image frames with no face detected are excluded from the sequential ordering and indexing. Besides, videos with fewer than $c$ frames containing detected faces are also omitted. The dlib library~\cite{dlib} is utilized for face detection and cropping where the face detector provides coordinates of the bounding box that locates the detected face. For the sequence of frames with frame indices $P=\{p_0, p_1, ..., p_{c-1}\}$ from a video, we fix the size $l$ of a squared bounding box $b$ for all faces by

\begin{equation}
    l=\max\{\max_i w_i, \max_i h_i\} \textrm{ for } 0\leq i<c \textrm{ and } i\in\mathbb{Z},
\label{eq2}
\end{equation}
where $w_i$ and $h_i$ are the widths and heights of each bounding box. We then locate the center of each face $f_i$ with the help of the corresponding bounding box $b_i$ and place the fixed squared bounding box $b$ at the centers for face cropping. %Other special cases and solutions are listed in Appendix Table A.

\subsection{Datasets Involved and Detailed Arrangements}
\label{datasets_involved}

Following the convention of the existing Deepfake detection work and considering the qualities of available benchmark datasets, we consider five datasets in experiments in this study, namely, FF++~\cite{Rossler2019FaceForensics}, FaceShifter~\cite{Li2019FaceShifter}, DFDC~\cite{DFDC}, Celeb-DF~\cite{Li2020CelebDF}, and DF1.0~\cite{jiang2020deeperforensics1}. In detail, early datasets~\cite{ExposingDeep2019Li} are excluded due to low quantity and diversity. Meanwhile, although WDF~\cite{WildDeepfake2020} is similar to real-life Deepfake materials, the videos collected from the internet are manually labeled without knowing the ground-truth labels, which leads to credibility issues. KoDF~\cite{KoDF2021Kwon} is the largest Deepfake dataset up to date, but its huge magnitude requires unreasonably large storage (\textrm{$\sim$}4 TB) that we are unable to acquire and process\footnote[5]{The 6 manipulation algorithms in KoDF are highly overlapped with the 8 manipulation algorithms in DFDC. This favorably suggests that the adoption of DFDC satisfies the demand for model diversity even without KoDF. }. All involved benchmark datasets follow the pre-processing scheme for face extraction as discussed in Section~\ref{pre-processing}, and special settings are mentioned in the following subsections when necessary.

\subsubsection{FaceForensics++}
\label{ffpp}

FaceForensics++ (FF++) is currently the most widely adopted dataset in the existing Deepfake detection studies. The dataset contains 1,000 real videos collected from YouTube and 4,000 Deepfake videos synthesized based on the real ones. In specific, four facial manipulation techniques are each applied to the 1,000 real videos to derive the corresponding 1,000 fake ones. Among the four facial manipulation techniques, FaceSwap (FS)~\cite{faceswap} and Deepfakes (DF)~\cite{deepfakes} are face-swapping algorithms that synthesize the faces by swapping facial identities, while Face2Face (F2F)~\cite{Thies2016Face2Face} and NeuralTextures (NT)~\cite{Thies2019Deferred} perform face reenactment by modifying facial attributes such as expressions and accessories.

The FF++ dataset has provided a subject-independent official dataset split with a ratio of 720:140:140 for training, validation, and testing. Meanwhile, three dataset qualities have been released, namely, Raw, HQ (c23), and LQ (c40), where the latter two are compressed with different video compression levels following the H.264 codec. In recent Deepfake detection work, FF++ is frequently adopted as the training dataset due to its manipulation diversity and data orderliness, and the HQ (c23) version is mostly utilized because it has a similar video compression level and video quality to the real-life Deepfake contents. In this survey, whenever necessary, we adopt FF++ for model training following the official dataset split. The key image frames are also extracted and employed since the performance enhancement by the participating key image frames in the training process has been proved in the early studies~\cite{DCPT2022Wang, Afchar2018MesoNet, Li2020CelebDF}. In the training process, unless specially designed, commonly used data augmentation is performed upon the real faces to construct a balanced training dataset for real and fake. In the testing phase, the testing set is constructed following the official split without further augmentation.

\subsubsection{Deepfake Detection Challenge}
\label{dfdc}

Deepfake Detection Challenge (DFDC) is one of the largest public Deepfake datasets with 23,654 real videos and 104,500 fake ones. Among the fake videos, there are eight synthetic techniques~\cite{deepfakes, nirkin2019fsgan, mmnn2012Huang, stylegan2021karras, nth2019zakharov, audioswap2019Adam} that have been applied based on the real ones. Due to its large data quantity, we randomly pick 10 of the 50 video folders from the official dataset and randomly shuffle 100 real videos and 100 fake ones from each folder. Since most existing approaches focus on detecting Deepfake visually and most benchmark datasets are published without audio, fake videos using the pure audio swap technique are easily classified as pristine because there is no visual artifact on the faces. Meanwhile, the official DFDC dataset only provides labels for real and fake while sub-labels for specific synthetic techniques are unavailable. Therefore, in this paper, the randomly picked 1,000 fake videos are manually examined to eliminate fake videos with the pure audio swap technique to omit noises in detection and guarantee a fair experimental setting. The selected videos are then fed through the data pre-processing scheme in Section~\ref{pre-processing} for model evaluation.

\subsubsection{Celeb-DF}
\label{cdf}

Celeb-DF is one of the most challenging benchmark Deepfake datasets that are publicly available. It contains 590 celebrity interview videos collected from YouTube and 5,639 face-swapped videos based on the real ones using an improved face-swapping algorithm with resolution enhancement, mismatched color correction~\cite{Reinhard2001Color}, inaccurate face mask adjustment, and temporal flickering reducing~\cite{Kalman1960Filtering} on the basic face-swapping auto-encoder architecture. A set of 518 official testing videos with high visual quality has failed most of the existing baseline models at a time because obvious visual artifacts can be barely found. We resort to the official testing set with 178 real videos and 340 fake ones for model evaluation.

\subsubsection{DeeperForensics-1.0}
\label{df1.0}

DeeperForensics-1.0 (DF1.0) is the first large-scale dataset that is manually added with deliberate distortions and perturbations to the clean face-swapped videos. A strengthened face-swapping algorithm, Deepfake Variational Auto-Encoder (DF-VAE), is introduced for superior synthetic performance with better reenactments on expression and pose, fewer style mismatches, and more stable temporal continuity. There are a total of 10,000 synthesized videos where 1,000 of them are face-swapped from lab-controlled source videos onto the FF++ real videos using DF-VAE and the rest 9,000 videos are derived using the 1,000 raw manipulated videos by applying combinations of seven distortions\footnote[6]{Change of color saturation, local block-wise distortion, change of color contrast, Gaussian blur, white Gaussian noise in color components, JPEG compression, and change of video constant rate factor.} under five intensity levels. Since the HQ and LQ versions of FF++ contain the same visual content and only differ in compression levels, DF1.0 with sufficient visual quality diversity in the manipulated videos serves as a perfect substitution for the LQ version of FF++ in the experiments, providing a convincing evaluation of all quality circumstances. Produced based on FF++, the dataset has only provided the official split ratio of 7:1:2 for the fake videos, and we thus execute model evaluation with merely the 2,000 fake testing videos.

\subsubsection{FaceShifter}
\label{faceshifter}

FaceShifter refers to a subject-agnostic GAN-based face-swapping algorithm that solves the facial occlusion challenge with a novel Heuristic Error Acknowledging Refinement Network (HEAR-Net). A subset with 1,000 synthetic videos is later included in the FF++ dataset by applying the FaceShifter face-swapping model to the 1,000 real videos. Since FaceShifter and FF++ share the same set of real videos, we take only the 140 fake videos for model evaluation following the FF++ official split.

\section{Detection Model Evaluation}
\label{model_evaluation}

In this section, we first adopt several state-of-the-art Deepfake detection models that are mainly designed regarding each of the three challenges as defined in Section~\ref{deepfake_detection} and report their detection performance on each benchmark testing set. Then the models are further discussed regarding the reliability following Algorithm~\ref{alg1} along with case studies on real-life Deepfake materials.

\subsection{Experiment Settings}
\label{experiment_settings}

Based on the Deepfake detection developing history and the three challenges of the current Deepfake detection domain, in the experiment, we selected several representative milestone baseline models and the most recent ones that have source code publicly available for reproduction. Specifically, Xception~\cite{Chollet2017Xception}, MAT~\cite{Zhao2021Multi-Attentional}, and RECCE~\cite{RECCE2022Cao} mainly attempt on the transferability challenge, Stage5~\cite{zhang2017beyond} and FSTMatching~\cite{Explaining2022Dong_ECCV} focus on the interpretability topic, and MetricLearning~\cite{cao2021metric} and LRNet~\cite{sun2021improving} are designed for the robustness issue. Models with publicly available trained weights are directly adopted for evaluation if the model is trained on FF++ or special arrangements other than the five benchmark datasets are necessary during training. The rest models are trained on FF++ in our experiment as discussed in Section~\ref{deepfake_preparation} and all models converge commonly. The selected models are tested on all benchmark datasets. To guarantee complete fairness, we applied optimal parameter settings as reported in the corresponding published papers during training and testing.

During model testing, we recorded the video-level Deepfake detection performance. In particular, detection results of the cropped faces of each video are averaged to a unique output for detectors that are designed for detecting individual images. Methods that can directly generate a single output for each video are fed with raw videos via the corresponding processing scheme as provided by their published source codes. The well-trained models are firstly evaluated on the FF++ testing set for the in-dataset setting, in other words, tested on the same dataset they have seen during training. Then, to further validate the model performance on unseen datasets, the cross-dataset evaluation is conducted to test the models on DFDC, Celeb-DF, DF1.0, and FaceShifter. We set $N=10$ for training and $N=20$ for testing regarding Eq. (\ref{eq1}) for frame extraction during data pre-processing when applicable. 

We adopted accuracy (ACC) and AUC score at the video level as the evaluation metrics. In detail, the accuracy refers to the proportion of the correctly classified data items regarding all testing data, and the AUC score represents the area under the receiver operating characteristic (ROC) curve. In other words, the AUC score demonstrates the probability that a random positive sample scores higher than a random negative sample from the testing set, that is, the ability of the classifier to distinguish between real and fake faces. For testing sets that contain only fake samples, the AUC score is inapplicable and thus withdrawn.

\begin{table}[b]
\begin{center}
\caption{Quantitative video-level accuracy (ACC) and AUC score performance comparison on each testing set. ($\dag$: trained weights directly adopted for evaluation.)}
\resizebox{\textwidth}{!}{
\begin{tabular}{lcccccccc}
\toprule\noalign{\smallskip}
\multirow{3}{*}{Model} & \multicolumn{8}{c}{Test Dataset} \\
\noalign{\smallskip}\cmidrule{2-9}\noalign{\smallskip}
& \multicolumn{2}{c}{FF++~\cite{Rossler2019FaceForensics} } & \multicolumn{2}{c}{DFDC~\cite{DFDC} } & \multicolumn{2}{c}{Celeb-DF~\cite{Li2020CelebDF}} & DF1.0~\cite{jiang2020deeperforensics1}  & FaceShifter~\cite{Li2019FaceShifter} \\
\noalign{\smallskip}\cmidrule{2-9}\noalign{\smallskip}
& ACC & AUC & ACC & AUC & ACC & AUC & ACC & ACC \\
\noalign{\smallskip}\midrule\noalign{\smallskip}
Xception~\cite{Chollet2017Xception} & 93.92\% & 97.31\% & 65.21\% & 71.57\% & 70.27\% & 70.71\% & 56.87\% & 57.55\% \\
MAT~\cite{Zhao2021Multi-Attentional} & 97.40\% & 99.67\% & 66.63\% & 74.83\% & 71.81\% & 77.16\% & 41.74\% & 18.71\% \\
RECCE~\cite{RECCE2022Cao} & 90.72\% & 95.26\% & 62.06\% & 66.94\% & 71.81\% & 77.90\% & 51.21\% & 56.12\% \\
Stage5$^\dag$~\cite{yang2021beyond_ijcai} & 19.97\% & 50.21\% & 51.02\% & 48.08\% & 34.36\% & 39.88\% & 0.00\% & 0.00\% \\
FSTMatching~\cite{Explaining2022Dong_ECCV} & 81.33\% & 77.01\% & 44.88\% & 39.90\% & 38.61\% & 44.27\% & 25.68\% & 13.67\% \\
MetricLearning$^\dag$~\cite{cao2021metric} & 80.03\% & 77.71\% & 48.98\% & 61.89\% & 65.64\% & 60.37\% & 100.00\% & 100.00\% \\
LRNet$^\dag$~\cite{sun2021improving} & 55.22\% & 67.82\% & 53.41\% & 53.91\% & 51.54\% & 59.72\% & 52.19\% & 41.30\% \\
\noalign{\smallskip}\bottomrule
\end{tabular}
}
\label{tab2}
\end{center}
\end{table}

\subsection{Model Performance on Benchmark Testing Sets}
\label{detection_results}

Model performance for both in- and cross-dataset evaluations is listed in Table~\ref{tab2}. It can be observed that all models of the transferability topic have derived reasonable detection performance on FF++ with accuracy values and AUC scores over 90\% since their goal is to achieve better performance in cross-dataset experiments after maintaining promising performance on seen data. In particular, MAT~\cite{Zhao2021Multi-Attentional} wins the comparison with the highest 97.40\% accuracy and 99.67\% AUC score. On the contrary, models that are designed specifically for interpretability or robustness purposes have exhibited relatively poor detection performance on FF++, and the potential causation is discussed together with their detection performance on other benchmark datasets in the following paragraphs.  

As for cross-dataset evaluation, most models have suffered a performance damping since the testing data are unseen during training. In specific, no model has reached over 80\% AUC scores on DFDC or Celeb-DF and some models even exhibit abnormal detection performance when validated on unseen fake testing sets solely (DF1.0 and FaceShifter). This may be caused by oblivious overfitting on real or fake data by the models. While models that are solving the transferability challenge have all exhibited normal and reasonable functionalities, hidden trouble can be discovered regarding the interpretability and robustness topics. Specifically, model weights of Stage5~\cite{yang2021beyond_ijcai} are adopted for testing because the model is trained with re-synthesized samples using exclusively GAN models under special settings, but this at the same time has led to unsatisfied results when detecting fake samples that are not synthesized using GAN architectures. FSTMatching~\cite{Explaining2022Dong_ECCV} spends huge computing power on disentangling source and target artifacts for the explanation, which thus results in the failure against other models although similarly trained on FF++. MetricLearning~\cite{cao2021metric} and LRNet~\cite{sun2021improving} both are proposed and trained to deal with highly compressed Deepfake contents in special scenarios. Unfortunately, they suffer performance fluctuation when the compression condition varies without expectation. Furthermore, LRNet~\cite{sun2021improving} executes detection based on facial landmarks solely, which is another main reason that leads to the unsatisfactory. 

Besides, models are generally unstable on different testing sets. For instance, RECCE~\cite{RECCE2022Cao} wins the competition on Celeb-DF for both accuracy and AUC score, but its detection ability deteriorates immensely when facing DFDC. On the other hand, MAT~\cite{Zhao2021Multi-Attentional} wins the battle on DFDC and achieves competitive performance on Celeb-DF, but an obvious overdependence on the real samples can be concluded from its poor accuracies on DF1.0 and FaceShfiter. Meanwhile, Xception has derived reasonable detection performance on each testing set even though not winning the comparison on any dataset. As a result, no model appears to be the overall winner according to Table~\ref{tab2} and it is hard to determine which model to use when facing an arbitrary candidate Deepfake suspect in real-life cases. 

Moreover, in most cases, a well-trained model usually achieves a higher AUC score than the accuracy on each testing set. The reason is that the threshold to classify real and fake with softmax or sigmoid function applied is always fixed at 0.5 for the accuracy evaluation upon the output scores within the range of [0, 1] where 0 refers to real and 1 represents fake, while the actual threshold for the optimal model performance is usually located differently regarding 0.5. Therefore, despite a classifier with a threshold value set to 0.5 does not perform well, the model may still distinguish between real and fake with a relatively high AUC score. However, it is also worth noting that although a high AUC score may reveal the model's ability to separate real and fake samples, the threshold may vary depending on different testing sets and different models. Hence, in order to stably determine real or fake, finding a fixed threshold to consistently satisfy the detection goal on arbitrary images and videos may help boost the overall model detection ability in the research domain.

\begin{table}[t]
\begin{center}
\caption{Dataset statistics of the sampling frame for model testing regarding the number of videos with cropped faces. Datasets with no real samples are marked with `--' sign.}
\resizebox{0.8\textwidth}{!}{
\begin{tabular}{lcccccc}
\toprule\noalign{\smallskip}
& FF++~\cite{Rossler2019FaceForensics} & DFDC~\cite{DFDC} & Celeb-DF~\cite{Li2020CelebDF} & DF1.0~\cite{jiang2020deeperforensics1} & FaceShifter~\cite{Li2019FaceShifter} & Total \\
\noalign{\smallskip}\midrule\noalign{\smallskip}
Num Real & 140 & 1,000 & 178 & -- & -- & 1,318 \\
Num Fake & 560 & 1,000 & 340 & 2,010 & 140 & 4,050 \\
Total & 700 & 2,000 & 518 & 2,010 & 140 & 5,368 \\
\noalign{\smallskip}\bottomrule
\end{tabular}
}
\label{tab3}
\end{center}
\end{table}

% \begin{table}[b]
% \begin{center}
% \caption{Dataset statistics of the sampling frame for model testing regarding the number of videos with cropped faces. Datasets with no real samples are marked with `--' sign.}
% \begin{tabular}{lccc}
% \toprule\noalign{\smallskip}
% Dataset & Num Real & Num Fake & Total \\
% \noalign{\smallskip}\midrule\noalign{\smallskip}
% FF++~\cite{Rossler2019FaceForensics} & 140 & 560 & 700 \\
% DFDC~\cite{DFDC} & 1,000 & 1,000 & 2,000 \\
% Celeb-DF~\cite{Li2020CelebDF} & 178 & 340 & 518 \\
% DF1.0~\cite{jiang2020deeperforensics1} & -- & 2,010 & 2,010 \\
% FaceShifter~\cite{Li2019FaceShifter} & -- & 140 & 140 \\
% Total & 1,318 & 4,050 & 5,368 \\
% \noalign{\smallskip}\bottomrule
% \end{tabular}
% \label{tab3}
% \end{center}
% \end{table}

\begin{table}
\begin{center}
\caption{Accuracy statistics (\%) with balanced sampling for 500 and 3,000 trials. }
\resizebox{\textwidth}{!}{
\begin{tabular}{lccccccccc}
\toprule
\multirow{2}{*}{Model Names} & \multirow{2}{*}{Num Trials} & \multirow{2}{*}{Statistics} & \multicolumn{7}{c}{Sample Sizes} \\
\cmidrule{4-10}
& & & 10 & 100 & 500 & 1,000 & 1,500 & 2,000 & 2,500 \\
\midrule
\multirow{8}{*}{Xception~\cite{Chollet2017Xception}} 
& \multirow{4}{*}{500}
& 90\% CI & 57.89--79.55 & 65.67--72.21 & 67.40--70.24 & 67.94--69.78 & 68.22--69.64 & 68.33--69.39 & 68.42--69.24 \\
& & 95\% CI & 55.81--81.63 & 65.04--72.84 & 67.13--70.51 & 67.76--69.96 & 68.08--69.78 & 68.23--69.49 & 68.34--69.31 \\
& & Mean & 68.72 & 68.94 & 68.82 & 68.86 & 68.93 & 68.86 & 68.83 \\
& & Std. & 14.70 & 4.44 & 1.92 & 1.25 & 0.97 & 0.72 & 0.56 \\
\cmidrule{2-10}
& \multirow{4}{*}{3,000}
& 90\% CI & 58.90--78.54 & 65.87--71.98 & 67.55--70.16 & 67.98--69.70 & 68.25--69.50 & 68.37--69.36 & 68.50--69.26 \\
& & 95\% CI & 57.02--80.42 & 65.29--72.57 & 67.31--70.41 & 67.82--69.87 & 68.13--69.62 & 68.28--69.45 & 68.42--69.34 \\
& & Mean & 68.72 & 68.93 & 68.86 & 68.84 & 68.88 & 68.86 & 68.88 \\
& & Std. & 14.62 & 4.55 & 1.94 & 1.28 & 0.93 & 0.73 & 0.57 \\
\midrule
\multirow{8}{*}{MAT~\cite{Zhao2021Multi-Attentional}} 
& \multirow{4}{*}{500}
& 90\% CI & 59.72--77.44 & 66.23--71.87 & 67.59--70.06 & 68.18--69.72 & 68.30--69.52 & 68.40--69.38 & 68.52--69.31 \\
& & 95\% CI & 58.02--79.14 & 65.69--72.41 & 67.35--70.30 & 68.03--69.87 & 68.19--69.64 & 68.31--69.47 & 68.45--69.39 \\
& & Mean & 68.58 & 69.05 & 68.82 & 68.95 & 68.91 & 68.89 & 68.92 \\
& & Std. & 13.17 & 4.19 & 1.83 & 1.15 & 0.91 & 0.73 & 0.59 \\
\cmidrule{2-10}
& \multirow{4}{*}{3,000}
& 90\% CI & 59.45--77.74 & 66.10--71.84 & 67.67--70.15 & 68.04--69.71 & 68.28--69.50 & 68.38--69.37 & 68.49--69.29 \\
& & 95\% CI & 57.70--79.49 & 65.55--72.39 & 67.43--70.39 & 67.88--69.87 & 68.16--69.62 & 68.29--69.47 & 68.41--69.37 \\
& & Mean & 68.60 & 68.97 & 68.91 & 68.87 & 68.89 & 68.88 & 68.89 \\
& & Std. & 13.61 & 4.27 & 1.85 & 1.24 & 0.91 & 0.74 & 0.60 \\
\midrule
\multirow{8}{*}{RECCE~\cite{RECCE2022Cao}} 
& \multirow{4}{*}{500}
& 90\% CI & 51.46--71.14 & 57.49--64.22 & 59.38--62.06 & 59.95--61.81 & 60.16--61.43 & 60.30--61.26 & 60.42--61.17 \\
& & 95\% CI & 49.56--73.04 & 56.84--64.86 & 59.13--62.32 & 59.77--61.99 & 60.03--61.55 & 60.20--61.35 & 60.35--61.24 \\
& & Mean & 61.30 & 60.85 & 60.72 & 60.88 & 60.79 & 60.78 & 60.79 \\
& & Std. & 14.63 & 5.00 & 1.99 & 1.38 & 0.94 & 0.72 & 0.56 \\
\cmidrule{2-10}
& \multirow{4}{*}{3,000}
& 90\% CI & 49.56--72.51 & 57.19--64.21 & 59.28--62.27 & 59.78--61.76 & 60.08--61.52 & 60.25--61.35 & 60.37--61.22 \\
& & 95\% CI & 47.36--74.71 & 56.52--64.88 & 58.99--62.55 & 59.59--61.95 & 59.94--61.66 & 60.15--61.46 & 60.29--61.30 \\
& & Mean & 61.04 & 60.70 & 60.77 & 60.77 & 60.80 & 60.80 & 60.80 \\
& & Std. & 15.59 & 4.77 & 2.03 & 1.35 & 0.98 & 0.75 & 0.58 \\
\midrule
\multirow{8}{*}{Stage5~\cite{yang2021beyond_ijcai}}
& \multirow{4}{*}{500}
& 90\% CI & 50.00--50.00 & 50.00--50.00 & 50.00--50.00 & 50.00--50.00 & 50.00--50.00 & 50.00--50.00 & 50.00--50.00 \\
& & 95\% CI & 50.00--50.00 & 50.00--50.00 & 50.00--50.00 & 50.00--50.00 & 50.00--50.00 & 50.00--50.00 & 50.00--50.00 \\
& & Mean & 50.00 & 50.00 & 50.00 & 50.00 & 50.00 & 50.00 & 50.00 \\
& & Std. & 0.00 & 0.00 & 0.00 & 0.00 & 0.00 & 0.00 & 0.00 \\
\cmidrule{2-10}
& \multirow{4}{*}{3,000}
& 90\% CI & 50.00--50.00 & 50.00--50.00 & 50.00--50.00 & 50.00--50.00 & 50.00--50.00 & 50.00--50.00 & 50.00--50.00 \\
& & 95\% CI & 50.00--50.00 & 50.00--50.00 & 50.00--50.00 & 50.00--50.00 & 50.00--50.00 & 50.00--50.00 & 50.00--50.00 \\
& & Mean & 50.00 & 50.00 & 50.00 & 50.00 & 50.00 & 50.00 & 50.00 \\
& & Std. & 0.00 & 0.00 & 0.00 & 0.00 & 0.00 & 0.00 & 0.00 \\
\midrule
\multirow{8}{*}{FSTMatching~\cite{Explaining2022Dong_ECCV}} 
& \multirow{4}{*}{500}
& 90\% CI & 44.61--59.71 & 49.85--54.76 & 51.42--53.37 & 51.68--53.01 & 51.91--52.87 & 51.96--52.78 & 52.06--52.69 \\
& & 95\% CI & 43.16--61.16 & 49.37--55.23 & 51.24--53.56 & 51.55--53.14 & 51.82--52.96 & 51.88--52.86 & 52.00--52.75 \\
& & Mean & 52.16 & 52.30 & 52.40 & 52.34 & 52.39 & 52.37 & 52.37 \\
& & Std. & 12.12 & 3.95 & 1.56 & 1.07 & 0.77 & 0.66 & 0.50 \\
\cmidrule{2-10}
& \multirow{4}{*}{3,000}
& 90\% CI & 44.62--59.82 & 50.15--54.80 & 51.42--53.42 & 51.73--53.07 & 51.85--52.84 & 51.98--52.78 & 52.06--52.70 \\
& & 95\% CI & 43.16--61.28 & 49.71--55.24 & 51.23--53.61 & 51.60--53.19 & 51.76--52.94 & 51.91--52.85 & 51.99--52.76 \\
& & Mean & 52.22 & 52.48 & 52.42 & 52.40 & 52.35 & 52.38 & 52.38 \\
& & Std. & 12.23 & 3.73 & 1.60 & 1.07 & 0.80 & 0.64 & 0.52 \\
\midrule
\multirow{8}{*}{MetricLearning~\cite{cao2021metric}} 
& \multirow{4}{*}{500}
& 90\% CI & 50.00--50.00 & 50.00--50.00 & 50.00--50.00 & 50.00--50.00 & 50.00--50.00 & 50.00--50.00 & 50.00--50.00 \\
& & 95\% CI & 50.00--50.00 & 50.00--50.00 & 50.00--50.00 & 50.00--50.00 & 50.00--50.00 & 50.00--50.00 & 50.00--50.00 \\
& & Mean & 50.00 & 50.00 & 50.00 & 50.00 & 50.00 & 50.00 & 50.00 \\
& & Std. & 0.00 & 0.00 & 0.00 & 0.00 & 0.00 & 0.00 & 0.00 \\
\cmidrule{2-10}
& \multirow{4}{*}{3,000}
& 90\% CI & 50.00--50.00 & 50.00--50.00 & 50.00--50.00 & 50.00--50.00 & 50.00--50.00 & 50.00--50.00 & 50.00--50.00 \\
& & 95\% CI & 50.00--50.00 & 50.00--50.00 & 50.00--50.00 & 50.00--50.00 & 50.00--50.00 & 50.00--50.00 & 50.00--50.00 \\
& & Mean & 50.00 & 50.00 & 50.00 & 50.00 & 50.00 & 50.00 & 50.00 \\
& & Std. & 0.00 & 0.00 & 0.00 & 0.00 & 0.00 & 0.00 & 0.00 \\
\midrule
\multirow{8}{*}{LRNet~\cite{sun2021improving}} 
& \multirow{4}{*}{500}
& 90\% CI & 41.70--62.46 & 48.22--55.13 & 50.53--53.24 & 51.09--52.97 & 51.31--52.72 & 51.42--52.51 & 51.62--52.37 \\
& & 95\% CI & 39.70--64.46 & 47.56--55.79 & 50.27--53.51 & 50.91--53.15 & 51.18--52.86 & 51.32--52.61 & 51.55--52.44 \\
& & Mean & 52.08 & 51.67 & 51.89 & 52.03 & 52.02 & 51.96 & 52.00 \\
& & Std. & 15.43 & 5.13 & 2.02 & 1.40 & 1.05 & 0.81 & 0.55 \\
\cmidrule{2-10}
& \multirow{4}{*}{3,000}
& 90\% CI & 41.70--61.81 & 49.05--55.25 & 50.74--53.29 & 51.15--52.80 & 51.38--52.64 & 51.51--52.49 & 51.62--52.36 \\
& & 95\% CI & 39.77--63.73 & 48.46--55.84 & 50.49--53.54 & 50.99--52.96 & 51.26--52.77 & 51.41--52.58 & 51.55--52.43 \\
& & Mean & 51.75 & 52.15 & 52.01 & 51.98 & 52.01 & 52.00 & 51.99 \\
& & Std. & 16.17 & 4.98 & 2.05 & 1.33 & 1.01 & 0.79 & 0.59 \\
\bottomrule
\end{tabular}
}
\label{tab4}
\end{center}
\end{table}

\begin{table}
\begin{center}
\caption{Accuracy statistics (\%) with imbalanced sampling for 500 and 3,000 trials. }
\resizebox{\textwidth}{!}{
\begin{tabular}{lccccccccc}
\toprule
\multirow{2}{*}{Model Names} & \multirow{2}{*}{Num Trials} & \multirow{2}{*}{Statistics} & \multicolumn{7}{c}{Sample Sizes} \\
\cmidrule{4-10}
& & & 10 & 100 & 500 & 1,000 & 1,500 & 2,000 & 2,500 \\
\midrule
\multirow{8}{*}{Xception~\cite{Chollet2017Xception}} 
& \multirow{4}{*}{500}
& 90\% CI & 55.90--75.50 & 62.70--69.19 & 64.66--67.62 & 65.08--67.19 & 65.48--67.06 & 65.55--66.91 & 65.58--66.89 \\
& & 95\% CI & 54.02--77.38 & 62.07--69.82 & 64.37--67.90 & 64.88--67.39 & 65.32--67.21 & 65.42--67.05 & 65.45--67.02 \\
& & Mean & 65.70 & 65.94 & 66.14 & 66.13 & 66.27 & 66.23 & 66.23 \\
& & Std. & 14.56 & 4.83 & 2.20 & 1.57 & 1.18 & 1.01 & 0.97 \\
\cmidrule{2-10}
& \multirow{4}{*}{3,000}
& 90\% CI & 55.36--77.38 & 62.75--69.79 & 64.74--67.81 & 65.06--67.25 & 65.32--67.06 & 65.44--66.99 & 65.51--66.94 \\
& & 95\% CI & 53.24--79.50 & 62.08--70.46 & 64.44--68.10 & 64.85--67.46 & 65.15--67.23 & 65.29--67.13 & 65.37--67.07 \\
& & Mean & 66.37 & 66.27 & 66.27 & 66.15 & 66.19 & 66.21 & 66.22 \\
& & Std. & 14.97 & 4.78 & 2.09 & 1.49 & 1.19 & 1.05 & 0.97 \\
\midrule
\multirow{8}{*}{MAT~\cite{Zhao2021Multi-Attentional}} 
& \multirow{4}{*}{500}
& 90\% CI & 50.11--70.45 & 57.75--64.32 & 59.17--61.98 & 59.61--61.72 & 59.82--61.55 & 59.93--61.38 & 59.93--61.28 \\
& & 95\% CI & 48.15--72.41 & 57.12--64.95 & 58.90--62.24 & 59.41--61.92 & 59.65--61.72 & 59.79--61.52 & 59.80--61.41 \\
& & Mean & 60.28 & 61.03 & 60.57 & 60.66 & 60.68 & 60.66 & 60.61 \\
& & Std. & 15.12 & 4.88 & 2.08 & 1.57 & 1.29 & 1.08 & 1.00 \\
\cmidrule{2-10}
& \multirow{4}{*}{3,000}
& 90\% CI & 49.79--70.58 & 57.22--63.89 & 59.21--62.13 & 59.57--61.64 & 59.80--61.52 & 59.90--61.39 & 59.97--61.26 \\
& & 95\% CI & 47.80--72.57 & 56.59--64.53 & 58.94--62.40 & 59.37--61.83 & 59.64--61.69 & 59.76--61.53 & 59.85--61.38 \\
& & Mean & 60.18 & 60.56 & 60.67 & 60.60 & 60.66 & 60.64 & 60.61 \\
& & Std. & 15.47 & 4.96 & 2.17 & 1.54 & 1.28 & 1.11 & 0.96 \\
\midrule
\multirow{8}{*}{RECCE~\cite{RECCE2022Cao}} 
& \multirow{4}{*}{500}
& 90\% CI & 53.35--73.89 & 59.48--66.01 & 61.53--64.47 & 62.15--64.17 & 62.37--64.00 & 62.30--63.70 & 62.43--63.76 \\
& & 95\% CI & 51.37--75.87 & 58.86--66.64 & 61.25--64.75 & 61.96--64.36 & 62.21--64.16 & 62.16--63.84 & 62.30--63.88 \\
& & Mean & 63.62 & 62.75 & 63.00 & 63.16 & 63.18 & 63.00 & 63.09 \\
& & Std. & 15.27 & 4.85 & 2.18 & 1.50 & 1.21 & 1.04 & 0.98 \\
\cmidrule{2-10}
& \multirow{4}{*}{3,000}
& 90\% CI & 52.85--73.05 & 59.88--66.32 & 61.64--64.53 & 62.01--64.07 & 62.25--63.89 & 62.30--63.77 & 62.36--63.66 \\
& & 95\% CI & 50.91--74.99 & 59.26--66.94 & 61.36--64.81 & 61.81--64.27 & 62.09--64.04 & 62.16--63.91 & 62.24--63.79 \\
& & Mean & 62.95 & 63.10 & 63.09 & 63.04 & 63.07 & 63.04 & 63.01 \\
& & Std. & 15.04 & 4.80 & 2.15 & 1.54 & 1.22 & 1.09 & 0.97 \\
\midrule
\multirow{8}{*}{Stage5~\cite{yang2021beyond_ijcai}}
& \multirow{4}{*}{500}
& 90\% CI & 16.38--33.50 & 22.20--27.30 & 23.73--26.06 & 24.04--25.65 & 24.20--25.34 & 24.33--25.28 & 24.37--25.15 \\
& & 95\% CI & 14.73--35.15 & 21.71--27.79 & 23.50--26.28 & 23.88--25.80 & 24.10--25.45 & 24.24--25.37 & 24.30--25.23 \\
& & Mean & 24.94 & 24.75 & 24.89 & 24.84 & 24.77 & 24.80 & 24.76 \\
& & Std. & 13.75 & 4.09 & 1.87 & 1.29 & 0.91 & 0.76 & 0.63 \\
\cmidrule{2-10}
& \multirow{4}{*}{3,000}
& 90\% CI & 16.38--33.20 & 22.13--27.53 & 23.61--25.96 & 24.02--25.54 & 24.21--25.37 & 24.31--25.25 & 24.40--25.17 \\
& & 95\% CI & 14.77--34.82 & 21.62--28.05 & 23.39--26.18 & 23.88--25.69 & 24.10--25.49 & 24.22--25.34 & 24.33--25.25 \\
& & Mean & 24.79 & 24.83 & 24.79 & 24.78 & 24.79 & 24.78 & 24.79 \\
& & Std. & 13.52 & 4.34 & 1.89 & 1.22 & 0.93 & 0.76 & 0.62 \\
\midrule
\multirow{8}{*}{FSTMatching~\cite{Explaining2022Dong_ECCV}} 
& \multirow{4}{*}{500}
& 90\% CI & 31.38--51.22 & 37.87--44.02 & 39.65--42.21 & 40.20--41.92 & 40.43--41.76 & 40.56--41.64 & 40.67--41.54 \\
& & 95\% CI & 29.47--53.13 & 37.28--44.61 & 39.40--42.45 & 40.03--42.08 & 40.31--41.88 & 40.45--41.74 & 40.58--41.62 \\
& & Mean & 41.30 & 40.95 & 40.93 & 41.06 & 41.09 & 41.10 & 41.10 \\
& & Std. & 15.93 & 4.94 & 2.06 & 1.38 & 1.06 & 0.87 & 0.70 \\
\cmidrule{2-10}
& \multirow{4}{*}{3,000}
& 90\% CI & 31.47--50.66 & 38.02--44.20 & 39.83--42.47 & 40.19--41.94 & 40.42--41.74 & 40.56--41.64 & 40.65--41.53 \\
& & 95\% CI & 29.63--52.50 & 37.43--44.79 & 39.58--42.72 & 40.02--42.11 & 40.29--41.87 & 40.45--41.75 & 40.57--41.62 \\
& & Mean & 41.07 & 41.11 & 41.15 & 41.06 & 41.08 & 41.10 & 41.09 \\
& & Std. & 15.43 & 4.97 & 2.12 & 1.41 & 1.06 & 0.87 & 0.71 \\
\midrule
\multirow{8}{*}{MetricLearning~\cite{cao2021metric}} 
& \multirow{4}{*}{500}
& 90\% CI & 66.17--83.91 & 72.88--77.98 & 74.09--76.40 & 74.42--76.06 & 74.65--75.78 & 74.69--75.67 & 74.86--75.59 \\
& & 95\% CI & 64.47--85.61 & 72.39--78.47 & 73.87--76.62 & 74.27--76.22 & 74.54--75.89 & 74.60--75.77 & 74.79--75.66 \\
& & Mean & 75.04 & 75.43 & 75.25 & 75.24 & 75.22 & 75.18 & 75.22 \\
& & Std. & 14.23 & 4.09 & 1.85 & 1.31 & 0.91 & 0.79 & 0.59 \\
\cmidrule{2-10}
& \multirow{4}{*}{3,000}
& 90\% CI & 66.53--83.60 & 72.81--78.12 & 74.05--76.36 & 74.46--75.97 & 74.66--75.83 & 74.75--75.67 & 74.85--75.62 \\
& & 95\% CI & 64.89--85.24 & 72.30--78.63 & 73.83--76.58 & 74.32--76.12 & 74.54--75.95 & 74.66--75.76 & 74.77--75.69 \\
& & Mean & 75.07 & 75.46 & 75.20 & 75.22 & 75.25 & 75.21 & 75.23 \\
& & Std. & 13.73 & 4.28 & 1.86 & 1.22 & 0.95 & 0.74 & 0.62 \\
\midrule
\multirow{8}{*}{LRNet~\cite{sun2021improving}} 
& \multirow{4}{*}{500}
& 90\% CI & 44.08--62.36 & 49.99--55.83 & 51.62--54.32 & 51.89--53.62 & 52.08--53.41 & 52.14--53.19 & 52.19--53.07 \\
& & 95\% CI & 42.32--64.12 & 49.43--56.40 & 51.36--54.58 & 51.72--53.79 & 51.96--53.53 & 52.04--53.29 & 52.10--53.16 \\
& & Mean & 53.22 & 52.91 & 52.97 & 52.75 & 52.74 & 52.66 & 52.63 \\
& & Std. & 14.68 & 4.69 & 2.17 & 1.39 & 1.06 & 0.84 & 0.71 \\
\cmidrule{2-10}
& \multirow{4}{*}{3,000}
& 90\% CI & 42.91--62.33 & 49.72--55.84 & 51.33--53.99 & 51.79--53.58 & 51.99--53.36 & 52.13--53.22 & 52.23--53.14 \\
& & 95\% CI & 41.05--64.19 & 49.13--56.43 & 51.08--54.25 & 51.62--53.76 & 51.86--53.49 & 52.03--53.33 & 52.15--53.23 \\
& & Mean & 52.62 & 52.78 & 52.66 & 52.69 & 52.68 & 52.68 & 52.69 \\
& & Std. & 15.61 & 4.92 & 2.14 & 1.44 & 1.10 & 0.88 & 0.73 \\
\bottomrule
\end{tabular}
}
\label{tab5}
\end{center}
\end{table}

\begin{table}
\begin{center}
\caption{AUC score statistics (\%) with balanced sampling for 500 and 3,000 trials. }
\resizebox{\textwidth}{!}{
\begin{tabular}{lccccccccc}
\toprule
\multirow{2}{*}{Model Names} & \multirow{2}{*}{Num Trials} & \multirow{2}{*}{Statistics} & \multicolumn{7}{c}{Sample Sizes} \\
\cmidrule{4-10}
& & & 10 & 100 & 500 & 1,000 & 1,500 & 2,000 & 2,500 \\
\midrule
\multirow{8}{*}{Xception~\cite{Chollet2017Xception}} 
& \multirow{4}{*}{500}
& 90\% CI & 68.57--84.16 & 74.21--78.76 & 75.67--77.53 & 76.06--77.32 & 76.18--77.13 & 76.29--77.04 & 76.38--76.92 \\
& & 95\% CI & 67.08--85.66 & 73.77--79.20 & 75.49--77.71 & 75.94--77.44 & 76.09--77.22 & 76.22--77.11 & 76.33--76.98 \\
& & Mean & 76.37 & 76.49 & 76.60 & 76.69 & 76.65 & 76.66 & 76.65 \\
& & Std. & 15.69 & 4.58 & 1.87 & 1.27 & 0.95 & 0.76 & 0.55 \\
\cmidrule{2-10}
& \multirow{4}{*}{3,000}
& 90\% CI & 69.14--84.68 & 74.50--79.02 & 75.69--77.64 & 76.02--77.29 & 76.23--77.14 & 76.30--77.02 & 76.36--76.93 \\
& & 95\% CI & 67.65--86.17 & 74.07--79.45 & 75.51--77.82 & 75.90--77.42 & 76.14--77.23 & 76.23--77.09 & 76.30--76.98 \\
& & Mean & 76.91 & 76.76 & 76.66 & 76.66 & 76.69 & 76.66 & 76.64 \\
& & Std. & 15.67 & 4.55 & 1.96 & 1.28 & 0.92 & 0.73 & 0.57 \\
\midrule
\multirow{8}{*}{MAT~\cite{Zhao2021Multi-Attentional}} 
& \multirow{4}{*}{500}
& 90\% CI & 66.83--83.07 & 73.00--77.79 & 74.79--76.75 & 74.88--76.21 & 75.13--76.16 & 75.31--76.12 & 75.31--75.93 \\
& & 95\% CI & 65.27--84.63 & 72.54--78.25 & 74.60--76.94 & 74.75--76.34 & 75.03--76.26 & 75.23--76.20 & 75.25--75.99 \\
& & Mean & 74.95 & 75.39 & 75.77 & 75.54 & 75.64 & 75.71 & 75.62 \\
& & Std. & 16.34 & 4.81 & 1.97 & 1.34 & 1.05 & 0.82 & 0.62 \\
\cmidrule{2-10}
& \multirow{4}{*}{3,000}
& 90\% CI & 67.60--83.56 & 73.37--78.00 & 74.68--76.61 & 74.94--76.27 & 75.13--76.12 & 75.24--76.03 & 75.33--75.96 \\
& & 95\% CI & 66.07--85.09 & 72.93--78.45 & 74.50--76.79 & 74.82--76.40 & 75.03--76.21 & 75.17--76.10 & 75.27--76.02 \\
& & Mean & 75.58 & 75.69 & 75.65 & 75.61 & 75.62 & 75.63 & 75.64 \\
& & Std. & 16.09 & 4.67 & 1.94 & 1.34 & 1.00 & 0.79 & 0.64 \\
\midrule
\multirow{8}{*}{RECCE~\cite{RECCE2022Cao}} 
& \multirow{4}{*}{500}
& 90\% CI & 57.27--75.43 & 63.82--69.22 & 65.88--68.13 & 66.02--67.53 & 66.18--67.32 & 66.39--67.27 & 66.40--67.14 \\
& & 95\% CI & 55.52--77.18 & 63.30--69.74 & 65.66--68.35 & 65.87--67.67 & 66.07--67.43 & 66.31--67.35 & 66.33--67.21 \\
& & Mean & 66.35 & 66.52 & 67.00 & 66.77 & 66.75 & 66.83 & 66.77 \\
& & Std. & 18.28 & 5.43 & 2.26 & 1.52 & 1.14 & 0.88 & 0.75 \\
\cmidrule{2-10}
& \multirow{4}{*}{3,000}
& 90\% CI & 57.80--75.48 & 64.02--69.42 & 65.65--67.90 & 66.05--67.59 & 66.25--67.41 & 66.35--67.26 & 66.44--67.19 \\
& & 95\% CI & 56.11--77.17 & 63.50--69.94 & 65.43--68.12 & 65.90--67.73 & 66.14--67.52 & 66.26--67.35 & 66.37--67.26 \\
& & Mean & 66.64 & 66.72 & 66.77 & 66.82 & 66.83 & 66.80 & 66.82 \\
& & Std. & 17.81 & 5.45 & 2.27 & 1.55 & 1.17 & 0.92 & 0.75 \\
\midrule
\multirow{8}{*}{Stage5~\cite{yang2021beyond_ijcai}}
& \multirow{4}{*}{500}
& 90\% CI & 42.69--57.45 & 48.82--52.75 & 49.70--51.45 & 49.80--50.95 & 50.07--50.95 & 50.17--50.88 & 50.20--50.76 \\
& & 95\% CI & 41.27--58.86 & 48.44--53.12 & 49.53--51.62 & 49.68--51.07 & 49.99--51.03 & 50.10--50.95 & 50.15--50.81 \\
& & Mean & 50.07 & 50.78 & 50.57 & 50.38 & 50.51 & 50.53 & 50.48 \\
& & Std. & 20.02 & 5.33 & 2.37 & 1.57 & 1.19 & 0.97 & 0.76 \\
\cmidrule{2-10}
& \multirow{4}{*}{3,000}
& 90\% CI & 43.18--57.79 & 48.23--52.59 & 49.59--51.37 & 49.81--51.03 & 50.06--50.96 & 50.13--50.83 & 50.22--50.75 \\
& & 95\% CI & 41.78--59.20 & 47.81--53.01 & 49.42--51.55 & 49.70--51.14 & 49.97--51.04 & 50.06--50.90 & 50.16--50.80 \\
& & Mean & 50.49 & 50.41 & 50.48 & 50.42 & 50.51 & 50.48 & 50.48 \\
& & Std. & 19.36 & 5.78 & 2.36 & 1.61 & 1.19 & 0.94 & 0.71 \\
\midrule
\multirow{8}{*}{FSTMatching~\cite{Explaining2022Dong_ECCV}} 
& \multirow{4}{*}{500}
& 90\% CI & 47.67--66.44 & 54.04--59.66 & 55.80--58.10 & 56.09--57.72 & 56.39--57.53 & 56.45--57.40 & 56.53--57.28 \\
& & 95\% CI & 45.87--68.24 & 53.49--60.20 & 55.58--58.32 & 55.93--57.88 & 56.29--57.64 & 56.36--57.49 & 56.46--57.35 \\
& & Mean & 57.06 & 56.85 & 56.95 & 56.91 & 56.96 & 56.92 & 56.90 \\
& & Std. & 18.88 & 5.66 & 2.31 & 1.64 & 1.15 & 0.96 & 0.76 \\
\cmidrule{2-10}
& \multirow{4}{*}{3,000}
& 90\% CI & 47.06--66.20 & 54.10--59.72 & 55.84--58.24 & 56.16--57.78 & 56.27--57.49 & 56.45--57.37 & 56.56--57.31 \\
& & 95\% CI & 45.23--68.04 & 53.56--60.26 & 55.61--58.47 & 56.00--57.93 & 56.16--57.61 & 56.36--57.45 & 56.49--57.38 \\
& & Mean & 56.63 & 56.91 & 57.04 & 56.97 & 56.88 & 56.91 & 56.93 \\
& & Std. & 19.29 & 5.66 & 2.42 & 1.63 & 1.23 & 0.92 & 0.75 \\
\midrule
\multirow{8}{*}{MetricLearning~\cite{cao2021metric}} 
& \multirow{4}{*}{500}
& 90\% CI & 64.48--76.77 & 67.55--71.48 & 68.57--70.18 & 69.01--70.08 & 68.99--69.81 & 69.22--69.78 & 69.28--69.69 \\
& & 95\% CI & 63.29--77.95 & 67.18--71.85 & 68.41--70.33 & 68.90--70.19 & 68.91--69.89 & 69.17--69.83 & 69.24--69.73 \\
& & Mean & 70.62 & 69.51 & 69.37 & 69.55 & 69.40 & 69.50 & 69.49 \\
& & Std. & 16.26 & 5.19 & 2.13 & 1.42 & 1.09 & 0.73 & 0.54 \\
\cmidrule{2-10}
& \multirow{4}{*}{3,000}
& 90\% CI & 60.11--78.39 & 66.82--72.23 & 68.35--70.58 & 68.78--70.23 & 68.91--69.96 & 69.10--69.90 & 69.20--69.78 \\
& & 95\% CI & 58.36--80.14 & 66.30--72.74 & 68.14--70.79 & 68.64--70.37 & 68.80--70.07 & 69.03--69.98 & 69.14--69.83 \\
& & Mean & 69.25 & 69.52 & 69.47 & 69.50 & 69.44 & 69.50 & 69.49 \\
& & Std. & 17.56 & 5.20 & 2.14 & 1.40 & 1.02 & 0.76 & 0.56 \\
\midrule
\multirow{8}{*}{LRNet~\cite{sun2021improving}} 
& \multirow{4}{*}{500}
& 90\% CI & 46.90--64.71 & 51.93--57.71 & 53.66--56.02 & 54.26--55.81 & 54.44--55.61 & 54.55--55.43 & 54.69--55.32 \\
& & 95\% CI & 45.19--66.43 & 51.37--58.27 & 53.43--56.25 & 54.11--55.96 & 54.32--55.73 & 54.46--55.52 & 54.63--55.38 \\
& & Mean & 55.81 & 54.82 & 54.84 & 55.04 & 55.03 & 54.99 & 55.01 \\
& & Std. & 17.93 & 5.83 & 2.38 & 1.57 & 1.19 & 0.89 & 0.64 \\
\cmidrule{2-10}
& \multirow{4}{*}{3,000}
& 90\% CI & 45.24--63.92 & 52.49--58.15 & 53.84--56.20 & 54.25--55.76 & 54.47--55.60 & 54.57--55.46 & 54.68--55.33 \\
& & 95\% CI & 43.45--65.71 & 51.95--58.69 & 53.62--56.43 & 54.10--55.90 & 54.36--55.71 & 54.48--55.55 & 54.62--55.40 \\
& & Mean & 54.58 & 55.32 & 55.02 & 55.00 & 55.03 & 55.02 & 55.01 \\
& & Std. & 18.82 & 5.70 & 2.37 & 1.52 & 1.14 & 0.90 & 0.66 \\
\bottomrule
\end{tabular}
}
\label{tab6}
\end{center}
\end{table}

\begin{table}
\begin{center}
\caption{AUC score statistics (\%) with imbalanced sampling for 500 and 3,000 trials. }
\resizebox{\textwidth}{!}{
\begin{tabular}{lccccccccc}
\toprule
\multirow{2}{*}{Model Names} & \multirow{2}{*}{Num Trials} & \multirow{2}{*}{Statistics} & \multicolumn{7}{c}{Sample Sizes} \\
\cmidrule{4-10}
& & & 10 & 100 & 500 & 1,000 & 1,500 & 2,000 & 2,500 \\
\midrule
\multirow{8}{*}{Xception~\cite{Chollet2017Xception}} 
& \multirow{4}{*}{500}
& 90\% CI & -- & 74.64--78.75 & 75.80--77.60 & 75.98--77.21 & 76.12--77.19 & 76.17--77.06 & 76.31--77.15 \\
& & 95\% CI & -- & 74.25--79.15 & 75.62--77.77 & 75.86--77.33 & 76.02--77.29 & 76.09--77.15 & 76.23--77.23 \\
& & Mean & -- & 76.70 & 76.70 & 76.60 & 76.65 & 76.62 & 76.73 \\
& & Std. & -- & 5.43 & 2.38 & 1.62 & 1.41 & 1.18 & 1.11 \\
\cmidrule{2-10}
& \multirow{4}{*}{3,000}
& 90\% CI & -- & 74.78--78.80 & 75.79--77.56 & 76.02--77.29 & 76.13--77.17 & 76.20--77.11 & 76.24--77.05 \\
& & 95\% CI & -- & 74.39--79.19 & 75.61--77.73 & 75.89--77.41 & 76.03--77.27 & 76.11--77.20 & 76.17--77.13 \\
& & Mean & -- & 76.79 & 76.67 & 76.65 & 76.65 & 76.66 & 76.65 \\
& & Std. & -- & 5.33 & 2.35 & 1.69 & 1.38 & 1.21 & 1.07 \\
\midrule
\multirow{8}{*}{MAT~\cite{Zhao2021Multi-Attentional}} 
& \multirow{4}{*}{500}
& 90\% CI & -- & 74.03--78.10 & 74.64--76.44 & 75.09--76.35 & 75.09--76.11 & 75.24--76.14 & 75.23--76.02 \\
& & 95\% CI & -- & 73.64--78.49 & 74.47--76.62 & 74.97--76.47 & 74.99--76.21 & 75.15--76.23 & 75.16--76.09 \\
& & Mean & -- & 76.07 & 75.54 & 75.72 & 75.60 & 75.69 & 75.63 \\
& & Std. & -- & 5.24 & 2.32 & 1.62 & 1.32 & 1.16 & 1.01 \\
\cmidrule{2-10}
& \multirow{4}{*}{3,000}
& 90\% CI & -- & 73.65--77.52 & 74.71--76.41 & 75.07--76.28 & 75.16--76.16 & 75.25--76.10 & 75.25--76.03 \\
& & 95\% CI & -- & 73.28--77.89 & 74.55--76.58 & 74.96--76.40 & 75.07--76.26 & 75.17--76.18 & 75.18--76.10 \\
& & Mean & -- & 75.58 & 75.56 & 75.68 & 75.66 & 75.68 & 75.64 \\
& & Std. & -- & 5.13 & 2.26 & 1.60 & 1.32 & 1.13 & 1.03 \\
\midrule
\multirow{8}{*}{RECCE~\cite{RECCE2022Cao}} 
& \multirow{4}{*}{500}
& 90\% CI & -- & 64.69--69.07 & 65.67--67.67 & 66.32--67.58 & 66.27--67.36 & 66.35--67.31 & 66.37--67.26 \\
& & 95\% CI & -- & 64.27--69.50 & 65.48--67.86 & 66.20--67.70 & 66.17--67.46 & 66.26--67.40 & 66.29--67.34 \\
& & Mean & -- & 66.88 & 66.67 & 66.95 & 66.81 & 66.83 & 66.82 \\
& & Std. & -- & 5.80 & 2.64 & 1.66 & 1.44 & 1.27 & 1.17 \\
\cmidrule{2-10}
& \multirow{4}{*}{3,000}
& 90\% CI & -- & 64.67--68.97 & 65.85--67.78 & 66.17--67.56 & 66.26--67.38 & 66.34--67.28 & 66.38--67.25 \\
& & 95\% CI & -- & 64.26--69.38 & 65.66--67.96 & 66.04--67.69 & 66.16--67.48 & 66.25--67.37 & 66.30--67.33 \\
& & Mean & -- & 66.82 & 66.81 & 66.87 & 66.82 & 66.81 & 66.81 \\
& & Std. & -- & 5.69 & 2.56 & 1.84 & 1.47 & 1.25 & 1.15 \\
\midrule
\multirow{8}{*}{Stage5~\cite{yang2021beyond_ijcai}}
& \multirow{4}{*}{500}
& 90\% CI & -- & 48.58--53.40 & 49.39--51.49 & 49.74--51.14 & 50.00--51.09 & 50.08--50.98 & 50.13--50.83 \\
& & 95\% CI & -- & 48.12--53.86 & 49.19--51.69 & 49.61--51.27 & 49.89--51.19 & 50.00--51.07 & 50.07--50.90 \\
& & Mean & -- & 50.99 & 50.44 & 50.44 & 50.54 & 50.53 & 50.48 \\
& & Std. & -- & 6.37 & 2.78 & 1.85 & 1.44 & 1.18 & 0.92 \\
\cmidrule{2-10}
& \multirow{4}{*}{3,000}
& 90\% CI & -- & 48.10--52.96 & 49.49--51.58 & 49.78--51.18 & 49.97--51.02 & 50.06--50.91 & 50.11--50.81 \\
& & 95\% CI & -- & 47.64--53.42 & 49.29--51.78 & 49.65--51.31 & 49.87--51.12 & 49.98--50.99 & 50.04--50.88 \\
& & Mean & -- & 50.53 & 50.53 & 50.48 & 50.49 & 50.49 & 50.46 \\
& & Std. & -- & 6.44 & 2.77 & 1.85 & 1.39 & 1.13 & 0.93 \\
\midrule
\multirow{8}{*}{FSTMatching~\cite{Explaining2022Dong_ECCV}} 
& \multirow{4}{*}{500}
& 90\% CI & -- & 54.79--59.43 & 55.72--57.70 & 56.33--57.62 & 56.45--57.46 & 56.53--57.33 & 56.64--57.30 \\
& & 95\% CI & -- & 54.34--59.87 & 55.53--57.89 & 56.20--57.74 & 56.35--57.56 & 56.45--57.40 & 56.57--57.37 \\
& & Mean & -- & 57.11 & 56.71 & 56.97 & 56.96 & 56.93 & 56.97 \\
& & Std. & -- & 6.14 & 2.62 & 1.71 & 1.34 & 1.06 & 0.88 \\
\cmidrule{2-10}
& \multirow{4}{*}{3,000}
& 90\% CI & -- & 54.53--59.18 & 56.08--58.08 & 56.21--57.55 & 56.42--57.44 & 56.52--57.36 & 56.60--57.27 \\
& & 95\% CI & -- & 54.08--59.63 & 55.89--58.27 & 56.09--57.68 & 56.33--57.53 & 56.44--57.44 & 56.54--57.34 \\
& & Mean & -- & 56.86 & 57.08 & 56.88 & 56.93 & 56.94 & 56.94 \\
& & Std. & -- & 6.16 & 2.64 & 1.77 & 1.34 & 1.11 & 0.89 \\
\midrule
\multirow{8}{*}{MetricLearning~\cite{cao2021metric}} 
& \multirow{4}{*}{500}
& 90\% CI & -- & 66.94--72.04 & 68.33--70.44 & 68.75--70.16 & 69.06--70.11 & 69.06--69.91 & 69.02--69.76 \\
& & 95\% CI & -- & 66.45--72.53 & 68.13--70.64 & 68.62--70.29 & 68.96--70.21 & 68.98--69.99 & 68.95--69.83 \\
& & Mean & -- & 69.49 & 69.38 & 69.46 & 69.59 & 69.48 & 69.39 \\
& & Std. & -- & 6.74 & 2.78 & 1.85 & 1.38 & 1.12 & 0.97 \\
\cmidrule{2-10}
& \multirow{4}{*}{3,000}
& 90\% CI & -- & 67.07--72.09 & 68.45--70.49 & 68.72--70.12 & 68.93--70.01 & 69.07--69.91 & 69.16--69.87 \\
& & 95\% CI & -- & 66.59--72.58 & 68.25--70.69 & 68.58--70.25 & 68.83--70.12 & 68.99--69.99 & 69.09--69.94 \\
& & Mean & -- & 69.58 & 69.47 & 69.42 & 69.47 & 69.49 & 69.51 \\
& & Std. & -- & 6.65 & 2.71 & 1.86 & 1.43 & 1.12 & 0.94 \\
\midrule
\multirow{8}{*}{LRNet~\cite{sun2021improving}} 
& \multirow{4}{*}{500}
& 90\% CI & -- & 52.42--57.52 & 54.03--56.28 & 54.33--55.77 & 54.56--55.71 & 54.48--55.37 & 54.59--55.42 \\
& & 95\% CI & -- & 51.93--58.01 & 53.81--56.50 & 54.19--55.91 & 54.45--55.81 & 54.40--55.46 & 54.51--55.50 \\
& & Mean & -- & 54.97 & 55.15 & 55.05 & 55.13 & 54.93 & 55.01 \\
& & Std. & -- & 6.56 & 2.90 & 1.85 & 1.47 & 1.14 & 1.06 \\
\cmidrule{2-10}
& \multirow{4}{*}{3,000}
& 90\% CI & -- & 52.53--57.66 & 53.86--56.03 & 54.32--55.78 & 54.43--55.56 & 54.56--55.46 & 54.64--55.39 \\
& & 95\% CI & -- & 52.04--58.16 & 53.65--56.24 & 54.17--55.92 & 54.32--55.67 & 54.47--55.54 & 54.57--55.46 \\
& & Mean & -- & 55.10 & 54.95 & 55.05 & 55.00 & 55.01 & 55.01 \\
& & Std. & -- & 6.80 & 2.88 & 1.95 & 1.50 & 1.19 & 0.99 \\
\bottomrule
\end{tabular}
}
\label{tab7}
\end{center}
\end{table}

\subsection{Model Reliability Experiment Results}
\label{reliability_results}

The model reliability evaluation is conducted on a sampling frame with 5,368 videos composed of the testing set of each benchmark dataset, and the detailed data quantity is listed in Table~\ref{tab3}. Following the workflow of Algorithm~\ref{alg1}, we proposed reliability analyses on the well-trained Deepfake detection approaches. We computed the 90\% and 95\% confidence intervals in experiments. Specifically, two values of $t$, 500 and 3,000, are chosen to ensure a sufficient number of trials when locating the sample means. Various sample sizes $s \in \{\textrm{10, 100, 500, 1,000, 1,500, 2,000, 2,500}\}$ are selected to find the settled confidence intervals. Besides, both sampling options, balanced and imbalanced sampling, are executed in experiments. Detailed results are listed in Tables~\ref{tab4} to~\ref{tab7} following the Cartesian product settings of balancing options $O=\{\textrm{True, False}\}$, the number of trials $T=\{\textrm{500, 3,000}\}$, and evaluation metrics $E=\{\textrm{ACC, AUC}\}$.

It can be easily observed that for all models in the tables, the mean values $\Bar{x}$ gradually settle as sample sizes become larger and the standard deviation (Std.) values $\sigma$ consistently decrease concurrently. Similarly, the 90\% and 95\% confidence intervals are progressively straitened and settled around the mean values. Moreover, statistical results with 3,000 sampling trials converge faster and are more stable than those with 500 trials as sample size increases for all experiments, and both trial numbers lead to similar final values once settled. 

Regarding the balanced sampling setting, the results generally match the model detection performance in Table~\ref{tab2}. On the other hand, since the constructed sampling frame is imbalanced for real and fake faces, an imbalanced sampling option may lead to more fake data than the real ones in the sample set. Consequently, for models that have shown relatively poor performance when evaluated on fake testing sets (DF1.0 and FaceShifter) solely, the mean values and confidence intervals of the accuracy are located at lower levels. Meanwhile, models that have achieved promising performance on merely the fake testing sets have led to higher accuracy values for mean and confidence intervals. The AUC score metric, as mentioned in Section~\ref{experiment_settings}, is impervious regarding the imbalanced dataset. Therefore, mean values and confidence intervals under balanced and imbalanced sampling options are generally identical.

With a closer look at the tables, the leading approaches, Xception~\cite{Chollet2017Xception} and MAT~\cite{Zhao2021Multi-Attentional}, both have achieved mean accuracies above 68\% and confidence intervals around 69\% with respect to the balanced sampling option. All models regarding the interpretability and robustness topics have derived accuracies and confidence intervals around 50\%. Stage5~\cite{yang2021beyond_ijcai} and MetricLearning~\cite{cao2021metric} convey results with all values being 50\% since the former recognizes all candidate samples as real and the latter classifies all as fake when checking the predicted labels accordingly. While Stage5~\cite{yang2021beyond_ijcai} may only be interpretable when facing GAN-based synthetic contents depending on its model design, MetricLearning~\cite{cao2021metric} relies on a fixed threshold of 5 upon the output value without softmax or sigmoid activation. Since a perfect threshold value may vary depending on the testing dataset, the fixed threshold may be the main fact that causes the mistake. By looking at the AUC scores in Table~\ref{tab6} and Table~\ref{tab7}, MetricLeaning~\cite{cao2021metric} actually has displayed a reasonable ability in distinguishing real and fake with a mean value of 69.49\% AUC score. As for FSTMatching~\cite{Explaining2022Dong_ECCV} and LRNet~\cite{sun2021improving}, the experimental results are generally consistent with the ones in Table~\ref{tab2}.

As for experiments under the imbalanced sampling setting, besides the foreseeably high and low performance by MetricLearning~\cite{cao2021metric} and Stage5~\cite{yang2021beyond_ijcai} as discussed, Xception~\cite{Chollet2017Xception} wins with the highest mean accuracy of 66.23\% due to its stable performance in detecting both real and fake faces. Regarding the AUC scores, ignoring the imperceptible differences and taking a look at Table~\ref{tab6}, 76.64\% is derived by Xception because of its ability to separate real and fake samples at a certain threshold even though performing relatively unsatisfactory with the threshold value of 0.5 regarding the softmax output scores. Besides, MAT~\cite{Zhao2021Multi-Attentional} is the only other model that has achieved above 70\% AUC score. In the remaining approaches, RECCE has reached above 65\% AUC scores, while the other two methods have performed relatively unsatisfactory in comparison.

It is also worth noting that despite the imbalanced sampling setting does not affect the final values of mean and confidence intervals, it may cause the AUC score incomputable for a tiny sample size. In particular, as shown in Table~\ref{tab7}, there is a high possibility to randomly draw 10 samples that belong to the same category, which then leads to an incomputable AUC score since the sample set lacks data from the other category. Meanwhile, it is unlikely to randomly draw 100 or even more samples that are of the same category even though there is a possibility theoretically.

\begin{figure}
\centering
\includegraphics[width=\textwidth]{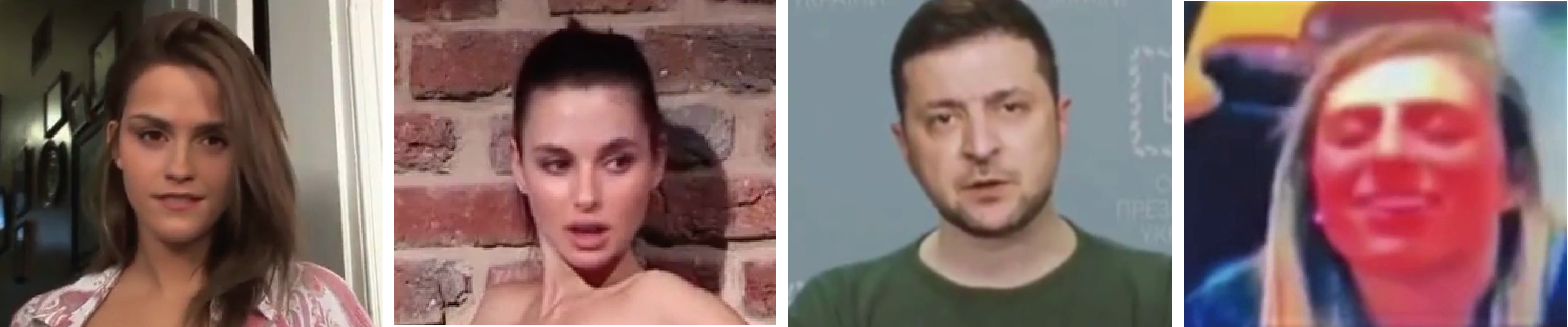}
\caption{Screenshots of the four real-life Deepfake videos for Deepfake detection in the case study. The videos are hyper-realistic with different resolutions and no obvious artifacts can be observed visually by human eyes. }
\label{fig1}
\end{figure}

\section{Real-life Case Study}
\label{case_study}

In this section, we made use of the experiment results of the model reliability study. Experiments are conducted to analyze the reliabilities of the existing Deepfake detection approaches when applied to real-life cases. Specifically, four famous Deepfake cases that have jeopardized individuals and society from 2018 to 2022 are considered in this case study (Fig.~\ref{fig1}). In 2018, when the technique of Deepfake had just been released shortly, the well-known actress Emma Watson who performed in the Harry Potter movie series was face-swapped onto porn videos~\cite{Kelion2018Emma}. In the same year, another famous actress Natalie Portman encountered a similar fake scandal because of Deepfake~\cite{Lee2018Portman}. The porn videos were widely spread at the time and had gravely influenced their reputations because the term `Deepfake' was unfamiliar to the public and people were easily tricked and believed the videos to be genuine upon their first appearances. Later in March 2021, a Bucks County mom was accused of creating Deepfake videos of the underage girls on her daughter’s cheerleader team and threatening them to quit the team~\cite{Rudy2021Cheerleader}. The videos exhibited the girls that were naked, drinking alcohol, or vaping, and are accused to be fake. Nevertheless, two months later in May, the prosecutors admitted that they could not prove the fake-video claims without reliable tools and evidence~\cite{Harwell2021CheerProve}. This is a representative case that confirms that anyone can become a victim of Deepfake nowadays. One of the most famous and most recent Deepfake events, the fake Zelensky video, had caused a short panic in the country during the Russia-Ukraine war~\cite{Miller2022Zelensky}. In that fake video, a synthetic president Zelensky was telling Ukrainians to put down their weapons and give up resistance. 

\subsection{Detection Results}
\label{case_study_results}

We obtained the available video clips of the four Deepfake cases from the internet and performed Deepfake detection using each of the well-trained models. Specifically, for video clips with sufficient numbers of image frames that contain faces, we randomly extracted 100 frames for face cropping when using frame-level detectors. As for the cheerleader case, as the sensitive contents are omitted, we were only able to acquire a total of 75 faces from the news clip. For frame-level detectors, the numbers of faces classified as real or fake by each model for each video are exhibited in Table~\ref{tab8}, and except MetricLearning~\cite{cao2021metric} that uses a fixed threshold, the softmax scores for each video are averaged to obtain a single score that indicates the model determined authenticity. In particular, except for MetricLearning, the fake scores lie in the range of [0, 1] where 0 refers to real and 1 refers to fake. For video-level detectors, the ultimate results are straightforwardly demonstrated in the table. As a result, given the fact that all four videos are known to be fake, about half of the selected models have made the correct classifications regarding both the number of fake faces and the average softmax score. Besides, we provided reliably quantified 95\% confidence intervals regarding the models' detection accuracy for reference when utilizing the results. 

Despite achieving high statistical values regarding the accuracy and AUC score metrics in early experiments, Xception has failed to classify all four fake videos such that most faces are classified as real and all average softmax scores are below 0.43. Besides, the fake Emma Watson and Natalie Portman videos have also tricked the MAT model such that roughly two-thirds of the faces are classified as real and the final scores are below 0.4. Stage5 and FSTMatching, which proved to be relatively unsatisfactory in early experiments and discussions, have both failed to detect all four fake videos. MetricLearning and LRNet, although performing poorly on lab-controlled datasets, have shown robust detection ability especially because the videos circulated and downloaded from the internet have suffered incrementally heavy compressions. The RECCE model, as a result, turns out to be the winner with the highest fake scores and the number of correctly classified faces when facing real-life Deepfake suspects.

\begin{table}
\begin{center}
\caption{Deepfake detection results on real-life cases by the well-trained models. 95\% confidence intervals following the balanced sampling option for accuracies are listed along the models. ($\dag$: threshold fixed as 5 where greater values refers to fake; $\ddag$: video-level detector with no intermediate frame-level result.)}
\resizebox{\textwidth}{!}{
\begin{tabular}{lccccccccc}
\toprule\noalign{\smallskip}
\multirow{3}{*}{Model} & \multicolumn{8}{c}{Real-life Videos} & \multirowcell{3}{\noalign{\smallskip}\noalign{\smallskip}\noalign{\smallskip}95\% CI\\\noalign{\smallskip}Balanced Sampling\\\noalign{\smallskip}ACC (\%)}  \\
\noalign{\smallskip}\cmidrule{2-9}\noalign{\smallskip}
& \multicolumn{2}{c}{Watson (2018)} & \multicolumn{2}{c}{Portman (2018)} & \multicolumn{2}{c}{Cheerleader (2021)} & \multicolumn{2}{c}{Zelensky (2022)} &\\
\noalign{\smallskip}\cmidrule{2-9}\noalign{\smallskip}
& \# Real / Fake & Fake Score & \# Real / Fake & Fake Score & \# Real / Fake & Fake Score & \# Real / Fake & Fake Score & \\
\noalign{\smallskip}\midrule\noalign{\smallskip}
Xception~\cite{Chollet2017Xception} & 78 / 22 & 0.290 (Real) & 59 / 41 & 0.426 (Real) & 68 / 7 & 0.401 (Real) & 69 / 31 & 0.328 (Real) & 68.42--69.34\\
MAT~\cite{Zhao2021Multi-Attentional} & 66 / 34 & 0.365 (Real) & 67 / 33 & 0.378 (Real) & 3 / 72 & 0.951 (Fake) & 34 / 66 & 0.630 (Fake) & 68.41--69.37\\
RECCE~\cite{RECCE2022Cao} & 10 / 90 & 0.779 (Fake) & 8 / 92 & 0.808 (Fake) & 0 / 75 & 0.902 (Fake) & 15 / 85 & 0.723 (Fake) & 60.29--61.30 \\
Stage5~\cite{yang2021beyond_ijcai} & 100 / 0 & 0.000 (Real) & 100 / 0 & 0.000 (Real) & 75 / 0 & 0.000 (Real) & 100 / 0 & 0.000 (Real) & 50.00--50.00 \\
FSTMatching~\cite{Explaining2022Dong_ECCV} & 94 / 6 & 0.259 (Real) & 89 / 11 & 0.261 (Real) & 75 / 0 & 0.113 (Real) & 95 / 5 & 0.245 (Real) & 51.99--52.76\\
MetricLearning$^\dag$~\cite{cao2021metric} & 35 / 65 & 11.698 (Fake) & 27 / 73 & 11.996 (Fake) & 23 / 52 & 11.928 (Fake) & 27 / 73 & 11.951 (Fake) & 50.00--50.00\\
LRNet$^\ddag$~\cite{sun2021improving} & -- & 0.655 (Fake) & -- & 0.667 (Fake) & -- & 0.667 (Fake) & -- & 0.538 (Fake) & 51.55--52.43 \\
\noalign{\smallskip}\bottomrule
\end{tabular}
}
\label{tab8}
\end{center}
\end{table}

\subsection{Deepfake Intelligence}
Considering real-life usages, accessible Deepfake detection models such as the ones being discussed in this survey can be gathered to an online platform and to provide real-time detection services. To go even further, similar to VirusTotal~\cite{virustotal}, a famous virus and malware intelligence platform that reports malicious threat intelligence, a fake video intelligence web portal can be established regarding the Deepfake detection research domain by endlessly integrating the detection models and real-life Deepfake intelligence beyond detection.

\section{Discussion}
\label{discussion}

Our study provides a scientific workflow to prove the reliability of the detection models when applied to real-life cases. Unlike research that simply enumerates detection performance on each benchmark dataset, the reliability study scheme derives statistical claims regarding the detectors on arbitrary candidate Deepfake suspects with the help of confidence intervals. Particularly, the interval values are reliable based on sufficient trials from a sampling frame that ideally imitates real-world Deepfake distribution according to CLT. Considering that the prosecutors are unable to prove the fake-video claim due to the challenge brought to the video evidence authentication standard by Deepfake, the experiment results in Section~\ref{reliability_study} have solved the problem favorably. Specifically, the model reliability study scheme can be qualified by expert witnesses for the validity of the detection models based on the expert's testimony following corresponding rules, and thus, the reliable statistical metrics regarding the detection performance may assist the video evidence for criminal investigation cases. The accuracies are picked to assist the claims while the AUC scores are more helpful at the research level. 

As a result, a reliable justification can be claimed based on values in Table~\ref{tab4} and Table~\ref{tab5} with the help of the confidence intervals and mean values once a sampling option is chosen. For example, suppose the RECCE model is used to help justify the authenticity of the cheerleader video of the `Deepfake mom' case following the balanced sampling results, a claim can be made such that the video is fake with accuracies lying in the range of [60.37\%, 61.22\%] and [60.29\%, 61.30\%] with 90\% and 95\% confidence levels, respectively. In other words, we are 90\% and 95\% confident to declare that the video is fake with accuracies in the range of [60.37\%, 61.22\%] and [60.29\%, 61.30\%], respectively. If the imbalanced sampling option is trusted, a similar claim can be concluded in the range of [62.36\%, 63.66\%] and [62.24\%, 63.79\%] accuracies for the 90\% and 95\% confidence levels, respectively. 

% Admittedly, the classification accuracies with the 95\% confidence interval as demonstrated in the model reliability study are not satisfied to act as evidence for criminal cases at the court. However, as the defendants and the victims of Deepfake cases are usually individuals, the civil case~\cite{civilcase} is more applicable to the Deepfake prosecution. According to the words of Lord Nicholas in Re H (Minors) [1996] AC 563~\cite{ac563}, based on the balance of probability standard, `a court is satisfied an event occurred if the court considers that, on the evidence, the occurrence of the event was more likely than not,' in other words, larger than the 50\% likelihood. Therefore, the accuracy values above 50\% can be sufficient to assist or act as evidence to prove the falsification of the cheerleader video under the qualification of expert testimony following proper legislation.

In real life, since the authenticity of the candidate suspect is normally unknown, based on the experiment results in this study, the dominant MAT model is likely to be adopted for Deepfake detection and the following conclusion can be provided following the balanced sampling setting: we are 95\% confident to classify the video as real (or fake) with an accuracy between 68.41\% and 69.37\%. If the imbalanced sampling setting is trusted, the winning Xception model can be employed to offer the justification such that the video is real (or fake) with an accuracy between 65.37\% and 67.07\% at the 95\% confidence level.

Meanwhile, several findings can be concluded based on the results in Table~\ref{tab2} and Tables~\ref{tab4} to~\ref{tab8}. Firstly, trade-offs are objectively unavoidable when attempting each of the three challenges. Specifically, regarding Table~\ref{tab2}, models with promising transferability on lab-controlled benchmark datasets might lack interpretability on their performance and robustness in sophisticated real-life scenarios. While successfully explaining the detection decision with pellucid evidence and common sense or smoothly resolving challenges in specific real-life conditions with robustness as reported in the published papers, there usually remains limited computational power and model ability to enrich the transferability and detection accuracy on unseen data. Secondly, although the detection performance in Sections~\ref{detection_results} and~\ref{reliability_results} is unremarkable compared to other approaches, the RECCE model appears to derive the best detection results on real-life Deepfake videos given the fact that we are aware that they are all fake, while on the contrary, the winning MAT and Xception models from early experiments both have failed in classifying multiple real-life fake videos. In other words, based on the experiment results, at the current research stage, a detection model showing promising performance on the benchmark datasets does not necessarily perform well on real-life Deepfake materials. This may be caused by the potential adversarial attacks such that the facial manipulation technique of the fake materials can easily fool models that are mainly based on certain feature extraction perspectives or techniques, and the detectors, therefore, need to be improved to better cooperate with the reliability study scheme. Lastly, for models that have a large gap between accuracy and AUC score values, it may be meaningful to locate a classification threshold other than 0.5 in order to achieve satisfactory detection accuracy since their high AUC scores have highlighted the ability to separate real and fake data.

\section{Conclusion}
\label{conclusion}
This paper provides a thorough survey of reliability-oriented Deepfake detection approaches by defining the three challenges of Deepfake detection research: transferability, interpretability, and robustness. While the early methods mainly solve puzzles on seen data, improvements by persistent attempts have gradually shown promising up-to-date detection performance on unseen benchmark datasets. Considering the lack of usage for the well-trained detection models to benefit real life and even specifically for criminal investigation, this paper conducts a comprehensive survey regarding model reliability by introducing an unprecedented model reliability study scheme that bridges the research gap and provides a reliable path for applying on-the-shelf detection models to assist prosecutions in court for Deepfake related cases. A barely discussed standardized data preparation workflow is simultaneously designed and presented for the reference of both starters and veterans in the domain. The reliable accuracies of the detection models derived by random sampling at the 90\% and 95\% confidence levels are informative and may be adopted as or to assist forensic evidence in court for Deepfake-related cases under the qualification of expert testimony. 

Based on the informative findings in validating the detection models, potential future research trends are worth discussing. Although a Deepfake detection model can be verified the reliability for real-life usages via the presented reliability study scheme in this paper, an ideal model that simultaneously solves transferability, interpretability, and robustness challenges is not yet accomplished in the current research domain. Consequently, obvious trade-offs have been observed when resolving each of the three challenges. Therefore, considering that the ground-truth attacks and labels of current and future Deepfake contents will not be visible to victims in Deepfake-related cases, researchers may continuously advance the detection model performance regarding each of the three challenges, but more importantly, a model that satisfies all three goals at the same time is urgently desired. At the same time, based on the model reliability study scheme that is first put forward in this study, subsequent improvements and discussions can also be conducted to achieve the general reliability goal progressively. For instance, tracing original sources of synthetic contents and recovering synthetic operation sequences are worth exploiting in future work to further enhance the reliability of a falsification. Moreover, although videos in this study are either real or fake as a whole, as the real-world Deepfake materials become complex, videos with only partial image frames being fake can lead to further potential risks, and the corresponding benchmark datasets and detectors are also desired. 

% \section{Acknowledgements}

% This work is supported by the National Natural Science Foundation of China (Grant No. U22A2030), the National Key R\&D Program of China (Grant No. 2022YFB3103500), and Hunan Provincial Funds for Distinguished Young Scholars (Grant No. 2024JJ2025).

\bibliographystyle{ACM-Reference-Format}
\bibliography{my_bib}


\begin{thebibliography}{198}

%%% the Web address.

\ifx \showCODEN    \undefined \def \showCODEN     #1{\unskip}     \fi
\ifx \showDOI      \undefined \def \showDOI       #1{#1}\fi
\ifx \showISBNx    \undefined \def \showISBNx     #1{\unskip}     \fi
\ifx \showISBNxiii \undefined \def \showISBNxiii  #1{\unskip}     \fi
\ifx \showISSN     \undefined \def \showISSN      #1{\unskip}     \fi
\ifx \showLCCN     \undefined \def \showLCCN      #1{\unskip}     \fi
\ifx \shownote     \undefined \def \shownote      #1{#1}          \fi
\ifx \showarticletitle \undefined \def \showarticletitle #1{#1}   \fi
\ifx \showURL      \undefined \def \showURL       {\relax}        \fi
% invisible to TeX
\providecommand\bibfield[2]{#2}
\providecommand\bibinfo[2]{#2}
\providecommand\natexlab[1]{#1}
\providecommand\showeprint[2][]{arXiv:#2}

\bibitem[Administration(2022)]%
        {FDA2022antigen}
\bibfield{author}{\bibinfo{person}{U.S Food \&~Drug Administration}.} \bibinfo{year}{2022}\natexlab{}.
\newblock \bibinfo{title}{COVID-19 Antigen Home Test Package Insert for Healthcare Providers}.
\newblock \bibinfo{howpublished}{\url{https://www.fda.gov/media/152698/download}}.
\newblock
\newblock
\shownote{Accessed: 2022-09-09}.


\bibitem[Afchar et~al\mbox{.}(2018)]%
        {Afchar2018MesoNet}
\bibfield{author}{\bibinfo{person}{Darius Afchar}, \bibinfo{person}{Vincent Nozick}, \bibinfo{person}{Junichi Yamagishi}, {and} \bibinfo{person}{Isao Echizen}.} \bibinfo{year}{2018}\natexlab{}.
\newblock \showarticletitle{MesoNet: a Compact Facial Video Forgery Detection Network}.
\newblock \bibinfo{journal}{\emph{2018 IEEE International Workshop on Information Forensics and Security (WIFS)}} (\bibinfo{year}{2018}), \bibinfo{pages}{1--7}.
\newblock


\bibitem[Agarwal et~al\mbox{.}(2020)]%
        {Agarwal2020Detecting_WIFS}
\bibfield{author}{\bibinfo{person}{Shruti Agarwal}, \bibinfo{person}{Hany Farid}, \bibinfo{person}{Tarek El-Gaaly}, {and} \bibinfo{person}{Ser-Nam Lim}.} \bibinfo{year}{2020}\natexlab{}.
\newblock \showarticletitle{Detecting Deep-Fake Videos from Appearance and Behavior}. In \bibinfo{booktitle}{\emph{2020 IEEE International Workshop on Information Forensics and Security (WIFS)}}. \bibinfo{pages}{1--6}.
\newblock
\urldef\tempurl%
\url{https://doi.org/10.1109/WIFS49906.2020.9360904}
\showDOI{\tempurl}


\bibitem[Amerini et~al\mbox{.}(2019)]%
        {Amerini2019Deepfake}
\bibfield{author}{\bibinfo{person}{Irene Amerini}, \bibinfo{person}{Leonardo Galteri}, \bibinfo{person}{Roberto Caldelli}, {and} \bibinfo{person}{Alberto Del~Bimbo}.} \bibinfo{year}{2019}\natexlab{}.
\newblock \showarticletitle{Deepfake Video Detection through Optical Flow Based CNN}. In \bibinfo{booktitle}{\emph{2019 IEEE/CVF International Conference on Computer Vision Workshop (ICCVW)}}. \bibinfo{publisher}{IEEE}, \bibinfo{pages}{1205--1207}.
\newblock


\bibitem[Aouf(2019)]%
        {Rima2019Museum}
\bibfield{author}{\bibinfo{person}{Rima~Sabina Aouf}.} \bibinfo{year}{2019}\natexlab{}.
\newblock \bibinfo{title}{Museum creates deepfake Salvador Dalí to greet visitors.}
\newblock \bibinfo{howpublished}{\url{https://www.dezeen.com/2019/05/24/salvador-dali-deepfake-dali-musuem-florida/}}.
\newblock
\newblock
\shownote{Accessed: 2022-09-08}.


\bibitem[Bai et~al\mbox{.}(2020)]%
        {bai2020fake}
\bibfield{author}{\bibinfo{person}{Yong Bai}, \bibinfo{person}{Yuanfang Guo}, \bibinfo{person}{Jinjie Wei}, \bibinfo{person}{Lin Lu}, \bibinfo{person}{Rui Wang}, {and} \bibinfo{person}{Yunhong Wang}.} \bibinfo{year}{2020}\natexlab{}.
\newblock \showarticletitle{Fake Generated Painting Detection Via Frequency Analysis}. In \bibinfo{booktitle}{\emph{2020 IEEE International Conference on Image Processing (ICIP)}}. \bibinfo{pages}{1256--1260}.
\newblock
\urldef\tempurl%
\url{https://doi.org/10.1109/ICIP40778.2020.9190892}
\showDOI{\tempurl}


\bibitem[Baldassarre et~al\mbox{.}(2022)]%
        {Quantitative2022Baldassarre_BMVC}
\bibfield{author}{\bibinfo{person}{Federico Baldassarre}, \bibinfo{person}{Quentin Debard}, \bibinfo{person}{Gonzalo~Fiz Pontiveros}, {and} \bibinfo{person}{Tri~Kurniawan Wijaya}.} \bibinfo{year}{2022}\natexlab{}.
\newblock \showarticletitle{Quantitative Metrics for Evaluating Explanations of Video DeepFake Detectors}. In \bibinfo{booktitle}{\emph{33rd British Machine Vision Conference 2022, {BMVC} 2022, London, UK, November 21-24, 2022}}. \bibinfo{publisher}{{BMVA} Press}.
\newblock
\urldef\tempurl%
\url{https://bmvc2022.mpi-inf.mpg.de/0972.pdf}
\showURL{%
\tempurl}


\bibitem[Binh and Woo(2022)]%
        {Binh2022ADD}
\bibfield{author}{\bibinfo{person}{Le~Minh Binh} {and} \bibinfo{person}{Simon Woo}.} \bibinfo{year}{2022}\natexlab{}.
\newblock \showarticletitle{ADD: Frequency Attention and Multi-View Based Knowledge Distillation to Detect Low-Quality Compressed Deepfake Images}.
\newblock \bibinfo{journal}{\emph{Proceedings of the AAAI Conference on Artificial Intelligence}} \bibinfo{volume}{36}, \bibinfo{number}{1} (\bibinfo{date}{Jun.} \bibinfo{year}{2022}), \bibinfo{pages}{122--130}.
\newblock
\urldef\tempurl%
\url{https://doi.org/10.1609/aaai.v36i1.19886}
\showDOI{\tempurl}


\bibitem[Bonettini et~al\mbox{.}(2021)]%
        {Bonettini2021ensemble}
\bibfield{author}{\bibinfo{person}{Nicolò Bonettini}, \bibinfo{person}{Edoardo~Daniele Cannas}, \bibinfo{person}{Sara Mandelli}, \bibinfo{person}{Luca Bondi}, \bibinfo{person}{Paolo Bestagini}, {and} \bibinfo{person}{Stefano Tubaro}.} \bibinfo{year}{2021}\natexlab{}.
\newblock \showarticletitle{Video Face Manipulation Detection Through Ensemble of CNNs}. In \bibinfo{booktitle}{\emph{2020 25th International Conference on Pattern Recognition (ICPR)}}. \bibinfo{pages}{5012--5019}.
\newblock
\urldef\tempurl%
\url{https://doi.org/10.1109/ICPR48806.2021.9412711}
\showDOI{\tempurl}


\bibitem[Brock et~al\mbox{.}(2019)]%
        {brock2018large}
\bibfield{author}{\bibinfo{person}{Andrew Brock}, \bibinfo{person}{Jeff Donahue}, {and} \bibinfo{person}{Karen Simonyan}.} \bibinfo{year}{2019}\natexlab{}.
\newblock \showarticletitle{Large Scale {GAN} Training for High Fidelity Natural Image Synthesis}. In \bibinfo{booktitle}{\emph{International Conference on Learning Representations}}.
\newblock
\urldef\tempurl%
\url{https://openreview.net/forum?id=B1xsqj09Fm}
\showURL{%
\tempurl}


\bibitem[Bromley et~al\mbox{.}(1993)]%
        {Bromley1993}
\bibfield{author}{\bibinfo{person}{Jane Bromley}, \bibinfo{person}{Isabelle Guyon}, \bibinfo{person}{Yann LeCun}, \bibinfo{person}{Eduard S\"{a}ckinger}, {and} \bibinfo{person}{Roopak Shah}.} \bibinfo{year}{1993}\natexlab{}.
\newblock \showarticletitle{Signature Verification Using a "Siamese" Time Delay Neural Network}. In \bibinfo{booktitle}{\emph{Proceedings of the 6th International Conference on Neural Information Processing Systems}} \emph{(\bibinfo{series}{NIPS'93})}. \bibinfo{pages}{737–744}.
\newblock


\bibitem[Brown(2010)]%
        {sampling_frame}
\bibfield{author}{\bibinfo{person}{R.S. Brown}.} \bibinfo{year}{2010}\natexlab{}.
\newblock \showarticletitle{Sampling}.
\newblock In \bibinfo{booktitle}{\emph{International Encyclopedia of Education (Third Edition)} (\bibinfo{edition}{third edition} ed.)}, \bibfield{editor}{\bibinfo{person}{Penelope Peterson}, \bibinfo{person}{Eva Baker}, {and} \bibinfo{person}{Barry McGaw}} (Eds.). \bibinfo{publisher}{Elsevier}, \bibinfo{address}{Oxford}, \bibinfo{pages}{142--146}.
\newblock
\showISBNx{978-0-08-044894-7}
\urldef\tempurl%
\url{https://doi.org/10.1016/B978-0-08-044894-7.00294-3}
\showDOI{\tempurl}


\bibitem[Cao et~al\mbox{.}(2022)]%
        {RECCE2022Cao}
\bibfield{author}{\bibinfo{person}{Junyi Cao}, \bibinfo{person}{Chao Ma}, \bibinfo{person}{Taiping Yao}, \bibinfo{person}{Shen Chen}, \bibinfo{person}{Shouhong Ding}, {and} \bibinfo{person}{Xiaokang Yang}.} \bibinfo{year}{2022}\natexlab{}.
\newblock \showarticletitle{End-to-End Reconstruction-Classification Learning for Face Forgery Detection}. In \bibinfo{booktitle}{\emph{Proceedings of the IEEE/CVF Conference on Computer Vision and Pattern Recognition (CVPR)}}. \bibinfo{pages}{4113--4122}.
\newblock


\bibitem[Cao et~al\mbox{.}(2021)]%
        {cao2021metric}
\bibfield{author}{\bibinfo{person}{Shenhao Cao}, \bibinfo{person}{Qin Zou}, \bibinfo{person}{Xiuqing Mao}, \bibinfo{person}{Dengpan Ye}, {and} \bibinfo{person}{Zhongyuan Wang}.} \bibinfo{year}{2021}\natexlab{}.
\newblock \showarticletitle{Metric Learning for Anti-Compression Facial Forgery Detection}. In \bibinfo{booktitle}{\emph{Proceedings of the 29th ACM International Conference on Multimedia (ACM MM 2021)}}. \bibinfo{pages}{1929--1937}.
\newblock


\bibitem[Carlini and Farid(2020)]%
        {Carlini2020Evading_CVPR_Workshops}
\bibfield{author}{\bibinfo{person}{Nicholas Carlini} {and} \bibinfo{person}{Hany Farid}.} \bibinfo{year}{2020}\natexlab{}.
\newblock \showarticletitle{Evading Deepfake-Image Detectors With White- and Black-Box Attacks}. In \bibinfo{booktitle}{\emph{Proceedings of the IEEE/CVF Conference on Computer Vision and Pattern Recognition (CVPR) Workshops}}.
\newblock


\bibitem[Chen et~al\mbox{.}(2022a)]%
        {Chen2022SNIS}
\bibfield{author}{\bibinfo{person}{Jiaxin Chen}, \bibinfo{person}{Xin Liao}, \bibinfo{person}{Wei Wang}, \bibinfo{person}{Zhenxing Qian}, \bibinfo{person}{Zheng Qin}, {and} \bibinfo{person}{Yaonan Wang}.} \bibinfo{year}{2022}\natexlab{a}.
\newblock \showarticletitle{SNIS: A Signal Noise Separation-based Network for Post-processed Image Forgery Detection}.
\newblock \bibinfo{journal}{\emph{IEEE Transactions on Circuits and Systems for Video Technology}} (\bibinfo{year}{2022}), \bibinfo{pages}{1--1}.
\newblock
\urldef\tempurl%
\url{https://doi.org/10.1109/TCSVT.2022.3204753}
\showDOI{\tempurl}


\bibitem[Chen et~al\mbox{.}(2022b)]%
        {SLADD2022Chen}
\bibfield{author}{\bibinfo{person}{Liang Chen}, \bibinfo{person}{Yong Zhang}, \bibinfo{person}{Yibing Song}, \bibinfo{person}{Lingqiao Liu}, {and} \bibinfo{person}{Jue Wang}.} \bibinfo{year}{2022}\natexlab{b}.
\newblock \showarticletitle{Self-supervised Learning of Adversarial Examples: Towards Good Generalizations for DeepFake Detections}. In \bibinfo{booktitle}{\emph{CVPR}}.
\newblock


\bibitem[Chen et~al\mbox{.}(2022c)]%
        {chen2022ost_nips}
\bibfield{author}{\bibinfo{person}{Liang Chen}, \bibinfo{person}{Yong Zhang}, \bibinfo{person}{Yibing Song}, \bibinfo{person}{Jue Wang}, {and} \bibinfo{person}{Lingqiao Liu}.} \bibinfo{year}{2022}\natexlab{c}.
\newblock \showarticletitle{{OST}: Improving Generalization of DeepFake Detection via One-Shot Test-Time Training}. In \bibinfo{booktitle}{\emph{Advances in Neural Information Processing Systems}}, \bibfield{editor}{\bibinfo{person}{Alice~H. Oh}, \bibinfo{person}{Alekh Agarwal}, \bibinfo{person}{Danielle Belgrave}, {and} \bibinfo{person}{Kyunghyun Cho}} (Eds.).
\newblock


\bibitem[Chen et~al\mbox{.}(2020)]%
        {Chen2020SimSwap}
\bibfield{author}{\bibinfo{person}{Renwang Chen}, \bibinfo{person}{Xuanhong Chen}, \bibinfo{person}{Bingbing Ni}, {and} \bibinfo{person}{Yanhao Ge}.} \bibinfo{year}{2020}\natexlab{}.
\newblock \showarticletitle{SimSwap: An Efficient Framework For High Fidelity Face Swapping}. In \bibinfo{booktitle}{\emph{{MM} '20: The 28th {ACM} International Conference on Multimedia}}.
\newblock


\bibitem[Chen et~al\mbox{.}(2021)]%
        {LocalRel2021Chen}
\bibfield{author}{\bibinfo{person}{Shen Chen}, \bibinfo{person}{Taiping Yao}, \bibinfo{person}{Yang Chen}, \bibinfo{person}{Shouhong Ding}, \bibinfo{person}{Jilin Li}, {and} \bibinfo{person}{Rongrong Ji}.} \bibinfo{year}{2021}\natexlab{}.
\newblock \showarticletitle{Local Relation Learning for Face Forgery Detection}.
\newblock \bibinfo{journal}{\emph{Proceedings of the AAAI Conference on Artificial Intelligence}} \bibinfo{volume}{35}, \bibinfo{number}{2} (\bibinfo{date}{May} \bibinfo{year}{2021}), \bibinfo{pages}{1081--1088}.
\newblock
\urldef\tempurl%
\url{https://ojs.aaai.org/index.php/AAAI/article/view/16193}
\showURL{%
\tempurl}


\bibitem[Cheng et~al\mbox{.}(2023)]%
        {VFD2023Cheng}
\bibfield{author}{\bibinfo{person}{Harry Cheng}, \bibinfo{person}{Yangyang Guo}, \bibinfo{person}{Tianyi Wang}, \bibinfo{person}{Qi Li}, \bibinfo{person}{Xiaojun Chang}, {and} \bibinfo{person}{Liqiang Nie}.} \bibinfo{year}{2023}\natexlab{}.
\newblock \showarticletitle{Voice-Face Homogeneity Tells Deepfake}.
\newblock \bibinfo{journal}{\emph{ACM Transactions on Multimedia Computing, Communications, and Applications}} \bibinfo{volume}{20}, \bibinfo{number}{3}, Article \bibinfo{articleno}{76} (\bibinfo{date}{nov} \bibinfo{year}{2023}), \bibinfo{numpages}{22}~pages.
\newblock
\showISSN{1551-6857}
\urldef\tempurl%
\url{https://doi.org/10.1145/3625231}
\showDOI{\tempurl}


\bibitem[Chinchilla(2021)]%
        {Rudy2021Cheerleader}
\bibfield{author}{\bibinfo{person}{Rudy Chinchilla}.} \bibinfo{year}{2021}\natexlab{}.
\newblock \bibinfo{title}{Mom Made Deepfake Nudes of Daughter's Cheer Teammates to Harass Them: Police}.
\newblock \bibinfo{howpublished}{\url{https://www.nbcphiladelphia.com/news/local/mom-made-deepfake-nudes-of-daughters-cheer-teammates-to-harass-them-police/2740906/}}.
\newblock
\newblock
\shownote{Accessed: 2022-09-08}.


\bibitem[Choi et~al\mbox{.}(2018)]%
        {stargan2018_choi}
\bibfield{author}{\bibinfo{person}{Yunjey Choi}, \bibinfo{person}{Minje Choi}, \bibinfo{person}{Munyoung Kim}, \bibinfo{person}{Jung-Woo Ha}, \bibinfo{person}{Sunghun Kim}, {and} \bibinfo{person}{Jaegul Choo}.} \bibinfo{year}{2018}\natexlab{}.
\newblock \showarticletitle{StarGAN: Unified Generative Adversarial Networks for Multi-domain Image-to-Image Translation}. In \bibinfo{booktitle}{\emph{2018 IEEE/CVF Conference on Computer Vision and Pattern Recognition}}. \bibinfo{pages}{8789--8797}.
\newblock
\urldef\tempurl%
\url{https://doi.org/10.1109/CVPR.2018.00916}
\showDOI{\tempurl}


\bibitem[Chollet(2017)]%
        {Chollet2017Xception}
\bibfield{author}{\bibinfo{person}{Francois Chollet}.} \bibinfo{year}{2017}\natexlab{}.
\newblock \showarticletitle{Xception: Deep Learning With Depthwise Separable Convolutions}. In \bibinfo{booktitle}{\emph{IEEE Conference on Computer Vision and Patten Recognition (CVPR)}}. \bibinfo{pages}{1251--1258}.
\newblock


\bibitem[Chuming et~al\mbox{.}(2021)]%
        {Practical2021Yang}
\bibfield{author}{\bibinfo{person}{Yang Chuming}, \bibinfo{person}{Daniel Wu}, {and} \bibinfo{person}{Ken Hong}.} \bibinfo{year}{2021}\natexlab{}.
\newblock \showarticletitle{Practical Deepfake Detection: Vulnerabilities in Global Contexts}. In \bibinfo{booktitle}{\emph{Responsible AI (RAT) - ICLR 2021 workshop}}.
\newblock


\bibitem[Ciftci et~al\mbox{.}(2020)]%
        {Ciftci2020FakeCatcher}
\bibfield{author}{\bibinfo{person}{Umur~Aybars Ciftci}, \bibinfo{person}{Ilke Demir}, {and} \bibinfo{person}{Lijun Yin}.} \bibinfo{year}{2020}\natexlab{}.
\newblock \showarticletitle{FakeCatcher: Detection of Synthetic Portrait Videos using Biological Signals}.
\newblock \bibinfo{journal}{\emph{IEEE Transactions on Pattern Analysis and Machine Intelligence}} (\bibinfo{year}{2020}), \bibinfo{pages}{1--1}.
\newblock
\urldef\tempurl%
\url{https://doi.org/10.1109/TPAMI.2020.3009287}
\showDOI{\tempurl}


\bibitem[Committee(2022)]%
        {EU2022antigen}
\bibfield{author}{\bibinfo{person}{EU~Health~Security Committee}.} \bibinfo{year}{2022}\natexlab{}.
\newblock \bibinfo{title}{EU Common list of COVID-19 antigen tests}.
\newblock \bibinfo{howpublished}{\url{https://health.ec.europa.eu/system/files/2022-07/covid-19\_eu-common-list-antigen-tests\_en.pdf}}.
\newblock
\newblock
\shownote{Accessed: 2022-09-09}.


\bibitem[Cozzolino et~al\mbox{.}(2014)]%
        {Cozzolino2014Image}
\bibfield{author}{\bibinfo{person}{Davide Cozzolino}, \bibinfo{person}{Diego Gragnaniello}, {and} \bibinfo{person}{Luisa Verdoliva}.} \bibinfo{year}{2014}\natexlab{}.
\newblock \showarticletitle{Image forgery localization through the fusion of camera-based, feature-based and pixel-based techniques}. In \bibinfo{booktitle}{\emph{2014 IEEE International Conference on Image Processing (ICIP)}}. \bibinfo{pages}{5302--5306}.
\newblock
\urldef\tempurl%
\url{https://doi.org/10.1109/ICIP.2014.7026073}
\showDOI{\tempurl}


\bibitem[Cozzolino and Verdoliva(2020)]%
        {Cozzolino2020noiseprint}
\bibfield{author}{\bibinfo{person}{D. Cozzolino} {and} \bibinfo{person}{L. Verdoliva}.} \bibinfo{year}{2020}\natexlab{}.
\newblock \showarticletitle{Noiseprint: A CNN-Based Camera Model Fingerprint}.
\newblock \bibinfo{journal}{\emph{IEEE Transactions on Information Forensics and Security}}  \bibinfo{volume}{15} (\bibinfo{year}{2020}), \bibinfo{pages}{144--159}.
\newblock
\urldef\tempurl%
\url{https://doi.org/10.1109/TIFS.2019.2916364}
\showDOI{\tempurl}


\bibitem[Cozzolino Giovanni Poggi Luisa~Verdoliva(2019)]%
        {Verdoliva2019Extracting}
\bibfield{author}{\bibinfo{person}{Davide Cozzolino Giovanni Poggi Luisa~Verdoliva}.} \bibinfo{year}{2019}\natexlab{}.
\newblock \showarticletitle{Extracting camera-based fingerprints for video forensics}. In \bibinfo{booktitle}{\emph{Proceedings of the IEEE/CVF Conference on Computer Vision and Pattern Recognition (CVPR) Workshops}}.
\newblock


\bibitem[Dang et~al\mbox{.}(2020)]%
        {Dang2020on}
\bibfield{author}{\bibinfo{person}{Hao Dang}, \bibinfo{person}{Feng Liu}, \bibinfo{person}{Joel Stehouwer}, \bibinfo{person}{Xiaoming Liu}, {and} \bibinfo{person}{Anil~K. Jain}.} \bibinfo{year}{2020}\natexlab{}.
\newblock \showarticletitle{On the Detection of Digital Face Manipulation}. In \bibinfo{booktitle}{\emph{Proceedings of the IEEE/CVF Conference on Computer Vision and Pattern Recognition (CVPR)}}.
\newblock


\bibitem[de~Weever and Wilczek(2020)]%
        {Weever2020DeepfakeDT}
\bibfield{author}{\bibinfo{person}{Catherine de Weever} {and} \bibinfo{person}{S. Wilczek}.} \bibinfo{year}{2020}\natexlab{}.
\newblock \showarticletitle{Deepfake detection through PRNU and logistic regression analyses}.
\newblock


\bibitem[deepfakes(2018)]%
        {Guilloux2018FakeApp}
\bibfield{author}{\bibinfo{person}{deepfakes}.} \bibinfo{year}{2018}\natexlab{}.
\newblock \bibinfo{title}{FakeApp}.
\newblock \bibinfo{howpublished}{\url{https://www.malavida.com/en/soft/fakeapp/}}.
\newblock
\newblock
\shownote{Accessed: 2022-09-08}.


\bibitem[deepfakes(2019)]%
        {deepfakes}
\bibfield{author}{\bibinfo{person}{deepfakes}.} \bibinfo{year}{2019}\natexlab{}.
\newblock \bibinfo{title}{{DeepFakes}}.
\newblock \bibinfo{howpublished}{\url{https://github.com/deepfakes/}}.
\newblock
\newblock
\shownote{Accessed: 2022-09-08}.


\bibitem[Deng et~al\mbox{.}(2019)]%
        {Deng2019ArcFace}
\bibfield{author}{\bibinfo{person}{Jiankang Deng}, \bibinfo{person}{Jia Guo}, \bibinfo{person}{Niannan Xue}, {and} \bibinfo{person}{Stefanos Zafeiriou}.} \bibinfo{year}{2019}\natexlab{}.
\newblock \showarticletitle{ArcFace: Additive Angular Margin Loss for Deep Face Recognition}. In \bibinfo{booktitle}{\emph{Proceedings of the IEEE/CVF Conference on Computer Vision and Pattern Recognition (CVPR)}}.
\newblock


\bibitem[Ding et~al\mbox{.}(2018)]%
        {Ding2018ExprGANFE}
\bibfield{author}{\bibinfo{person}{Hui Ding}, \bibinfo{person}{Kumar Sricharan}, {and} \bibinfo{person}{Rama Chellappa}.} \bibinfo{year}{2018}\natexlab{}.
\newblock \showarticletitle{ExprGAN: Facial Expression Editing with Controllable Expression Intensity}. In \bibinfo{booktitle}{\emph{AAAI}}.
\newblock


\bibitem[Dolhansky et~al\mbox{.}(2020)]%
        {DFDC}
\bibfield{author}{\bibinfo{person}{Brian Dolhansky}, \bibinfo{person}{Joanna Bitton}, \bibinfo{person}{Ben Pflaum}, \bibinfo{person}{Jikuo Lu}, \bibinfo{person}{Russ Howes}, \bibinfo{person}{Menglin Wang}, {and} \bibinfo{person}{Cristian~Canton Ferrer}.} \bibinfo{year}{2020}\natexlab{}.
\newblock \bibinfo{title}{The DeepFake Detection Challenge (DFDC) Dataset}.
\newblock
\newblock
\urldef\tempurl%
\url{https://doi.org/10.48550/ARXIV.2006.07397}
\showDOI{\tempurl}


\bibitem[Dolhansky et~al\mbox{.}(2019)]%
        {DFDC_Preview}
\bibfield{author}{\bibinfo{person}{Brian Dolhansky}, \bibinfo{person}{Russ Howes}, \bibinfo{person}{Ben Pflaum}, \bibinfo{person}{Nicole Baram}, {and} \bibinfo{person}{Cristian~Canton Ferrer}.} \bibinfo{year}{2019}\natexlab{}.
\newblock \bibinfo{title}{The Deepfake Detection Challenge (DFDC) Preview Dataset}.
\newblock
\newblock
\urldef\tempurl%
\url{https://doi.org/10.48550/ARXIV.1910.08854}
\showDOI{\tempurl}


\bibitem[Dong et~al\mbox{.}(2022)]%
        {Explaining2022Dong_ECCV}
\bibfield{author}{\bibinfo{person}{Shichao Dong}, \bibinfo{person}{Jin Wang}, \bibinfo{person}{Jiajun Liang}, \bibinfo{person}{Haoqiang Fan}, {and} \bibinfo{person}{Renhe Ji}.} \bibinfo{year}{2022}\natexlab{}.
\newblock \showarticletitle{Explaining Deepfake Detection by Analysing Image Matching}. In \bibinfo{booktitle}{\emph{Computer Vision -- ECCV 2022}}, \bibfield{editor}{\bibinfo{person}{Shai Avidan}, \bibinfo{person}{Gabriel Brostow}, \bibinfo{person}{Moustapha Ciss{\'e}}, \bibinfo{person}{Giovanni~Maria Farinella}, {and} \bibinfo{person}{Tal Hassner}} (Eds.). \bibinfo{publisher}{Springer Nature Switzerland}, \bibinfo{address}{Cham}, \bibinfo{pages}{18--35}.
\newblock
\showISBNx{978-3-031-19781-9}


\bibitem[Dosovitskiy et~al\mbox{.}(2021)]%
        {Alexey2021An}
\bibfield{author}{\bibinfo{person}{Alexey Dosovitskiy}, \bibinfo{person}{Lucas Beyer}, \bibinfo{person}{Alexander Kolesnikov}, \bibinfo{person}{Dirk Weissenborn}, \bibinfo{person}{Xiaohua Zhai}, \bibinfo{person}{Thomas Unterthiner}, \bibinfo{person}{Mostafa Dehghani}, \bibinfo{person}{Matthias Minderer}, \bibinfo{person}{Georg Heigold}, \bibinfo{person}{Sylvain Gelly}, \bibinfo{person}{Jakob Uszkoreit}, {and} \bibinfo{person}{Neil Houlsby}.} \bibinfo{year}{2021}\natexlab{}.
\newblock \showarticletitle{An Image is Worth 16x16 Words: Transformers for Image Recognition at Scale}. In \bibinfo{booktitle}{\emph{International Conference on Learning Representations}}.
\newblock


\bibitem[Durall et~al\mbox{.}(2020)]%
        {durall2020unmasking}
\bibfield{author}{\bibinfo{person}{Ricard Durall}, \bibinfo{person}{Margret Keuper}, \bibinfo{person}{Franz-Josef Pfreundt}, {and} \bibinfo{person}{Janis Keuper}.} \bibinfo{year}{2020}\natexlab{}.
\newblock \bibinfo{title}{Unmasking DeepFakes with simple Features}.
\newblock
\newblock
\urldef\tempurl%
\url{https://doi.org/10.48550/ARXIV.1911.00686}
\showDOI{\tempurl}


\bibitem[Farid(2022)]%
        {Farid2022Creating}
\bibfield{author}{\bibinfo{person}{Hany Farid}.} \bibinfo{year}{2022}\natexlab{}.
\newblock \showarticletitle{Creating, Using, Misusing, and Detecting Deep Fakes}.
\newblock \bibinfo{journal}{\emph{Journal of Online Trust and Safety}} \bibinfo{volume}{1}, \bibinfo{number}{4} (\bibinfo{date}{Sep.} \bibinfo{year}{2022}).
\newblock
\urldef\tempurl%
\url{https://doi.org/10.54501/jots.v1i4.56}
\showDOI{\tempurl}


\bibitem[Ferrara et~al\mbox{.}(2012)]%
        {Ferrara2012ImageFL}
\bibfield{author}{\bibinfo{person}{Pasquale Ferrara}, \bibinfo{person}{Tiziano Bianchi}, \bibinfo{person}{Alessia~De Rosa}, {and} \bibinfo{person}{Alessandro Piva}.} \bibinfo{year}{2012}\natexlab{}.
\newblock \showarticletitle{Image Forgery Localization via Fine-Grained Analysis of CFA Artifacts}.
\newblock \bibinfo{journal}{\emph{IEEE Transactions on Information Forensics and Security}}  \bibinfo{volume}{7} (\bibinfo{year}{2012}), \bibinfo{pages}{1566--1577}.
\newblock


\bibitem[FFmpeg(2021)]%
        {ffmpeg}
\bibfield{author}{\bibinfo{person}{FFmpeg}.} \bibinfo{year}{2021}\natexlab{}.
\newblock \bibinfo{title}{{FFmpeg}}.
\newblock \bibinfo{howpublished}{\url{https://www.ffmpeg.org/}}.
\newblock
\newblock
\shownote{Accessed: 2021-08-29}.


\bibitem[Fischer(2011)]%
        {CLT2011Fischer}
\bibfield{author}{\bibinfo{person}{Hans Fischer}.} \bibinfo{year}{2011}\natexlab{}.
\newblock \bibinfo{booktitle}{\emph{The Central Limit Theorem from Laplace to Cauchy: Changes in Stochastic Objectives and in Analytical Methods}}.
\newblock \bibinfo{publisher}{Springer New York}, \bibinfo{address}{New York, NY}, \bibinfo{pages}{17--74}.
\newblock
\showISBNx{978-0-387-87857-7}
\urldef\tempurl%
\url{https://doi.org/10.1007/978-0-387-87857-7\_2}
\showDOI{\tempurl}


\bibitem[Frank et~al\mbox{.}(2020)]%
        {LeveFreq2020Frank}
\bibfield{author}{\bibinfo{person}{Joel Frank}, \bibinfo{person}{Thorsten Eisenhofer}, \bibinfo{person}{Lea Sch\"{o}nherr}, \bibinfo{person}{Asja Fischer}, \bibinfo{person}{Dorothea Kolossa}, {and} \bibinfo{person}{Thorsten Holz}.} \bibinfo{year}{2020}\natexlab{}.
\newblock \showarticletitle{Leveraging Frequency Analysis for Deep Fake Image Recognition}. In \bibinfo{booktitle}{\emph{Proceedings of the 37th International Conference on Machine Learning}} \emph{(\bibinfo{series}{ICML'20})}. \bibinfo{publisher}{JMLR.org}, Article \bibinfo{articleno}{304}, \bibinfo{numpages}{12}~pages.
\newblock


\bibitem[Fridrich and Kodovsky(2012)]%
        {Fridrich2012Rich}
\bibfield{author}{\bibinfo{person}{Jessica Fridrich} {and} \bibinfo{person}{Jan Kodovsky}.} \bibinfo{year}{2012}\natexlab{}.
\newblock \showarticletitle{Rich Models for Steganalysis of Digital Images}.
\newblock \bibinfo{journal}{\emph{IEEE Transactions on Information Forensics and Security}} \bibinfo{volume}{7}, \bibinfo{number}{3} (\bibinfo{year}{2012}), \bibinfo{pages}{868--882}.
\newblock
\urldef\tempurl%
\url{https://doi.org/10.1109/TIFS.2012.2190402}
\showDOI{\tempurl}


\bibitem[Gandhi and Jain(2020)]%
        {adversarial2020gandhi}
\bibfield{author}{\bibinfo{person}{Apurva Gandhi} {and} \bibinfo{person}{Shomik Jain}.} \bibinfo{year}{2020}\natexlab{}.
\newblock \showarticletitle{Adversarial Perturbations Fool Deepfake Detectors}. In \bibinfo{booktitle}{\emph{2020 International Joint Conference on Neural Networks (IJCNN)}}. \bibinfo{pages}{1--8}.
\newblock
\urldef\tempurl%
\url{https://doi.org/10.1109/IJCNN48605.2020.9207034}
\showDOI{\tempurl}


\bibitem[Gao et~al\mbox{.}(2021)]%
        {gao2021info}
\bibfield{author}{\bibinfo{person}{Gege Gao}, \bibinfo{person}{Huaibo Huang}, \bibinfo{person}{Chaoyou Fu}, \bibinfo{person}{Zhaoyang Li}, {and} \bibinfo{person}{Ran He}.} \bibinfo{year}{2021}\natexlab{}.
\newblock \showarticletitle{Information Bottleneck Disentanglement for Identity Swapping}. In \bibinfo{booktitle}{\emph{2021 IEEE/CVF Conference on Computer Vision and Pattern Recognition (CVPR)}}. \bibinfo{pages}{3403--3412}.
\newblock
\urldef\tempurl%
\url{https://doi.org/10.1109/CVPR46437.2021.00341}
\showDOI{\tempurl}


\bibitem[Goodfellow et~al\mbox{.}(2014)]%
        {Goodfellow2014Generative}
\bibfield{author}{\bibinfo{person}{Ian Goodfellow}, \bibinfo{person}{Jean Pouget-Abadie}, \bibinfo{person}{Mehdi Mirza}, \bibinfo{person}{Bing Xu}, \bibinfo{person}{David Warde-Farley}, \bibinfo{person}{Sherjil Ozair}, \bibinfo{person}{Aaron Courville}, {and} \bibinfo{person}{Yoshua Bengio}.} \bibinfo{year}{2014}\natexlab{}.
\newblock \showarticletitle{Generative Adversarial Nets}.
\newblock In \bibinfo{booktitle}{\emph{The 27th Neural Information Processing Systems Advances}}. \bibinfo{pages}{2672--2680}.
\newblock


\bibitem[Goodman et~al\mbox{.}(2020)]%
        {goodman2020advbox}
\bibfield{author}{\bibinfo{person}{Dou Goodman}, \bibinfo{person}{Hao Xin}, \bibinfo{person}{Wang Yang}, \bibinfo{person}{Wu Yuesheng}, \bibinfo{person}{Xiong Junfeng}, {and} \bibinfo{person}{Zhang Huan}.} \bibinfo{year}{2020}\natexlab{}.
\newblock \bibinfo{title}{Advbox: a toolbox to generate adversarial examples that fool neural networks}.
\newblock
\newblock
\showeprint[arxiv]{2001.05574}~[cs.LG]


\bibitem[Government(2022)]%
        {HK2022antigen}
\bibfield{author}{\bibinfo{person}{Hong Kong Special Administrative~Region Government}.} \bibinfo{year}{2022}\natexlab{}.
\newblock \bibinfo{title}{Rapid Antigen Test (RAT) for COVID-19}.
\newblock \bibinfo{howpublished}{\url{https://www.coronavirus.gov.hk/pdf/RapAgTest\_FAQ\_ENG.pdf}}.
\newblock
\newblock
\shownote{Accessed: 2022-09-09}.


\bibitem[Gu et~al\mbox{.}(2022)]%
        {Gu2022Exploiting}
\bibfield{author}{\bibinfo{person}{Qiqi Gu}, \bibinfo{person}{Shen Chen}, \bibinfo{person}{Taiping Yao}, \bibinfo{person}{Yang Chen}, \bibinfo{person}{Shouhong Ding}, {and} \bibinfo{person}{Ran Yi}.} \bibinfo{year}{2022}\natexlab{}.
\newblock \showarticletitle{Exploiting Fine-Grained Face Forgery Clues via Progressive Enhancement Learning}.
\newblock \bibinfo{journal}{\emph{Proceedings of the AAAI Conference on Artificial Intelligence}} \bibinfo{volume}{36}, \bibinfo{number}{1} (\bibinfo{date}{Jun.} \bibinfo{year}{2022}), \bibinfo{pages}{735--743}.
\newblock
\urldef\tempurl%
\url{https://doi.org/10.1609/aaai.v36i1.19954}
\showDOI{\tempurl}


\bibitem[Guan et~al\mbox{.}(2022)]%
        {Delving2022Guan_NIPS}
\bibfield{author}{\bibinfo{person}{Jiazhi Guan}, \bibinfo{person}{Hang Zhou}, \bibinfo{person}{Zhibin Hong}, \bibinfo{person}{Errui Ding}, \bibinfo{person}{Jingdong Wang}, \bibinfo{person}{Chengbin Quan}, {and} \bibinfo{person}{Youjian Zhao}.} \bibinfo{year}{2022}\natexlab{}.
\newblock \showarticletitle{Delving into Sequential Patches for Deepfake Detection}. In \bibinfo{booktitle}{\emph{Advances in Neural Information Processing Systems}}, \bibfield{editor}{\bibinfo{person}{Alice~H. Oh}, \bibinfo{person}{Alekh Agarwal}, \bibinfo{person}{Danielle Belgrave}, {and} \bibinfo{person}{Kyunghyun Cho}} (Eds.).
\newblock


\bibitem[Guarnera et~al\mbox{.}(2020a)]%
        {Guarnera2020DeepFake}
\bibfield{author}{\bibinfo{person}{Luca Guarnera}, \bibinfo{person}{Oliver Giudice}, {and} \bibinfo{person}{Sebastiano Battiato}.} \bibinfo{year}{2020}\natexlab{a}.
\newblock \showarticletitle{DeepFake Detection by Analyzing Convolutional Traces}. In \bibinfo{booktitle}{\emph{Proceedings of the IEEE/CVF Conference on Computer Vision and Pattern Recognition (CVPR) Workshops}}.
\newblock


\bibitem[Guarnera et~al\mbox{.}(2020b)]%
        {Guarnera2020Fighting}
\bibfield{author}{\bibinfo{person}{Luca Guarnera}, \bibinfo{person}{Oliver Giudice}, {and} \bibinfo{person}{Sebastiano Battiato}.} \bibinfo{year}{2020}\natexlab{b}.
\newblock \showarticletitle{Fighting Deepfake by Exposing the Convolutional Traces on Images}.
\newblock \bibinfo{journal}{\emph{IEEE Access}}  \bibinfo{volume}{8} (\bibinfo{year}{2020}), \bibinfo{pages}{165085--165098}.
\newblock
\urldef\tempurl%
\url{https://doi.org/10.1109/ACCESS.2020.3023037}
\showDOI{\tempurl}


\bibitem[Gully(2019)]%
        {DFD2019Nick}
\bibfield{author}{\bibinfo{person}{Nick Dufourand~Andrew Gully}.} \bibinfo{year}{2019}\natexlab{}.
\newblock \bibinfo{title}{Contributing Data to Deepfake Detection Research}.
\newblock \bibinfo{howpublished}{\url{https://ai.googleblog.com/2019/09/contributing-data-to-deepfake-detection.html}}.
\newblock
\newblock
\shownote{Accessed: 2022-09-08}.


\bibitem[Guo et~al\mbox{.}(2021)]%
        {GUO2021fake}
\bibfield{author}{\bibinfo{person}{Zhiqing Guo}, \bibinfo{person}{Gaobo Yang}, \bibinfo{person}{Jiyou Chen}, {and} \bibinfo{person}{Xingming Sun}.} \bibinfo{year}{2021}\natexlab{}.
\newblock \showarticletitle{Fake face detection via adaptive manipulation traces extraction network}.
\newblock \bibinfo{journal}{\emph{Computer Vision and Image Understanding}}  \bibinfo{volume}{204} (\bibinfo{year}{2021}), \bibinfo{pages}{103170}.
\newblock
\showISSN{1077-3142}
\urldef\tempurl%
\url{https://doi.org/10.1016/j.cviu.2021.103170}
\showDOI{\tempurl}


\bibitem[Guo et~al\mbox{.}(2022)]%
        {Guo2022Exposing}
\bibfield{author}{\bibinfo{person}{Zhiqing Guo}, \bibinfo{person}{Gaobo Yang}, \bibinfo{person}{Jiyou Chen}, {and} \bibinfo{person}{Xingming Sun}.} \bibinfo{year}{2022}\natexlab{}.
\newblock \bibinfo{title}{Exposing Deepfake Face Forgeries with Guided Residuals}.
\newblock
\newblock
\urldef\tempurl%
\url{https://doi.org/10.48550/ARXIV.2205.00753}
\showDOI{\tempurl}


\bibitem[Güera and Delp(2018)]%
        {deepfake2018david}
\bibfield{author}{\bibinfo{person}{David Güera} {and} \bibinfo{person}{Edward~J. Delp}.} \bibinfo{year}{2018}\natexlab{}.
\newblock \showarticletitle{Deepfake Video Detection Using Recurrent Neural Networks}. In \bibinfo{booktitle}{\emph{2018 15th IEEE International Conference on Advanced Video and Signal Based Surveillance (AVSS)}}. \bibinfo{pages}{1--6}.
\newblock
\urldef\tempurl%
\url{https://doi.org/10.1109/AVSS.2018.8639163}
\showDOI{\tempurl}


\bibitem[Haliassos et~al\mbox{.}(2022)]%
        {Leveraging2022Haliassos_CVPR}
\bibfield{author}{\bibinfo{person}{Alexandros Haliassos}, \bibinfo{person}{Rodrigo Mira}, \bibinfo{person}{Stavros Petridis}, {and} \bibinfo{person}{Maja Pantic}.} \bibinfo{year}{2022}\natexlab{}.
\newblock \showarticletitle{Leveraging Real Talking Faces via Self-Supervision for Robust Forgery Detection}. In \bibinfo{booktitle}{\emph{2022 IEEE/CVF Conference on Computer Vision and Pattern Recognition (CVPR)}}. \bibinfo{pages}{14930--14942}.
\newblock
\urldef\tempurl%
\url{https://doi.org/10.1109/CVPR52688.2022.01453}
\showDOI{\tempurl}


\bibitem[Harwell(2021)]%
        {Harwell2021CheerProve}
\bibfield{author}{\bibinfo{person}{Drew Harwell}.} \bibinfo{year}{2021}\natexlab{}.
\newblock \bibinfo{title}{Remember the ‘deepfake cheerleader mom’? Prosecutors now admit they can’t prove fake-video claims}.
\newblock \bibinfo{howpublished}{\url{https://www.washingtonpost.com/technology/2021/05/14/deepfake-cheer-mom-claims-dropped/}}.
\newblock
\newblock
\shownote{Accessed: 2022-09-08}.


\bibitem[He et~al\mbox{.}(2015)]%
        {He2015spatial}
\bibfield{author}{\bibinfo{person}{Kaiming He}, \bibinfo{person}{Xiangyu Zhang}, \bibinfo{person}{Shaoqing Ren}, {and} \bibinfo{person}{Jian Sun}.} \bibinfo{year}{2015}\natexlab{}.
\newblock \showarticletitle{Spatial Pyramid Pooling in Deep Convolutional Networks for Visual Recognition}.
\newblock \bibinfo{journal}{\emph{IEEE Transactions on Pattern Analysis and Machine Intelligence}} \bibinfo{volume}{37}, \bibinfo{number}{9} (\bibinfo{year}{2015}), \bibinfo{pages}{1904--1916}.
\newblock
\urldef\tempurl%
\url{https://doi.org/10.1109/TPAMI.2015.2389824}
\showDOI{\tempurl}


\bibitem[He et~al\mbox{.}(2016)]%
        {He2016Deep}
\bibfield{author}{\bibinfo{person}{Kaiming He}, \bibinfo{person}{Xiangyu Zhang}, \bibinfo{person}{Shaoqing Ren}, {and} \bibinfo{person}{Jian Sun}.} \bibinfo{year}{2016}\natexlab{}.
\newblock \showarticletitle{Deep residual learning for image recognition}. In \bibinfo{booktitle}{\emph{IEEE Conference on Computer Vision and Pattern Recognition}}. \bibinfo{pages}{770--778}.
\newblock
\urldef\tempurl%
\url{https://doi.org/10.1109/CVPR.2016.90}
\showDOI{\tempurl}


\bibitem[He et~al\mbox{.}(2021)]%
        {yang2021beyond_ijcai}
\bibfield{author}{\bibinfo{person}{Yang He}, \bibinfo{person}{Ning Yu}, \bibinfo{person}{Margret Keuper}, {and} \bibinfo{person}{Mario Fritz}.} \bibinfo{year}{2021}\natexlab{}.
\newblock \showarticletitle{Beyond the Spectrum: Detecting Deepfakes via Re-synthesis}. In \bibinfo{booktitle}{\emph{30th International Joint Conference on Artificial Intelligence (IJCAI)}}.
\newblock


\bibitem[Hearst et~al\mbox{.}(1998)]%
        {SVM1998_Hearst}
\bibfield{author}{\bibinfo{person}{M.A. Hearst}, \bibinfo{person}{S.T. Dumais}, \bibinfo{person}{E. Osuna}, \bibinfo{person}{J. Platt}, {and} \bibinfo{person}{B. Scholkopf}.} \bibinfo{year}{1998}\natexlab{}.
\newblock \showarticletitle{Support vector machines}.
\newblock \bibinfo{journal}{\emph{IEEE Intelligent Systems and their Applications}} \bibinfo{volume}{13}, \bibinfo{number}{4} (\bibinfo{year}{1998}), \bibinfo{pages}{18--28}.
\newblock
\urldef\tempurl%
\url{https://doi.org/10.1109/5254.708428}
\showDOI{\tempurl}


\bibitem[Heo et~al\mbox{.}(2022)]%
        {Heo2022DeepFake}
\bibfield{author}{\bibinfo{person}{Young-Jin Heo}, \bibinfo{person}{Woon-Ha Yeo}, {and} \bibinfo{person}{Byung-Gyu Kim}.} \bibinfo{year}{2022}\natexlab{}.
\newblock \showarticletitle{DeepFake detection algorithm based on improved vision transformer}.
\newblock \bibinfo{journal}{\emph{Applied Intelligence}} (\bibinfo{year}{2022}).
\newblock
\urldef\tempurl%
\url{https://doi.org/10.1007/s10489-022-03867-9}
\showDOI{\tempurl}


\bibitem[Hochreiter and Schmidhuber(1997)]%
        {lstm}
\bibfield{author}{\bibinfo{person}{Sepp Hochreiter} {and} \bibinfo{person}{Jürgen Schmidhuber}.} \bibinfo{year}{1997}\natexlab{}.
\newblock \showarticletitle{Long Short-term Memory}.
\newblock \bibinfo{journal}{\emph{Neural computation}}  \bibinfo{volume}{9} (\bibinfo{date}{12} \bibinfo{year}{1997}), \bibinfo{pages}{1735--80}.
\newblock
\urldef\tempurl%
\url{https://doi.org/10.1162/neco.1997.9.8.1735}
\showDOI{\tempurl}


\bibitem[Hooda et~al\mbox{.}(2022)]%
        {hooda2022towards}
\bibfield{author}{\bibinfo{person}{Ashish Hooda}, \bibinfo{person}{Neal Mangaokar}, \bibinfo{person}{Ryan Feng}, \bibinfo{person}{Kassem Fawaz}, \bibinfo{person}{Somesh Jha}, {and} \bibinfo{person}{Atul Prakash}.} \bibinfo{year}{2022}\natexlab{}.
\newblock \bibinfo{title}{Towards Adversarially Robust Deepfake Detection: An Ensemble Approach}.
\newblock
\newblock
\urldef\tempurl%
\url{https://doi.org/10.48550/ARXIV.2202.05687}
\showDOI{\tempurl}


\bibitem[Hsu et~al\mbox{.}(2020)]%
        {hsu2020deep}
\bibfield{author}{\bibinfo{person}{Chih-Chung Hsu}, \bibinfo{person}{Yi-Xiu Zhuang}, {and} \bibinfo{person}{Chia-Yen Lee}.} \bibinfo{year}{2020}\natexlab{}.
\newblock \showarticletitle{Deep Fake Image Detection Based on Pairwise Learning}.
\newblock \bibinfo{journal}{\emph{Applied Sciences}} \bibinfo{volume}{10}, \bibinfo{number}{1} (\bibinfo{year}{2020}).
\newblock
\showISSN{2076-3417}
\urldef\tempurl%
\url{https://doi.org/10.3390/app10010370}
\showDOI{\tempurl}


\bibitem[Hu et~al\mbox{.}(2022a)]%
        {Hu2022FInfer}
\bibfield{author}{\bibinfo{person}{Juan Hu}, \bibinfo{person}{Xin Liao}, \bibinfo{person}{Jinwen Liang}, \bibinfo{person}{Wenbo Zhou}, {and} \bibinfo{person}{Zheng Qin}.} \bibinfo{year}{2022}\natexlab{a}.
\newblock \showarticletitle{FInfer: Frame Inference-Based Deepfake Detection for High-Visual-Quality Videos}.
\newblock \bibinfo{journal}{\emph{Proceedings of the AAAI Conference on Artificial Intelligence}} \bibinfo{volume}{36}, \bibinfo{number}{1} (\bibinfo{date}{Jun.} \bibinfo{year}{2022}), \bibinfo{pages}{951--959}.
\newblock
\urldef\tempurl%
\url{https://doi.org/10.1609/aaai.v36i1.19978}
\showDOI{\tempurl}


\bibitem[Hu et~al\mbox{.}(2022b)]%
        {Hu2021Detecting}
\bibfield{author}{\bibinfo{person}{Juan Hu}, \bibinfo{person}{Xin Liao}, \bibinfo{person}{Wei Wang}, {and} \bibinfo{person}{Zheng Qin}.} \bibinfo{year}{2022}\natexlab{b}.
\newblock \showarticletitle{Detecting Compressed Deepfake Videos in Social Networks Using Frame-Temporality Two-Stream Convolutional Network}.
\newblock \bibinfo{journal}{\emph{IEEE Transactions on Circuits and Systems for Video Technology}} \bibinfo{volume}{32}, \bibinfo{number}{3} (\bibinfo{year}{2022}), \bibinfo{pages}{1089--1102}.
\newblock
\urldef\tempurl%
\url{https://doi.org/10.1109/TCSVT.2021.3074259}
\showDOI{\tempurl}


\bibitem[Huang and De~La~Torre(2012)]%
        {mmnn2012Huang}
\bibfield{author}{\bibinfo{person}{Dong Huang} {and} \bibinfo{person}{Fernando De~La~Torre}.} \bibinfo{year}{2012}\natexlab{}.
\newblock \showarticletitle{Facial Action Transfer with Personalized Bilinear Regression}. In \bibinfo{booktitle}{\emph{Computer Vision -- ECCV 2012}}, \bibfield{editor}{\bibinfo{person}{Andrew Fitzgibbon}, \bibinfo{person}{Svetlana Lazebnik}, \bibinfo{person}{Pietro Perona}, \bibinfo{person}{Yoichi Sato}, {and} \bibinfo{person}{Cordelia Schmid}} (Eds.). \bibinfo{publisher}{Springer Berlin Heidelberg}, \bibinfo{address}{Berlin, Heidelberg}, \bibinfo{pages}{144--158}.
\newblock
\showISBNx{978-3-642-33709-3}


\bibitem[Huang et~al\mbox{.}(2017)]%
        {dense_net}
\bibfield{author}{\bibinfo{person}{Gao Huang}, \bibinfo{person}{Zhuang Liu}, \bibinfo{person}{Laurens Van Der~Maaten}, {and} \bibinfo{person}{Kilian~Q. Weinberger}.} \bibinfo{year}{2017}\natexlab{}.
\newblock \showarticletitle{Densely Connected Convolutional Networks}. In \bibinfo{booktitle}{\emph{2017 IEEE Conference on Computer Vision and Pattern Recognition (CVPR)}}. \bibinfo{pages}{2261--2269}.
\newblock
\urldef\tempurl%
\url{https://doi.org/10.1109/CVPR.2017.243}
\showDOI{\tempurl}


\bibitem[Hussain et~al\mbox{.}(2022)]%
        {Hussain2022Exposing}
\bibfield{author}{\bibinfo{person}{Shehzeen Hussain}, \bibinfo{person}{Paarth Neekhara}, \bibinfo{person}{Brian Dolhansky}, \bibinfo{person}{Joanna Bitton}, \bibinfo{person}{Cristian~Canton Ferrer}, \bibinfo{person}{Julian McAuley}, {and} \bibinfo{person}{Farinaz Koushanfar}.} \bibinfo{year}{2022}\natexlab{}.
\newblock \showarticletitle{Exposing Vulnerabilities of Deepfake Detection Systems with Robust Attacks}.
\newblock \bibinfo{journal}{\emph{Digital Threats}} \bibinfo{volume}{3}, \bibinfo{number}{3}, Article \bibinfo{articleno}{30} (\bibinfo{date}{feb} \bibinfo{year}{2022}), \bibinfo{numpages}{23}~pages.
\newblock
\showISSN{2692-1626}
\urldef\tempurl%
\url{https://doi.org/10.1145/3464307}
\showDOI{\tempurl}


\bibitem[Inc(2021)]%
        {wombo}
\bibfield{author}{\bibinfo{person}{Wombo~Studios Inc}.} \bibinfo{year}{2021}\natexlab{}.
\newblock \bibinfo{title}{Wombo: Make your selfies sing}.
\newblock \bibinfo{howpublished}{\url{https://play.google.com/store/apps/details?id=com.womboai.wombo&hl=en&gl=US}}.
\newblock
\newblock
\shownote{Accessed: 2022-09-08}.


\bibitem[Jafar et~al\mbox{.}(2020)]%
        {Jafar2020forensics}
\bibfield{author}{\bibinfo{person}{Mousa~Tayseer Jafar}, \bibinfo{person}{Mohammad Ababneh}, \bibinfo{person}{Mohammad Al-Zoube}, {and} \bibinfo{person}{Ammar Elhassan}.} \bibinfo{year}{2020}\natexlab{}.
\newblock \showarticletitle{Forensics and Analysis of Deepfake Videos}. In \bibinfo{booktitle}{\emph{2020 11th International Conference on Information and Communication Systems (ICICS)}}. \bibinfo{pages}{053--058}.
\newblock
\urldef\tempurl%
\url{https://doi.org/10.1109/ICICS49469.2020.239493}
\showDOI{\tempurl}


\bibitem[Jeon et~al\mbox{.}(2020)]%
        {Jeon2020FDFtNet}
\bibfield{author}{\bibinfo{person}{Hyeonseong Jeon}, \bibinfo{person}{Youngoh Bang}, {and} \bibinfo{person}{Simon~S. Woo}.} \bibinfo{year}{2020}\natexlab{}.
\newblock \showarticletitle{FDFtNet: Facing Off Fake Images Using Fake Detection Fine-Tuning Network}. In \bibinfo{booktitle}{\emph{ICT Systems Security and Privacy Protection}}, \bibfield{editor}{\bibinfo{person}{Marko H{\"o}lbl}, \bibinfo{person}{Kai Rannenberg}, {and} \bibinfo{person}{Tatjana Welzer}} (Eds.). \bibinfo{publisher}{Springer International Publishing}, \bibinfo{address}{Cham}, \bibinfo{pages}{416--430}.
\newblock


\bibitem[Jeong et~al\mbox{.}(2022)]%
        {Jeon2022FrePGAN}
\bibfield{author}{\bibinfo{person}{Yonghyun Jeong}, \bibinfo{person}{Doyeon Kim}, \bibinfo{person}{Youngmin Ro}, {and} \bibinfo{person}{Jongwon Choi}.} \bibinfo{year}{2022}\natexlab{}.
\newblock \showarticletitle{FrePGAN: Robust Deepfake Detection Using Frequency-Level Perturbations}.
\newblock \bibinfo{journal}{\emph{Proceedings of the AAAI Conference on Artificial Intelligence}} \bibinfo{volume}{36}, \bibinfo{number}{1} (\bibinfo{date}{Jun.} \bibinfo{year}{2022}), \bibinfo{pages}{1060--1068}.
\newblock
\urldef\tempurl%
\url{https://doi.org/10.1609/aaai.v36i1.19990}
\showDOI{\tempurl}


\bibitem[Jia et~al\mbox{.}(2022)]%
        {Exploring2022Jia_CVPR}
\bibfield{author}{\bibinfo{person}{Shuai Jia}, \bibinfo{person}{Chao Ma}, \bibinfo{person}{Taiping Yao}, \bibinfo{person}{Bangjie Yin}, \bibinfo{person}{Shouhong Ding}, {and} \bibinfo{person}{Xiaokang Yang}.} \bibinfo{year}{2022}\natexlab{}.
\newblock \showarticletitle{Exploring Frequency Adversarial Attacks for Face Forgery Detection}. In \bibinfo{booktitle}{\emph{{IEEE/CVF} Conference on Computer Vision and Pattern Recognition, {CVPR} 2022, New Orleans, LA, USA, June 18-24, 2022}}. \bibinfo{publisher}{{IEEE}}, \bibinfo{pages}{4093--4102}.
\newblock
\urldef\tempurl%
\url{https://doi.org/10.1109/CVPR52688.2022.00407}
\showDOI{\tempurl}


\bibitem[Jiang et~al\mbox{.}(2020)]%
        {jiang2020deeperforensics1}
\bibfield{author}{\bibinfo{person}{Liming Jiang}, \bibinfo{person}{Ren Li}, \bibinfo{person}{Wayne Wu}, \bibinfo{person}{Chen Qian}, {and} \bibinfo{person}{Chen~Change Loy}.} \bibinfo{year}{2020}\natexlab{}.
\newblock \showarticletitle{{DeeperForensics-1.0}: A Large-Scale Dataset for Real-World Face Forgery Detection}. In \bibinfo{booktitle}{\emph{CVPR}}. \bibinfo{pages}{2889--2898}.
\newblock


\bibitem[Jung et~al\mbox{.}(2020)]%
        {DeepVision2020Jung}
\bibfield{author}{\bibinfo{person}{Tackhyun Jung}, \bibinfo{person}{Sangwon Kim}, {and} \bibinfo{person}{Keecheon Kim}.} \bibinfo{year}{2020}\natexlab{}.
\newblock \showarticletitle{DeepVision: Deepfakes Detection Using Human Eye Blinking Pattern}.
\newblock \bibinfo{journal}{\emph{IEEE Access}}  \bibinfo{volume}{8} (\bibinfo{year}{2020}), \bibinfo{pages}{83144--83154}.
\newblock
\urldef\tempurl%
\url{https://doi.org/10.1109/ACCESS.2020.2988660}
\showDOI{\tempurl}


\bibitem[Kalman(1960)]%
        {Kalman1960Filtering}
\bibfield{author}{\bibinfo{person}{R.~E. Kalman}.} \bibinfo{year}{1960}\natexlab{}.
\newblock \showarticletitle{{A New Approach to Linear Filtering and Prediction Problems}}.
\newblock \bibinfo{journal}{\emph{Journal of Basic Engineering}} \bibinfo{volume}{82}, \bibinfo{number}{1} (\bibinfo{date}{03} \bibinfo{year}{1960}), \bibinfo{pages}{35--45}.
\newblock
\showISSN{0021-9223}
\urldef\tempurl%
\url{https://doi.org/10.1115/1.3662552}
\showDOI{\tempurl}


\bibitem[Kang et~al\mbox{.}(2022)]%
        {one2022kang}
\bibfield{author}{\bibinfo{person}{Wonjun Kang}, \bibinfo{person}{Geonsu Lee}, \bibinfo{person}{Hyung~Il Koo}, {and} \bibinfo{person}{Nam~Ik Cho}.} \bibinfo{year}{2022}\natexlab{}.
\newblock \bibinfo{title}{One-Shot Face Reenactment on Megapixels}.
\newblock
\newblock
\urldef\tempurl%
\url{https://doi.org/10.48550/ARXIV.2205.13368}
\showDOI{\tempurl}


\bibitem[Karras et~al\mbox{.}(2018)]%
        {karras2018progressive}
\bibfield{author}{\bibinfo{person}{Tero Karras}, \bibinfo{person}{Timo Aila}, \bibinfo{person}{Samuli Laine}, {and} \bibinfo{person}{Jaakko Lehtinen}.} \bibinfo{year}{2018}\natexlab{}.
\newblock \showarticletitle{Progressive Growing of {GAN}s for Improved Quality, Stability, and Variation}. In \bibinfo{booktitle}{\emph{International Conference on Learning Representations}}.
\newblock
\urldef\tempurl%
\url{https://openreview.net/forum?id=Hk99zCeAb}
\showURL{%
\tempurl}


\bibitem[Karras et~al\mbox{.}(2021)]%
        {stylegan2021karras}
\bibfield{author}{\bibinfo{person}{T. Karras}, \bibinfo{person}{S. Laine}, {and} \bibinfo{person}{T. Aila}.} \bibinfo{year}{2021}\natexlab{}.
\newblock \showarticletitle{A Style-Based Generator Architecture for Generative Adversarial Networks}.
\newblock \bibinfo{journal}{\emph{IEEE Transactions on Pattern Analysis and Machine Intelligence}} \bibinfo{volume}{43}, \bibinfo{number}{12} (\bibinfo{date}{dec} \bibinfo{year}{2021}), \bibinfo{pages}{4217--4228}.
\newblock
\showISSN{1939-3539}
\urldef\tempurl%
\url{https://doi.org/10.1109/TPAMI.2020.2970919}
\showDOI{\tempurl}


\bibitem[Katro(2022)]%
        {Katro2022BucksProb}
\bibfield{author}{\bibinfo{person}{Katie Katro}.} \bibinfo{year}{2022}\natexlab{}.
\newblock \bibinfo{title}{Bucks County mother gets probation in harassment case involving daughter's cheerleading rivals.}
\newblock \bibinfo{howpublished}{\url{https://6abc.com/raffaela-spone-bucks-county-pa-cheerleaders-harassment-case-victory-vipers-squad/11939419/}}.
\newblock
\newblock
\shownote{Accessed: 2022-09-08}.


\bibitem[Kelion(2018)]%
        {Kelion2018Emma}
\bibfield{author}{\bibinfo{person}{Leo Kelion}.} \bibinfo{year}{2018}\natexlab{}.
\newblock \bibinfo{title}{Deepfake porn videos deleted from internet by Gfycat}.
\newblock \bibinfo{howpublished}{\url{https://www.bbc.com/news/technology-42905185}}.
\newblock
\newblock
\shownote{Accessed: 2022-09-08}.


\bibitem[Kietzmann et~al\mbox{.}(2020)]%
        {Kietzmann2020TrickOrTreat}
\bibfield{author}{\bibinfo{person}{Jan Kietzmann}, \bibinfo{person}{Linda~W. Lee}, \bibinfo{person}{Ian~P. McCarthy}, {and} \bibinfo{person}{Tim~C. Kietzmann}.} \bibinfo{year}{2020}\natexlab{}.
\newblock \showarticletitle{Deepfakes: Trick or treat?}
\newblock \bibinfo{journal}{\emph{Business Horizons}} \bibinfo{volume}{63}, \bibinfo{number}{2} (\bibinfo{year}{2020}), \bibinfo{pages}{135--146}.
\newblock
\showISSN{0007-6813}
\urldef\tempurl%
\url{https://doi.org/10.1016/j.bushor.2019.11.006}
\showDOI{\tempurl}
\newblock
\shownote{ARTIFICIAL INTELLIGENCE AND MACHINE LEARNING}.


\bibitem[King(2021)]%
        {dlib}
\bibfield{author}{\bibinfo{person}{Davis King}.} \bibinfo{year}{2021}\natexlab{}.
\newblock \bibinfo{title}{{dlib 19.22.1}}.
\newblock \bibinfo{howpublished}{\url{https://pypi.org/project/dlib/}}.
\newblock
\newblock
\shownote{Accessed: 2021-08-29}.


\bibitem[Kingma and Welling(2014)]%
        {Kingma2014Autoencoder}
\bibfield{author}{\bibinfo{person}{Diederik~P. Kingma} {and} \bibinfo{person}{Max Welling}.} \bibinfo{year}{2014}\natexlab{}.
\newblock \showarticletitle{Auto-Encoding Variational Bayes}. In \bibinfo{booktitle}{\emph{2nd International Conference on Learning Representations, {ICLR} 2014, Banff, AB, Canada, April 14-16, 2014, Conference Track Proceedings}}, \bibfield{editor}{\bibinfo{person}{Yoshua Bengio} {and} \bibinfo{person}{Yann LeCun}} (Eds.).
\newblock


\bibitem[Koopman et~al\mbox{.}(2018)]%
        {Koopman2018detection}
\bibfield{author}{\bibinfo{person}{Marissa Koopman}, \bibinfo{person}{Andrea Macarulla~Rodriguez}, {and} \bibinfo{person}{Zeno Geradts}.} \bibinfo{year}{2018}\natexlab{}.
\newblock \showarticletitle{Detection of Deepfake Video Manipulation}. In \bibinfo{booktitle}{\emph{Proceedings of the 20th Irish Machine Vision and Image Processing conference}}. \bibinfo{pages}{133–136}.
\newblock


\bibitem[Korshunov and Marcel(2018)]%
        {DeepfakeTIMIT2018Korshunov}
\bibfield{author}{\bibinfo{person}{Pavel Korshunov} {and} \bibinfo{person}{S{\'{e}}bastien Marcel}.} \bibinfo{year}{2018}\natexlab{}.
\newblock \showarticletitle{DeepFakes: a New Threat to Face Recognition? Assessment and Detection}.
\newblock \bibinfo{journal}{\emph{CoRR}}  \bibinfo{volume}{abs/1812.08685} (\bibinfo{year}{2018}).
\newblock
\showeprint[arXiv]{1812.08685}
\urldef\tempurl%
\url{http://arxiv.org/abs/1812.08685}
\showURL{%
\tempurl}


\bibitem[Kumar et~al\mbox{.}(2020)]%
        {kumar2020detecting}
\bibfield{author}{\bibinfo{person}{Prabhat Kumar}, \bibinfo{person}{Mayank Vatsa}, {and} \bibinfo{person}{Richa Singh}.} \bibinfo{year}{2020}\natexlab{}.
\newblock \showarticletitle{Detecting Face2Face Facial Reenactment in Videos}. In \bibinfo{booktitle}{\emph{2020 IEEE Winter Conference on Applications of Computer Vision (WACV)}}. \bibinfo{pages}{2578--2586}.
\newblock
\urldef\tempurl%
\url{https://doi.org/10.1109/WACV45572.2020.9093628}
\showDOI{\tempurl}


\bibitem[Kwon et~al\mbox{.}(2021)]%
        {KoDF2021Kwon}
\bibfield{author}{\bibinfo{person}{Patrick Kwon}, \bibinfo{person}{Jaeseong You}, \bibinfo{person}{Gyuhyeon Nam}, \bibinfo{person}{Sungwoo Park}, {and} \bibinfo{person}{Gyeongsu Chae}.} \bibinfo{year}{2021}\natexlab{}.
\newblock \showarticletitle{KoDF: A Large-Scale Korean DeepFake Detection Dataset}. In \bibinfo{booktitle}{\emph{Proceedings of the IEEE/CVF International Conference on Computer Vision (ICCV)}}. \bibinfo{pages}{10744--10753}.
\newblock


\bibitem[Labs(2017)]%
        {faceswaplive}
\bibfield{author}{\bibinfo{person}{Laan Labs}.} \bibinfo{year}{2017}\natexlab{}.
\newblock \bibinfo{title}{Face Swap Live}.
\newblock \bibinfo{howpublished}{\url{https://play.google.com/store/apps/details?id=com.laan.labs.faceswaplive&hl=en&gl=US}}.
\newblock
\newblock
\shownote{Accessed: 2022-09-28}.


\bibitem[Lee(2018)]%
        {Lee2018Portman}
\bibfield{author}{\bibinfo{person}{Dave Lee}.} \bibinfo{year}{2018}\natexlab{}.
\newblock \bibinfo{title}{Deepfakes porn has serious consequences}.
\newblock \bibinfo{howpublished}{\url{https://www.bbc.com/news/technology-42912529}}.
\newblock
\newblock
\shownote{Accessed: 2022-09-08}.


\bibitem[Lempitsky et~al\mbox{.}(2018)]%
        {Lempitsky2018Deep}
\bibfield{author}{\bibinfo{person}{Victor Lempitsky}, \bibinfo{person}{Andrea Vedaldi}, {and} \bibinfo{person}{Dmitry Ulyanov}.} \bibinfo{year}{2018}\natexlab{}.
\newblock \showarticletitle{Deep Image Prior}. In \bibinfo{booktitle}{\emph{2018 IEEE/CVF Conference on Computer Vision and Pattern Recognition}}. \bibinfo{pages}{9446--9454}.
\newblock
\urldef\tempurl%
\url{https://doi.org/10.1109/CVPR.2018.00984}
\showDOI{\tempurl}


\bibitem[Li et~al\mbox{.}(2021)]%
        {FreqAware2021Li}
\bibfield{author}{\bibinfo{person}{Jiaming Li}, \bibinfo{person}{Hongtao Xie}, \bibinfo{person}{Jiahong Li}, \bibinfo{person}{Zhongyuan Wang}, {and} \bibinfo{person}{Yongdong Zhang}.} \bibinfo{year}{2021}\natexlab{}.
\newblock \showarticletitle{Frequency-Aware Discriminative Feature Learning Supervised by Single-Center Loss for Face Forgery Detection}. In \bibinfo{booktitle}{\emph{{IEEE} Conference on Computer Vision and Pattern Recognition, {CVPR} 2021, virtual, June 19-25, 2021}}. \bibinfo{publisher}{Computer Vision Foundation / {IEEE}}, \bibinfo{pages}{6458--6467}.
\newblock
\urldef\tempurl%
\url{https://doi.org/10.1109/CVPR46437.2021.00639}
\showDOI{\tempurl}


\bibitem[Li et~al\mbox{.}(2019)]%
        {Li2019FaceShifter}
\bibfield{author}{\bibinfo{person}{Lingzhi Li}, \bibinfo{person}{Jianmin Bao}, \bibinfo{person}{Hao Yang}, \bibinfo{person}{Dong Chen}, {and} \bibinfo{person}{Fang Wen}.} \bibinfo{year}{2019}\natexlab{}.
\newblock \showarticletitle{FaceShifter: Towards High Fidelity And Occlusion Aware Face Swapping}.
\newblock \bibinfo{journal}{\emph{arXiv preprint arXiv:1912.13457}} (\bibinfo{year}{2019}).
\newblock


\bibitem[Li et~al\mbox{.}(2020a)]%
        {Li2020face}
\bibfield{author}{\bibinfo{person}{Lingzhi Li}, \bibinfo{person}{Jianmin Bao}, \bibinfo{person}{Ting Zhang}, \bibinfo{person}{Hao Yang}, \bibinfo{person}{Dong Chen}, \bibinfo{person}{Fang Wen}, {and} \bibinfo{person}{Baining Guo}.} \bibinfo{year}{2020}\natexlab{a}.
\newblock \showarticletitle{Face X-Ray for More General Face Forgery Detection}. In \bibinfo{booktitle}{\emph{2020 IEEE/CVF Conference on Computer Vision and Pattern Recognition (CVPR)}}. \bibinfo{pages}{5000--5009}.
\newblock
\urldef\tempurl%
\url{https://doi.org/10.1109/CVPR42600.2020.00505}
\showDOI{\tempurl}


\bibitem[Li et~al\mbox{.}(2018)]%
        {InIctuOculi2018Li}
\bibfield{author}{\bibinfo{person}{Yuezun Li}, \bibinfo{person}{Ming-Ching Chang}, {and} \bibinfo{person}{Siwei Lyu}.} \bibinfo{year}{2018}\natexlab{}.
\newblock \showarticletitle{In Ictu Oculi: Exposing AI Created Fake Videos by Detecting Eye Blinking}. In \bibinfo{booktitle}{\emph{2018 IEEE International Workshop on Information Forensics and Security (WIFS)}}. \bibinfo{pages}{1--7}.
\newblock
\urldef\tempurl%
\url{https://doi.org/10.1109/WIFS.2018.8630787}
\showDOI{\tempurl}


\bibitem[Li and Lyu(2019)]%
        {li2019exposing}
\bibfield{author}{\bibinfo{person}{Yuezun Li} {and} \bibinfo{person}{Siwei Lyu}.} \bibinfo{year}{2019}\natexlab{}.
\newblock \showarticletitle{Exposing DeepFake Videos By Detecting Face Warping Artifacts}. In \bibinfo{booktitle}{\emph{IEEE Conference on Computer Vision and Pattern Recognition Workshops (CVPRW)}}.
\newblock


\bibitem[Li et~al\mbox{.}(2020b)]%
        {Li2020CelebDF}
\bibfield{author}{\bibinfo{person}{Yuezun Li}, \bibinfo{person}{Xin Yang}, \bibinfo{person}{Pu Sun}, \bibinfo{person}{Honggang Qi}, {and} \bibinfo{person}{Siwei Lyu}.} \bibinfo{year}{2020}\natexlab{b}.
\newblock \showarticletitle{Celeb-DF: A Large-Scale Challenging Dataset for DeepFake Forensics}. In \bibinfo{booktitle}{\emph{2020 IEEE/CVF Conference on Computer Vision and Pattern Recognition (CVPR)}}. \bibinfo{pages}{3204--3213}.
\newblock
\urldef\tempurl%
\url{https://doi.org/10.1109/CVPR42600.2020.00327}
\showDOI{\tempurl}


\bibitem[Liang et~al\mbox{.}(2022)]%
        {Exploring2022Liang_ECCV}
\bibfield{author}{\bibinfo{person}{Jiahao Liang}, \bibinfo{person}{Huafeng Shi}, {and} \bibinfo{person}{Weihong Deng}.} \bibinfo{year}{2022}\natexlab{}.
\newblock \showarticletitle{Exploring Disentangled Content Information for Face Forgery Detection}. In \bibinfo{booktitle}{\emph{Computer Vision -- ECCV 2022}}, \bibfield{editor}{\bibinfo{person}{Shai Avidan}, \bibinfo{person}{Gabriel Brostow}, \bibinfo{person}{Moustapha Ciss{\'e}}, \bibinfo{person}{Giovanni~Maria Farinella}, {and} \bibinfo{person}{Tal Hassner}} (Eds.). \bibinfo{publisher}{Springer Nature Switzerland}, \bibinfo{address}{Cham}, \bibinfo{pages}{128--145}.
\newblock
\showISBNx{978-3-031-19781-9}


\bibitem[Lin et~al\mbox{.}(2022)]%
        {Lin2022Exploiting}
\bibfield{author}{\bibinfo{person}{Dongdong Lin}, \bibinfo{person}{Benedetta Tondi}, \bibinfo{person}{Bin Li}, {and} \bibinfo{person}{Mauro Barni}.} \bibinfo{year}{2022}\natexlab{}.
\newblock \showarticletitle{Exploiting temporal information to prevent the transferability of adversarial examples against deep fake detectors}. In \bibinfo{booktitle}{\emph{2022 IEEE International Joint Conference on Biometrics (IJCB)}}. \bibinfo{pages}{1--8}.
\newblock
\urldef\tempurl%
\url{https://doi.org/10.1109/IJCB54206.2022.10007959}
\showDOI{\tempurl}


\bibitem[Liu et~al\mbox{.}(2021a)]%
        {SpatialPhase2021Liu}
\bibfield{author}{\bibinfo{person}{Honggu Liu}, \bibinfo{person}{Xiaodan Li}, \bibinfo{person}{Wenbo Zhou}, \bibinfo{person}{Yuefeng Chen}, \bibinfo{person}{Yuan He}, \bibinfo{person}{Hui Xue}, \bibinfo{person}{Weiming Zhang}, {and} \bibinfo{person}{Nenghai Yu}.} \bibinfo{year}{2021}\natexlab{a}.
\newblock \showarticletitle{Spatial-Phase Shallow Learning: Rethinking Face Forgery Detection in Frequency Domain}. In \bibinfo{booktitle}{\emph{{IEEE} Conference on Computer Vision and Pattern Recognition, {CVPR} 2021, virtual, June 19-25, 2021}}. \bibinfo{publisher}{Computer Vision Foundation / {IEEE}}, \bibinfo{pages}{772--781}.
\newblock
\urldef\tempurl%
\url{https://doi.org/10.1109/CVPR46437.2021.00083}
\showDOI{\tempurl}


\bibitem[Liu et~al\mbox{.}(2021b)]%
        {liu2021Swin}
\bibfield{author}{\bibinfo{person}{Ze Liu}, \bibinfo{person}{Yutong Lin}, \bibinfo{person}{Yue Cao}, \bibinfo{person}{Han Hu}, \bibinfo{person}{Yixuan Wei}, \bibinfo{person}{Zheng Zhang}, \bibinfo{person}{Stephen Lin}, {and} \bibinfo{person}{Baining Guo}.} \bibinfo{year}{2021}\natexlab{b}.
\newblock \showarticletitle{Swin Transformer: Hierarchical Vision Transformer using Shifted Windows}. In \bibinfo{booktitle}{\emph{Proceedings of the IEEE/CVF International Conference on Computer Vision (ICCV)}}.
\newblock


\bibitem[Loyola-González(2019)]%
        {Black-Box2019Loyola}
\bibfield{author}{\bibinfo{person}{Octavio Loyola-González}.} \bibinfo{year}{2019}\natexlab{}.
\newblock \showarticletitle{Black-Box vs. White-Box: Understanding Their Advantages and Weaknesses From a Practical Point of View}.
\newblock \bibinfo{journal}{\emph{IEEE Access}}  \bibinfo{volume}{7} (\bibinfo{year}{2019}), \bibinfo{pages}{154096--154113}.
\newblock
\urldef\tempurl%
\url{https://doi.org/10.1109/ACCESS.2019.2949286}
\showDOI{\tempurl}


\bibitem[Ltd(2017)]%
        {faceapp}
\bibfield{author}{\bibinfo{person}{FaceApp~Technology Ltd}.} \bibinfo{year}{2017}\natexlab{}.
\newblock \bibinfo{title}{FaceApp: Face Editor}.
\newblock \bibinfo{howpublished}{\url{https://play.google.com/store/apps/details?id=io.faceapp&hl=en&gl=US}}.
\newblock
\newblock
\shownote{Accessed: 2022-09-28}.


\bibitem[Lu(2018)]%
        {faceswap-GAN}
\bibfield{author}{\bibinfo{person}{Shao-An Lu}.} \bibinfo{year}{2018}\natexlab{}.
\newblock \bibinfo{title}{faceswap-GAN}.
\newblock \bibinfo{howpublished}{\url{https://github.com/shaoanlu/faceswap-GAN}}.
\newblock
\newblock
\shownote{Accessed: 2022-09-08}.


\bibitem[Lukas et~al\mbox{.}(2006)]%
        {Lukas2006Digital}
\bibfield{author}{\bibinfo{person}{J. Lukas}, \bibinfo{person}{J. Fridrich}, {and} \bibinfo{person}{M. Goljan}.} \bibinfo{year}{2006}\natexlab{}.
\newblock \showarticletitle{Digital camera identification from sensor pattern noise}.
\newblock \bibinfo{journal}{\emph{IEEE Transactions on Information Forensics and Security}} \bibinfo{volume}{1}, \bibinfo{number}{2} (\bibinfo{year}{2006}), \bibinfo{pages}{205--214}.
\newblock
\urldef\tempurl%
\url{https://doi.org/10.1109/TIFS.2006.873602}
\showDOI{\tempurl}


\bibitem[Luo et~al\mbox{.}(2021)]%
        {Luo2021Generalizing}
\bibfield{author}{\bibinfo{person}{Yuchen Luo}, \bibinfo{person}{Yong Zhang}, \bibinfo{person}{Junchi Yan}, {and} \bibinfo{person}{Wei Liu}.} \bibinfo{year}{2021}\natexlab{}.
\newblock \showarticletitle{Generalizing Face Forgery Detection With High-Frequency Features}. In \bibinfo{booktitle}{\emph{IEEE Conference on Computer Vision and Patten Recognition (CVPR)}}. \bibinfo{pages}{16317--16326}.
\newblock


\bibitem[Marcon et~al\mbox{.}(2021)]%
        {Marcon2021Detection}
\bibfield{author}{\bibinfo{person}{Federico Marcon}, \bibinfo{person}{Cecilia Pasquini}, {and} \bibinfo{person}{Giulia Boato}.} \bibinfo{year}{2021}\natexlab{}.
\newblock \showarticletitle{Detection of Manipulated Face Videos over Social Networks: A Large-Scale Study}.
\newblock \bibinfo{journal}{\emph{Journal of Imaging}} \bibinfo{volume}{7}, \bibinfo{number}{10} (\bibinfo{year}{2021}).
\newblock
\showISSN{2313-433X}
\urldef\tempurl%
\url{https://doi.org/10.3390/jimaging7100193}
\showDOI{\tempurl}


\bibitem[MarekKowalski(2019)]%
        {faceswap}
\bibfield{author}{\bibinfo{person}{Marek MarekKowalski}.} \bibinfo{year}{2019}\natexlab{}.
\newblock \bibinfo{title}{{FaceSwap}}.
\newblock \bibinfo{howpublished}{\url{https://github.com/MarekKowalski/FaceSwap}}.
\newblock
\newblock
\shownote{Accessed: 2021-08-29}.


\bibitem[Marra et~al\mbox{.}(2017)]%
        {Marra2017blind}
\bibfield{author}{\bibinfo{person}{Francesco Marra}, \bibinfo{person}{Giovanni Poggi}, \bibinfo{person}{Carlo Sansone}, {and} \bibinfo{person}{Luisa Verdoliva}.} \bibinfo{year}{2017}\natexlab{}.
\newblock \showarticletitle{Blind PRNU-Based Image Clustering for Source Identification}.
\newblock \bibinfo{journal}{\emph{IEEE Transactions on Information Forensics and Security}} \bibinfo{volume}{12}, \bibinfo{number}{9} (\bibinfo{year}{2017}), \bibinfo{pages}{2197--2211}.
\newblock
\urldef\tempurl%
\url{https://doi.org/10.1109/TIFS.2017.2701335}
\showDOI{\tempurl}


\bibitem[Martino et~al\mbox{.}(2018)]%
        {randsample2018Martino}
\bibfield{author}{\bibinfo{person}{Luca Martino}, \bibinfo{person}{David Luengo}, {and} \bibinfo{person}{Joaqu{\'i}n M{\'i}guez}.} \bibinfo{year}{2018}\natexlab{}.
\newblock \bibinfo{booktitle}{\emph{Direct Methods}}.
\newblock \bibinfo{publisher}{Springer International Publishing}, \bibinfo{address}{Cham}, \bibinfo{pages}{27--63}.
\newblock
\showISBNx{978-3-319-72634-2}
\urldef\tempurl%
\url{https://doi.org/10.1007/978-3-319-72634-2\_2}
\showDOI{\tempurl}


\bibitem[Masi et~al\mbox{.}(2020)]%
        {Masi2020Two-Branch}
\bibfield{author}{\bibinfo{person}{Iacopo Masi}, \bibinfo{person}{Aditya Killekar}, \bibinfo{person}{Royston~Marian Mascarenhas}, \bibinfo{person}{Shenoy~Pratik Gurudatt}, {and} \bibinfo{person}{Wael AbdAlmageed}.} \bibinfo{year}{2020}\natexlab{}.
\newblock \showarticletitle{Two-Branch Recurrent Network for Isolating Deepfakes in Videos}. In \bibinfo{booktitle}{\emph{Computer Vision -- ECCV 2020}}, \bibfield{editor}{\bibinfo{person}{Andrea Vedaldi}, \bibinfo{person}{Horst Bischof}, \bibinfo{person}{Thomas Brox}, {and} \bibinfo{person}{Jan-Michael Frahm}} (Eds.). \bibinfo{publisher}{Springer International Publishing}, \bibinfo{address}{Cham}, \bibinfo{pages}{667--684}.
\newblock
\showISBNx{978-3-030-58571-6}


\bibitem[Matern et~al\mbox{.}(2019)]%
        {VisualArtif2019Matern}
\bibfield{author}{\bibinfo{person}{Falko Matern}, \bibinfo{person}{Christian Riess}, {and} \bibinfo{person}{Marc Stamminger}.} \bibinfo{year}{2019}\natexlab{}.
\newblock \showarticletitle{Exploiting Visual Artifacts to Expose Deepfakes and Face Manipulations}. In \bibinfo{booktitle}{\emph{2019 IEEE Winter Applications of Computer Vision Workshops (WACVW)}}. \bibinfo{pages}{83--92}.
\newblock
\urldef\tempurl%
\url{https://doi.org/10.1109/WACVW.2019.00020}
\showDOI{\tempurl}


\bibitem[Miller(2022)]%
        {Miller2022Zelensky}
\bibfield{author}{\bibinfo{person}{Joshua~Rhett Miller}.} \bibinfo{year}{2022}\natexlab{}.
\newblock \bibinfo{title}{Deepfake video of Zelensky telling Ukrainians to surrender removed from social platforms}.
\newblock \bibinfo{howpublished}{\url{https://nypost.com/2022/03/17/deepfake-video-shows-volodymyr-zelensky-telling-ukrainians-to-surrender/}}.
\newblock
\newblock
\shownote{Accessed: 2022-09-08}.


\bibitem[Mirsky and Lee(2021)]%
        {Creation2021Mirsky}
\bibfield{author}{\bibinfo{person}{Yisroel Mirsky} {and} \bibinfo{person}{Wenke Lee}.} \bibinfo{year}{2021}\natexlab{}.
\newblock \showarticletitle{The Creation and Detection of Deepfakes: A Survey}.
\newblock \bibinfo{journal}{\emph{ACM Comput. Surv.}} \bibinfo{volume}{54}, \bibinfo{number}{1}, Article \bibinfo{articleno}{7} (\bibinfo{date}{jan} \bibinfo{year}{2021}), \bibinfo{numpages}{41}~pages.
\newblock
\showISSN{0360-0300}
\urldef\tempurl%
\url{https://doi.org/10.1145/3425780}
\showDOI{\tempurl}


\bibitem[(Momo)(2019)]%
        {ZAO2019}
\bibfield{author}{\bibinfo{person}{Hello Group~Inc. (Momo)}.} \bibinfo{year}{2019}\natexlab{}.
\newblock \bibinfo{title}{ZAO}.
\newblock \bibinfo{howpublished}{\url{https://apps.apple.com/cn/app/id1465199127}}.
\newblock
\newblock
\shownote{Accessed: 2022-09-08}.


\bibitem[Natsume et~al\mbox{.}(2018)]%
        {Natsume2018RSGAN}
\bibfield{author}{\bibinfo{person}{Ryota Natsume}, \bibinfo{person}{Tatsuya Yatagawa}, {and} \bibinfo{person}{Shigeo Morishima}.} \bibinfo{year}{2018}\natexlab{}.
\newblock \showarticletitle{RSGAN: Face Swapping and Editing Using Face and Hair Representation in Latent Spaces}. In \bibinfo{booktitle}{\emph{ACM SIGGRAPH 2018 Posters}} (Vancouver, British Columbia, Canada) \emph{(\bibinfo{series}{SIGGRAPH '18})}. \bibinfo{publisher}{Association for Computing Machinery}, \bibinfo{address}{New York, NY, USA}, Article \bibinfo{articleno}{69}, \bibinfo{numpages}{2}~pages.
\newblock
\showISBNx{9781450358170}
\urldef\tempurl%
\url{https://doi.org/10.1145/3230744.3230818}
\showDOI{\tempurl}


\bibitem[Nguyen et~al\mbox{.}(2019a)]%
        {Nguyen2019MultitaskLF}
\bibfield{author}{\bibinfo{person}{Huy~Hoang Nguyen}, \bibinfo{person}{Fuming Fang}, \bibinfo{person}{Junichi Yamagishi}, {and} \bibinfo{person}{Isao Echizen}.} \bibinfo{year}{2019}\natexlab{a}.
\newblock \showarticletitle{Multi-task Learning for Detecting and Segmenting Manipulated Facial Images and Videos}.
\newblock \bibinfo{journal}{\emph{2019 IEEE 10th International Conference on Biometrics Theory, Applications and Systems (BTAS)}} (\bibinfo{year}{2019}), \bibinfo{pages}{1--8}.
\newblock


\bibitem[Nguyen et~al\mbox{.}(2019b)]%
        {nguyen2019use}
\bibfield{author}{\bibinfo{person}{Huy~H. Nguyen}, \bibinfo{person}{Junichi Yamagishi}, {and} \bibinfo{person}{Isao Echizen}.} \bibinfo{year}{2019}\natexlab{b}.
\newblock \bibinfo{title}{Use of a Capsule Network to Detect Fake Images and Videos}.
\newblock
\newblock
\showeprint[arxiv]{1910.12467}~[cs.CV]


\bibitem[Nightingale et~al\mbox{.}(2021)]%
        {sophie2022synthetic}
\bibfield{author}{\bibinfo{person}{Sophie~J. Nightingale}, \bibinfo{person}{Shruti Agarwal}, \bibinfo{person}{Erik Härkönen}, \bibinfo{person}{Jaakko Lehtinen}, {and} \bibinfo{person}{Hany Farid}.} \bibinfo{year}{2021}\natexlab{}.
\newblock \showarticletitle{Synthetic faces: how perceptually convincing are they?}
\newblock \bibinfo{journal}{\emph{Journal of Vision}} \bibinfo{volume}{21}, \bibinfo{number}{9} (\bibinfo{year}{2021}), \bibinfo{pages}{2015}.
\newblock
\urldef\tempurl%
\url{https://doi.org/10.1167/jov.21.9.2015}
\showDOI{\tempurl}


\bibitem[Nightingale and Farid(2022a)]%
        {hany2022ai-synthesized}
\bibfield{author}{\bibinfo{person}{Sophie~J. Nightingale} {and} \bibinfo{person}{Hany Farid}.} \bibinfo{year}{2022}\natexlab{a}.
\newblock \showarticletitle{AI-synthesized faces are indistinguishable from real faces and more trustworthy}.
\newblock \bibinfo{journal}{\emph{Proceedings of the National Academy of Sciences}} \bibinfo{volume}{119}, \bibinfo{number}{8} (\bibinfo{year}{2022}), \bibinfo{pages}{e2120481119}.
\newblock
\urldef\tempurl%
\url{https://doi.org/10.1073/pnas.2120481119}
\showDOI{\tempurl}


\bibitem[Nightingale and Farid(2022b)]%
        {sophie2022synthesized}
\bibfield{author}{\bibinfo{person}{Sophie~J. Nightingale} {and} \bibinfo{person}{Hany Farid}.} \bibinfo{year}{2022}\natexlab{b}.
\newblock \showarticletitle{Synthetic Faces Are More Trustworthy Than Real Faces}.
\newblock \bibinfo{journal}{\emph{Proceedings of the National Academy of Sciences}} \bibinfo{volume}{22}, \bibinfo{number}{14} (\bibinfo{year}{2022}), \bibinfo{pages}{3068}.
\newblock
\urldef\tempurl%
\url{https://doi.org/10.1167/jov.22.14.3068}
\showDOI{\tempurl}


\bibitem[Nirkin et~al\mbox{.}(2019)]%
        {nirkin2019fsgan}
\bibfield{author}{\bibinfo{person}{Yuval Nirkin}, \bibinfo{person}{Yosi Keller}, {and} \bibinfo{person}{Tal Hassner}.} \bibinfo{year}{2019}\natexlab{}.
\newblock \showarticletitle{{FSGAN}: Subject agnostic face swapping and reenactment}. In \bibinfo{booktitle}{\emph{Proceedings of the IEEE International Conference on Computer Vision}}. \bibinfo{pages}{7184--7193}.
\newblock


\bibitem[Nirkin et~al\mbox{.}(2022)]%
        {Nirkin2022DeepFake}
\bibfield{author}{\bibinfo{person}{Yuval Nirkin}, \bibinfo{person}{Lior Wolf}, \bibinfo{person}{Yosi Keller}, {and} \bibinfo{person}{Tal Hassner}.} \bibinfo{year}{2022}\natexlab{}.
\newblock \showarticletitle{DeepFake Detection Based on Discrepancies Between Faces and Their Context}.
\newblock \bibinfo{journal}{\emph{IEEE Transactions on Pattern Analysis and Machine Intelligence}} \bibinfo{volume}{44}, \bibinfo{number}{10} (\bibinfo{year}{2022}), \bibinfo{pages}{6111--6121}.
\newblock
\urldef\tempurl%
\url{https://doi.org/10.1109/TPAMI.2021.3093446}
\showDOI{\tempurl}


\bibitem[Organization(2022)]%
        {WHO2022}
\bibfield{author}{\bibinfo{person}{World~Health Organization}.} \bibinfo{year}{2022}\natexlab{}.
\newblock \bibinfo{title}{Use of SARS-CoV-2 antigen-detection rapid diagnostic tests for COVID-19 self-testing}.
\newblock \bibinfo{howpublished}{\url{https://apps.who.int/iris/bitstream/handle/10665/352350/WHO-2019-nCoV-Ag-RDTs-Self-testing-2022.1-eng.pdf?sequence=1}}.
\newblock
\newblock
\shownote{Accessed: 2022-09-09}.


\bibitem[Osterman et~al\mbox{.}(2021)]%
        {Osterman2021Antigen}
\bibfield{author}{\bibinfo{person}{Andreas Osterman}, \bibinfo{person}{Maximilian Iglhaut}, \bibinfo{person}{Andreas Lehner}, \bibinfo{person}{Patricia Sp{\"a}th}, \bibinfo{person}{Marcel Stern}, \bibinfo{person}{Hanna Autenrieth}, \bibinfo{person}{Maximilian Muenchhoff}, \bibinfo{person}{Alexander Graf}, \bibinfo{person}{Stefan Krebs}, \bibinfo{person}{Helmut Blum}, \bibinfo{person}{Armin Baiker}, \bibinfo{person}{Natascha Grzimek-Koschewa}, \bibinfo{person}{Ulrike Protzer}, \bibinfo{person}{Lars Kaderali}, \bibinfo{person}{Hanna-Mari Baldauf}, {and} \bibinfo{person}{Oliver~T. Keppler}.} \bibinfo{year}{2021}\natexlab{}.
\newblock \showarticletitle{Comparison of four commercial, automated antigen tests to detect SARS-CoV-2 variants of concern}.
\newblock \bibinfo{journal}{\emph{Medical Microbiology and Immunology}} \bibinfo{volume}{210}, \bibinfo{number}{5} (\bibinfo{date}{01 Dec} \bibinfo{year}{2021}), \bibinfo{pages}{263--275}.
\newblock
\showISSN{1432-1831}
\urldef\tempurl%
\url{https://doi.org/10.1007/s00430-021-00719-0}
\showDOI{\tempurl}


\bibitem[Pan et~al\mbox{.}(2012)]%
        {Pan2012Exposing}
\bibfield{author}{\bibinfo{person}{Xunyu Pan}, \bibinfo{person}{Xing Zhang}, {and} \bibinfo{person}{Siwei Lyu}.} \bibinfo{year}{2012}\natexlab{}.
\newblock \showarticletitle{Exposing image splicing with inconsistent local noise variances}. In \bibinfo{booktitle}{\emph{2012 IEEE International Conference on Computational Photography (ICCP)}}. \bibinfo{pages}{1--10}.
\newblock
\urldef\tempurl%
\url{https://doi.org/10.1109/ICCPhot.2012.6215223}
\showDOI{\tempurl}


\bibitem[Park et~al\mbox{.}(2019)]%
        {gaugan2019park}
\bibfield{author}{\bibinfo{person}{Taesung Park}, \bibinfo{person}{Ming-Yu Liu}, \bibinfo{person}{Ting-Chun Wang}, {and} \bibinfo{person}{Jun-Yan Zhu}.} \bibinfo{year}{2019}\natexlab{}.
\newblock \showarticletitle{Semantic Image Synthesis With Spatially-Adaptive Normalization}. In \bibinfo{booktitle}{\emph{2019 IEEE/CVF Conference on Computer Vision and Pattern Recognition (CVPR)}}. \bibinfo{pages}{2332--2341}.
\newblock
\urldef\tempurl%
\url{https://doi.org/10.1109/CVPR.2019.00244}
\showDOI{\tempurl}


\bibitem[Peng et~al\mbox{.}(2017)]%
        {Peng2017Optimized3L}
\bibfield{author}{\bibinfo{person}{Bo Peng}, \bibinfo{person}{Wei Wang}, \bibinfo{person}{Jing Dong}, {and} \bibinfo{person}{Tieniu Tan}.} \bibinfo{year}{2017}\natexlab{}.
\newblock \showarticletitle{Optimized 3D Lighting Environment Estimation for Image Forgery Detection}.
\newblock \bibinfo{journal}{\emph{IEEE Transactions on Information Forensics and Security}}  \bibinfo{volume}{12} (\bibinfo{year}{2017}), \bibinfo{pages}{479--494}.
\newblock


\bibitem[Peng et~al\mbox{.}(2021)]%
        {Peng2021Conformer}
\bibfield{author}{\bibinfo{person}{Z. Peng}, \bibinfo{person}{W. Huang}, \bibinfo{person}{S. Gu}, \bibinfo{person}{L. Xie}, \bibinfo{person}{Y. Wang}, \bibinfo{person}{J. Jiao}, {and} \bibinfo{person}{Q. Ye}.} \bibinfo{year}{2021}\natexlab{}.
\newblock \showarticletitle{Conformer: Local Features Coupling Global Representations for Visual Recognition}. In \bibinfo{booktitle}{\emph{2021 IEEE/CVF International Conference on Computer Vision (ICCV)}}. \bibinfo{publisher}{IEEE Computer Society}, \bibinfo{address}{Los Alamitos, CA, USA}, \bibinfo{pages}{357--366}.
\newblock
\urldef\tempurl%
\url{https://doi.org/10.1109/ICCV48922.2021.00042}
\showDOI{\tempurl}


\bibitem[Perov et~al\mbox{.}(2021)]%
        {perov2021deepfacelab}
\bibfield{author}{\bibinfo{person}{Ivan Perov}, \bibinfo{person}{Daiheng Gao}, \bibinfo{person}{Nikolay Chervoniy}, \bibinfo{person}{Kunlin Liu}, \bibinfo{person}{Sugasa Marangonda}, \bibinfo{person}{Chris Umé}, \bibinfo{person}{Mr. Dpfks}, \bibinfo{person}{Carl~Shift Facenheim}, \bibinfo{person}{Luis RP}, \bibinfo{person}{Jian Jiang}, \bibinfo{person}{Sheng Zhang}, \bibinfo{person}{Pingyu Wu}, \bibinfo{person}{Bo Zhou}, {and} \bibinfo{person}{Weiming Zhang}.} \bibinfo{year}{2021}\natexlab{}.
\newblock \bibinfo{title}{DeepFaceLab: Integrated, flexible and extensible face-swapping framework}.
\newblock
\newblock
\showeprint[arxiv]{2005.05535}~[cs.CV]


\bibitem[Picetti et~al\mbox{.}(2020)]%
        {picetti2020dippas}
\bibfield{author}{\bibinfo{person}{Francesco Picetti}, \bibinfo{person}{Sara Mandelli}, \bibinfo{person}{Paolo Bestagini}, \bibinfo{person}{Vincenzo Lipari}, {and} \bibinfo{person}{Stefano Tubaro}.} \bibinfo{year}{2020}\natexlab{}.
\newblock \bibinfo{title}{DIPPAS: A Deep Image Prior PRNU Anonymization Scheme}.
\newblock
\newblock
\showeprint[arxiv]{2012.03581}~[cs.MM]


\bibitem[Polyak et~al\mbox{.}(2019)]%
        {audioswap2019Adam}
\bibfield{author}{\bibinfo{person}{Adam Polyak}, \bibinfo{person}{Lior Wolf}, {and} \bibinfo{person}{Yaniv Taigman}.} \bibinfo{year}{2019}\natexlab{}.
\newblock \bibinfo{title}{TTS Skins: Speaker Conversion via ASR}.
\newblock
\newblock
\urldef\tempurl%
\url{https://doi.org/10.48550/ARXIV.1904.08983}
\showDOI{\tempurl}


\bibitem[Qi et~al\mbox{.}(2020)]%
        {Qi2020DeepRhythm}
\bibfield{author}{\bibinfo{person}{Hua Qi}, \bibinfo{person}{Qing Guo}, \bibinfo{person}{Felix Juefei-Xu}, \bibinfo{person}{Xiaofei Xie}, \bibinfo{person}{Lei Ma}, \bibinfo{person}{Wei Feng}, \bibinfo{person}{Yang Liu}, {and} \bibinfo{person}{Jianjun Zhao}.} \bibinfo{year}{2020}\natexlab{}.
\newblock \showarticletitle{DeepRhythm: Exposing DeepFakes with Attentional Visual Heartbeat Rhythms}. In \bibinfo{booktitle}{\emph{Proceedings of the 28th ACM International Conference on Multimedia}} (Seattle, WA, USA) \emph{(\bibinfo{series}{MM '20})}. \bibinfo{publisher}{Association for Computing Machinery}, \bibinfo{address}{New York, NY, USA}, \bibinfo{pages}{4318–4327}.
\newblock
\showISBNx{9781450379885}
\urldef\tempurl%
\url{https://doi.org/10.1145/3394171.3413707}
\showDOI{\tempurl}


\bibitem[Qian et~al\mbox{.}(2020)]%
        {Qian2020Thinking}
\bibfield{author}{\bibinfo{person}{Yuyang Qian}, \bibinfo{person}{Guojun Yin}, \bibinfo{person}{Lu Sheng}, \bibinfo{person}{Zixuan Chen}, {and} \bibinfo{person}{Jing Shao}.} \bibinfo{year}{2020}\natexlab{}.
\newblock \showarticletitle{Thinking in Frequency: Face Forgery Detection by Mining Frequency-Aware Clues}. In \bibinfo{booktitle}{\emph{Computer Vision -- ECCV 2020}}, \bibfield{editor}{\bibinfo{person}{Andrea Vedaldi}, \bibinfo{person}{Horst Bischof}, \bibinfo{person}{Thomas Brox}, {and} \bibinfo{person}{Jan-Michael Frahm}} (Eds.). \bibinfo{publisher}{Springer International Publishing}, \bibinfo{address}{Cham}, \bibinfo{pages}{86--103}.
\newblock
\showISBNx{978-3-030-58610-2}


\bibitem[Reinhard et~al\mbox{.}(2001)]%
        {Reinhard2001Color}
\bibfield{author}{\bibinfo{person}{E. Reinhard}, \bibinfo{person}{M. Adhikhmin}, \bibinfo{person}{B. Gooch}, {and} \bibinfo{person}{P. Shirley}.} \bibinfo{year}{2001}\natexlab{}.
\newblock \showarticletitle{Color transfer between images}.
\newblock \bibinfo{journal}{\emph{IEEE Computer Graphics and Applications}} \bibinfo{volume}{21}, \bibinfo{number}{5} (\bibinfo{year}{2001}), \bibinfo{pages}{34--41}.
\newblock
\urldef\tempurl%
\url{https://doi.org/10.1109/38.946629}
\showDOI{\tempurl}


\bibitem[revise(2019)]%
        {Deepfakes2019Learn}
\bibfield{author}{\bibinfo{person}{Learn~\& revise}.} \bibinfo{year}{2019}\natexlab{}.
\newblock \bibinfo{title}{Deepfakes: What are they and why would I make one?}
\newblock \bibinfo{howpublished}{\url{https://www.bbc.co.uk/bitesize/articles/zfkwcqt}}.
\newblock
\newblock
\shownote{Accessed: 2022-09-08}.


\bibitem[Rossler et~al\mbox{.}(2019)]%
        {Rossler2019FaceForensics}
\bibfield{author}{\bibinfo{person}{Andreas Rossler}, \bibinfo{person}{Davide Cozzolino}, \bibinfo{person}{Luisa Verdoliva}, \bibinfo{person}{Christian Riess}, \bibinfo{person}{Justus Thies}, {and} \bibinfo{person}{Matthias Niessner}.} \bibinfo{year}{2019}\natexlab{}.
\newblock \showarticletitle{FaceForensics++: Learning to Detect Manipulated Facial Images}. In \bibinfo{booktitle}{\emph{Proceedings of the IEEE/CVF International Conference on Computer Vision (ICCV)}}. \bibinfo{pages}{1--11}.
\newblock


\bibitem[Sabir et~al\mbox{.}(2019)]%
        {Sabir2019RecurrentCS}
\bibfield{author}{\bibinfo{person}{Ekraam Sabir}, \bibinfo{person}{Jiaxin Cheng}, \bibinfo{person}{Ayush Jaiswal}, \bibinfo{person}{Wael AbdAlmageed}, \bibinfo{person}{Iacopo Masi}, {and} \bibinfo{person}{P. Natarajan}.} \bibinfo{year}{2019}\natexlab{}.
\newblock \showarticletitle{Recurrent Convolutional Strategies for Face Manipulation Detection in Videos}. In \bibinfo{booktitle}{\emph{CVPR Workshops}}.
\newblock


\bibitem[Sabour et~al\mbox{.}(2017)]%
        {Sabour2017dynamic}
\bibfield{author}{\bibinfo{person}{Sara Sabour}, \bibinfo{person}{Nicholas Frosst}, {and} \bibinfo{person}{Geoffrey~E. Hinton}.} \bibinfo{year}{2017}\natexlab{}.
\newblock \showarticletitle{Dynamic Routing between Capsules}. In \bibinfo{booktitle}{\emph{Proceedings of the 31st International Conference on Neural Information Processing Systems}} (Long Beach, California, USA) \emph{(\bibinfo{series}{NIPS'17})}. \bibinfo{publisher}{Curran Associates Inc.}, \bibinfo{address}{Red Hook, NY, USA}, \bibinfo{pages}{3859–3869}.
\newblock
\showISBNx{9781510860964}


\bibitem[Saito et~al\mbox{.}(2017)]%
        {Saito2017ATheoretical}
\bibfield{author}{\bibinfo{person}{Shota Saito}, \bibinfo{person}{Yoichi Tomioka}, {and} \bibinfo{person}{Hitoshi Kitazawa}.} \bibinfo{year}{2017}\natexlab{}.
\newblock \showarticletitle{A Theoretical Framework for Estimating False Acceptance Rate of PRNU-Based Camera Identification}.
\newblock \bibinfo{journal}{\emph{IEEE Transactions on Information Forensics and Security}} \bibinfo{volume}{12}, \bibinfo{number}{9} (\bibinfo{year}{2017}), \bibinfo{pages}{2026--2035}.
\newblock
\urldef\tempurl%
\url{https://doi.org/10.1109/TIFS.2017.2692683}
\showDOI{\tempurl}


\bibitem[ScienceDaily(2020)]%
        {ScienceDaily2020}
\bibfield{author}{\bibinfo{person}{ScienceDaily}.} \bibinfo{year}{2020}\natexlab{}.
\newblock \bibinfo{title}{`Deepfakes' ranked as most serious AI crime threat}.
\newblock \bibinfo{howpublished}{\url{https://www.sciencedaily.com/releases/2020/08/200804085908.htm}}.
\newblock
\newblock
\shownote{Accessed: 2021-05-01}.


\bibitem[Shahriyar and Wright(2022)]%
        {Evaluating2022Shahriyar}
\bibfield{author}{\bibinfo{person}{Shaikh~Akib Shahriyar} {and} \bibinfo{person}{Matthew Wright}.} \bibinfo{year}{2022}\natexlab{}.
\newblock \showarticletitle{Evaluating Robustness of Sequence-Based Deepfake Detector Models by Adversarial Perturbation}. In \bibinfo{booktitle}{\emph{Proceedings of the 1st Workshop on Security Implications of Deepfakes and Cheapfakes}} \emph{(\bibinfo{series}{WDC '22})}. \bibinfo{publisher}{Association for Computing Machinery}, \bibinfo{address}{New York, NY, USA}, \bibinfo{pages}{13–18}.
\newblock
\showISBNx{9781450391788}
\urldef\tempurl%
\url{https://doi.org/10.1145/3494109.3527194}
\showDOI{\tempurl}


\bibitem[Shang et~al\mbox{.}(2021)]%
        {PRRNet2021SHANG}
\bibfield{author}{\bibinfo{person}{Zhihua Shang}, \bibinfo{person}{Hongtao Xie}, \bibinfo{person}{Zhengjun Zha}, \bibinfo{person}{Lingyun Yu}, \bibinfo{person}{Yan Li}, {and} \bibinfo{person}{Yongdong Zhang}.} \bibinfo{year}{2021}\natexlab{}.
\newblock \showarticletitle{PRRNet: Pixel-Region relation network for face forgery detection}.
\newblock \bibinfo{journal}{\emph{Pattern Recognition}}  \bibinfo{volume}{116} (\bibinfo{year}{2021}), \bibinfo{pages}{107950}.
\newblock
\showISSN{0031-3203}
\urldef\tempurl%
\url{https://doi.org/10.1016/j.patcog.2021.107950}
\showDOI{\tempurl}


\bibitem[Shiohara and Yamasaki(2022)]%
        {SBIs2022Shiohara}
\bibfield{author}{\bibinfo{person}{Kaede Shiohara} {and} \bibinfo{person}{Toshihiko Yamasaki}.} \bibinfo{year}{2022}\natexlab{}.
\newblock \showarticletitle{Detecting Deepfakes with Self-Blended Images}. In \bibinfo{booktitle}{\emph{Proceedings of the IEEE/CVF Conference on Computer Vision and Pattern Recognition}}. \bibinfo{pages}{18720--18729}.
\newblock


\bibitem[Shvets(2022)]%
        {Shvets2022Reface}
\bibfield{author}{\bibinfo{person}{Dima Shvets}.} \bibinfo{year}{2022}\natexlab{}.
\newblock \bibinfo{title}{Reface}.
\newblock \bibinfo{howpublished}{\url{https://hey.reface.ai/}}.
\newblock
\newblock
\shownote{Accessed: 2022-09-08}.


\bibitem[Simonyan and Zisserman(2015)]%
        {Simonyan2015very}
\bibfield{author}{\bibinfo{person}{Karen Simonyan} {and} \bibinfo{person}{Andrew Zisserman}.} \bibinfo{year}{2015}\natexlab{}.
\newblock \showarticletitle{Very Deep Convolutional Networks for Large-Scale Image Recognition}. In \bibinfo{booktitle}{\emph{3rd International Conference on Learning Representations}}, \bibfield{editor}{\bibinfo{person}{Yoshua Bengio} {and} \bibinfo{person}{Yann LeCun}} (Eds.).
\newblock


\bibitem[Sistemas(2004)]%
        {virustotal}
\bibfield{author}{\bibinfo{person}{Hispasec Sistemas}.} \bibinfo{year}{2004}\natexlab{}.
\newblock \bibinfo{title}{VirusTotal}.
\newblock \bibinfo{howpublished}{\url{https://www.virustotal.com}}.
\newblock
\newblock
\shownote{Accessed: 2023-02-28}.


\bibitem[Sun et~al\mbox{.}(2022)]%
        {Sun2022Dual}
\bibfield{author}{\bibinfo{person}{Ke Sun}, \bibinfo{person}{Taiping Yao}, \bibinfo{person}{Shen Chen}, \bibinfo{person}{Shouhong Ding}, \bibinfo{person}{Jilin Li}, {and} \bibinfo{person}{Rongrong Ji}.} \bibinfo{year}{2022}\natexlab{}.
\newblock \showarticletitle{Dual Contrastive Learning for General Face Forgery Detection}.
\newblock \bibinfo{journal}{\emph{Proceedings of the AAAI Conference on Artificial Intelligence}} \bibinfo{volume}{36}, \bibinfo{number}{2} (\bibinfo{date}{Jun.} \bibinfo{year}{2022}), \bibinfo{pages}{2316--2324}.
\newblock
\urldef\tempurl%
\url{https://doi.org/10.1609/aaai.v36i2.20130}
\showDOI{\tempurl}


\bibitem[Sun et~al\mbox{.}(2021)]%
        {sun2021improving}
\bibfield{author}{\bibinfo{person}{Zekun Sun}, \bibinfo{person}{Yujie Han}, \bibinfo{person}{Zeyu Hua}, \bibinfo{person}{Na Ruan}, {and} \bibinfo{person}{Weijia Jia}.} \bibinfo{year}{2021}\natexlab{}.
\newblock \showarticletitle{Improving the Efficiency and Robustness of Deepfakes Detection through Precise Geometric Features}. In \bibinfo{booktitle}{\emph{Proceedings of the IEEE/CVF Conference on Computer Vision and Pattern Recognition (CVPR)}}. \bibinfo{pages}{3609--3618}.
\newblock


\bibitem[Tan and Le(2019)]%
        {Tan2019EfficientNet}
\bibfield{author}{\bibinfo{person}{Mingxing Tan} {and} \bibinfo{person}{Quoc Le}.} \bibinfo{year}{2019}\natexlab{}.
\newblock \showarticletitle{{E}fficient{N}et: Rethinking Model Scaling for Convolutional Neural Networks}. In \bibinfo{booktitle}{\emph{Proceedings of the 36th International Conference on Machine Learning}} \emph{(\bibinfo{series}{Proceedings of Machine Learning Research}, Vol.~\bibinfo{volume}{97})}. \bibinfo{publisher}{PMLR}, \bibinfo{pages}{6105--6114}.
\newblock


\bibitem[Tariq et~al\mbox{.}(2018)]%
        {tariq2018detecting}
\bibfield{author}{\bibinfo{person}{Shahroz Tariq}, \bibinfo{person}{Sangyup Lee}, \bibinfo{person}{Hoyoung Kim}, \bibinfo{person}{Youjin Shin}, {and} \bibinfo{person}{Simon~S. Woo}.} \bibinfo{year}{2018}\natexlab{}.
\newblock \showarticletitle{Detecting Both Machine and Human Created Fake Face Images In the Wild}. In \bibinfo{booktitle}{\emph{Proceedings of the 2nd International Workshop on Multimedia Privacy and Security}} (Toronto, Canada) \emph{(\bibinfo{series}{MPS '18})}. \bibinfo{publisher}{Association for Computing Machinery}, \bibinfo{address}{New York, NY, USA}, \bibinfo{pages}{81–87}.
\newblock
\showISBNx{9781450359887}
\urldef\tempurl%
\url{https://doi.org/10.1145/3267357.3267367}
\showDOI{\tempurl}


\bibitem[Thies et~al\mbox{.}(2019)]%
        {Thies2019Deferred}
\bibfield{author}{\bibinfo{person}{Justus Thies}, \bibinfo{person}{Michael Zollh\"{o}fer}, {and} \bibinfo{person}{Matthias Nie\ss{}ner}.} \bibinfo{year}{2019}\natexlab{}.
\newblock \showarticletitle{Deferred Neural Rendering: Image Synthesis Using Neural Textures}.
\newblock \bibinfo{journal}{\emph{ACM Trans. Graph.}} \bibinfo{volume}{38}, \bibinfo{number}{4}, Article \bibinfo{articleno}{66} (\bibinfo{date}{July} \bibinfo{year}{2019}), \bibinfo{numpages}{12}~pages.
\newblock
\showISSN{0730-0301}


\bibitem[Thies et~al\mbox{.}(2016)]%
        {Thies2016Face2Face}
\bibfield{author}{\bibinfo{person}{Justus Thies}, \bibinfo{person}{Michael Zollhofer}, \bibinfo{person}{Marc Stamminger}, \bibinfo{person}{Christian Theobalt}, {and} \bibinfo{person}{Matthias Niessner}.} \bibinfo{year}{2016}\natexlab{}.
\newblock \showarticletitle{Face2Face: Real-Time Face Capture and Reenactment of RGB Videos}. In \bibinfo{booktitle}{\emph{IEEE Conference on Computer Vision and Patten Recognition (CVPR)}}. \bibinfo{pages}{2387--2395}.
\newblock


\bibitem[Times(2021a)]%
        {zhangzhehan_scandal}
\bibfield{author}{\bibinfo{person}{Global Times}.} \bibinfo{year}{2021}\natexlab{a}.
\newblock \bibinfo{title}{Chinese social media platforms delete actor’s accounts of for hurting the nation after controversial photos of Yasukuni Shrine}.
\newblock \bibinfo{howpublished}{\url{https://www.globaltimes.cn/page/202108/1231473.shtml}}.
\newblock
\newblock
\shownote{Accessed: 2022-09-27}.


\bibitem[Times(2021b)]%
        {zhengshuang_scandal}
\bibfield{author}{\bibinfo{person}{Global Times}.} \bibinfo{year}{2021}\natexlab{b}.
\newblock \bibinfo{title}{Chinese surrogacy scandal actress Zheng Shuang fined \$46 million for tax evasion, shows banned}.
\newblock \bibinfo{howpublished}{\url{https://www.globaltimes.cn/page/202108/1232636.shtml}}.
\newblock
\newblock
\shownote{Accessed: 2022-09-27}.


\bibitem[Times(2021c)]%
        {zhaowei_scandal}
\bibfield{author}{\bibinfo{person}{Global Times}.} \bibinfo{year}{2021}\natexlab{c}.
\newblock \bibinfo{title}{Works of scandals-hit actress Zhao Wei removed from platforms, following ban on actor Zhang Zhehan for visiting Yasukuni Shrine}.
\newblock \bibinfo{howpublished}{\url{https://www.globaltimes.cn/page/202108/1232631.shtml}}.
\newblock
\newblock
\shownote{Accessed: 2022-09-27}.


\bibitem[Tolosana et~al\mbox{.}(2021)]%
        {Tolosana2021DeepFakes}
\bibfield{author}{\bibinfo{person}{Ruben Tolosana}, \bibinfo{person}{Sergio Romero-Tapiador}, \bibinfo{person}{Julian Fierrez}, {and} \bibinfo{person}{Ruben Vera-Rodriguez}.} \bibinfo{year}{2021}\natexlab{}.
\newblock \showarticletitle{DeepFakes Evolution: Analysis of Facial Regions and Fake Detection Performance}. In \bibinfo{booktitle}{\emph{Pattern Recognition. ICPR International Workshops and Challenges}}, \bibfield{editor}{\bibinfo{person}{Alberto Del~Bimbo}, \bibinfo{person}{Rita Cucchiara}, \bibinfo{person}{Stan Sclaroff}, \bibinfo{person}{Giovanni~Maria Farinella}, \bibinfo{person}{Tao Mei}, \bibinfo{person}{Marco Bertini}, \bibinfo{person}{Hugo~Jair Escalante}, {and} \bibinfo{person}{Roberto Vezzani}} (Eds.). \bibinfo{publisher}{Springer International Publishing}, \bibinfo{address}{Cham}, \bibinfo{pages}{442--456}.
\newblock
\showISBNx{978-3-030-68821-9}


\bibitem[Tolosana et~al\mbox{.}(2020)]%
        {TOLOSANA2020Deepfakes}
\bibfield{author}{\bibinfo{person}{Ruben Tolosana}, \bibinfo{person}{Ruben Vera-Rodriguez}, \bibinfo{person}{Julian Fierrez}, \bibinfo{person}{Aythami Morales}, {and} \bibinfo{person}{Javier Ortega-Garcia}.} \bibinfo{year}{2020}\natexlab{}.
\newblock \showarticletitle{Deepfakes and beyond: A Survey of face manipulation and fake detection}.
\newblock \bibinfo{journal}{\emph{Information Fusion}}  \bibinfo{volume}{64} (\bibinfo{year}{2020}), \bibinfo{pages}{131--148}.
\newblock
\showISSN{1566-2535}
\urldef\tempurl%
\url{https://doi.org/10.1016/j.inffus.2020.06.014}
\showDOI{\tempurl}


\bibitem[Tonucci(2005)]%
        {target_population}
\bibfield{author}{\bibinfo{person}{David Tonucci}.} \bibinfo{year}{2005}\natexlab{}.
\newblock \showarticletitle{44 - New and Emerging Testing Technology for Efficacy and Safety E valuation of Personal Care Delivery Systems}.
\newblock In \bibinfo{booktitle}{\emph{Delivery System Handbook for Personal Care and Cosmetic Products}}, \bibfield{editor}{\bibinfo{person}{Meyer~R. Rosen}} (Ed.). \bibinfo{publisher}{William Andrew Publishing}, \bibinfo{address}{Norwich, NY}, \bibinfo{pages}{911--929}.
\newblock
\showISBNx{978-0-8155-1504-3}
\urldef\tempurl%
\url{https://doi.org/10.1016/B978-081551504-3.50049-3}
\showDOI{\tempurl}


\bibitem[Touvron et~al\mbox{.}(2021)]%
        {Hugo2021Training}
\bibfield{author}{\bibinfo{person}{Hugo Touvron}, \bibinfo{person}{Matthieu Cord}, \bibinfo{person}{Matthijs Douze}, \bibinfo{person}{Francisco Massa}, \bibinfo{person}{Alexandre Sablayrolles}, {and} \bibinfo{person}{Herve Jegou}.} \bibinfo{year}{2021}\natexlab{}.
\newblock \showarticletitle{Training data-efficient image transformers \& distillation through attention}. In \bibinfo{booktitle}{\emph{Proceedings of the 38th International Conference on Machine Learning}} \emph{(\bibinfo{series}{Proceedings of Machine Learning Research}, Vol.~\bibinfo{volume}{139})}, \bibfield{editor}{\bibinfo{person}{Marina Meila} {and} \bibinfo{person}{Tong Zhang}} (Eds.). \bibinfo{publisher}{PMLR}, \bibinfo{pages}{10347--10357}.
\newblock


\bibitem[Trinh et~al\mbox{.}(2021)]%
        {Interpretable2021Trinh_WACV}
\bibfield{author}{\bibinfo{person}{Loc Trinh}, \bibinfo{person}{Michael Tsang}, \bibinfo{person}{Sirisha Rambhatla}, {and} \bibinfo{person}{Yan Liu}.} \bibinfo{year}{2021}\natexlab{}.
\newblock \showarticletitle{Interpretable and Trustworthy Deepfake Detection via Dynamic Prototypes}. In \bibinfo{booktitle}{\emph{Proceedings of the IEEE/CVF Winter Conference on Applications of Computer Vision (WACV)}}. \bibinfo{pages}{1973--1983}.
\newblock


\bibitem[Tripathy et~al\mbox{.}(2019)]%
        {ICface2019Tripathy}
\bibfield{author}{\bibinfo{person}{Soumya Tripathy}, \bibinfo{person}{Juho Kannala}, {and} \bibinfo{person}{Esa Rahtu}.} \bibinfo{year}{2019}\natexlab{}.
\newblock \showarticletitle{ICface: Interpretable and Controllable Face Reenactment Using GANs}.
\newblock \bibinfo{journal}{\emph{arXiv preprint arXiv:1904.01909}} (\bibinfo{year}{2019}).
\newblock


\bibitem[Vaswani et~al\mbox{.}(2017)]%
        {Vaswani2017Attention}
\bibfield{author}{\bibinfo{person}{Ashish Vaswani}, \bibinfo{person}{Noam Shazeer}, \bibinfo{person}{Niki Parmar}, \bibinfo{person}{Jakob Uszkoreit}, \bibinfo{person}{Llion Jones}, \bibinfo{person}{Aidan~N Gomez}, \bibinfo{person}{Lukasz Kaiser}, {and} \bibinfo{person}{Illia Polosukhin}.} \bibinfo{year}{2017}\natexlab{}.
\newblock \showarticletitle{Attention is All you Need}. In \bibinfo{booktitle}{\emph{Advances in Neural Information Processing Systems}}, \bibfield{editor}{\bibinfo{person}{I.~Guyon}, \bibinfo{person}{U.~V. Luxburg}, \bibinfo{person}{S.~Bengio}, \bibinfo{person}{H.~Wallach}, \bibinfo{person}{R.~Fergus}, \bibinfo{person}{S.~Vishwanathan}, {and} \bibinfo{person}{R.~Garnett}} (Eds.), Vol.~\bibinfo{volume}{30}. \bibinfo{publisher}{Curran Associates, Inc.}
\newblock


\bibitem[Wang et~al\mbox{.}(2020)]%
        {wang2020cnngen}
\bibfield{author}{\bibinfo{person}{Sheng-Yu Wang}, \bibinfo{person}{Oliver Wang}, \bibinfo{person}{Richard Zhang}, \bibinfo{person}{Andrew Owens}, {and} \bibinfo{person}{Alexei~A. Efros}.} \bibinfo{year}{2020}\natexlab{}.
\newblock \showarticletitle{CNN-Generated Images Are Surprisingly Easy to Spot… for Now}. In \bibinfo{booktitle}{\emph{2020 IEEE/CVF Conference on Computer Vision and Pattern Recognition (CVPR)}}. \bibinfo{pages}{8692--8701}.
\newblock
\urldef\tempurl%
\url{https://doi.org/10.1109/CVPR42600.2020.00872}
\showDOI{\tempurl}


\bibitem[Wang et~al\mbox{.}(2023)]%
        {DCPT2022Wang}
\bibfield{author}{\bibinfo{person}{Tianyi Wang}, \bibinfo{person}{Harry Cheng}, \bibinfo{person}{Kam~Pui Chow}, {and} \bibinfo{person}{Liqiang Nie}.} \bibinfo{year}{2023}\natexlab{}.
\newblock \showarticletitle{Deep Convolutional Pooling Transformer for Deepfake Detection}.
\newblock \bibinfo{journal}{\emph{ACM Transactions on Multimedia Computing, Communications, and Applications}} \bibinfo{volume}{19}, \bibinfo{number}{6}.
\newblock
\showISSN{1551-6857}


\bibitem[Wang and Chow(2023)]%
        {Wang2023NoiseDF_AAAI}
\bibfield{author}{\bibinfo{person}{Tianyi Wang} {and} \bibinfo{person}{Kam~Pui Chow}.} \bibinfo{year}{2023}\natexlab{}.
\newblock \showarticletitle{Noise Based Deepfake Detection via Multi-Head Relative-Interaction}.
\newblock \bibinfo{journal}{\emph{Proceedings of the AAAI Conference on Artificial Intelligence}} (\bibinfo{year}{2023}).
\newblock


\bibitem[Wang et~al\mbox{.}(2022a)]%
        {DeepfakeNoise2022Wang}
\bibfield{author}{\bibinfo{person}{Tianyi Wang}, \bibinfo{person}{Ming Liu}, \bibinfo{person}{Wei Cao}, {and} \bibinfo{person}{Kam~Pui Chow}.} \bibinfo{year}{2022}\natexlab{a}.
\newblock \showarticletitle{Deepfake noise investigation and detection}.
\newblock \bibinfo{journal}{\emph{Forensic Science International: Digital Investigation}}  \bibinfo{volume}{42} (\bibinfo{year}{2022}), \bibinfo{pages}{301395}.
\newblock
\showISSN{2666-2817}
\urldef\tempurl%
\url{https://doi.org/10.1016/j.fsidi.2022.301395}
\showDOI{\tempurl}
\newblock
\shownote{Proceedings of the Twenty-Second Annual DFRWS USA}.


\bibitem[Wang et~al\mbox{.}(2021b)]%
        {Wang2021Pyramid}
\bibfield{author}{\bibinfo{person}{Wenhai Wang}, \bibinfo{person}{Enze Xie}, \bibinfo{person}{Xiang Li}, \bibinfo{person}{Deng-Ping Fan}, \bibinfo{person}{Kaitao Song}, \bibinfo{person}{Ding Liang}, \bibinfo{person}{Tong Lu}, \bibinfo{person}{Ping Luo}, {and} \bibinfo{person}{Ling Shao}.} \bibinfo{year}{2021}\natexlab{b}.
\newblock \showarticletitle{Pyramid Vision Transformer: A Versatile Backbone for Dense Prediction Without Convolutions}. In \bibinfo{booktitle}{\emph{Proceedings of the IEEE/CVF International Conference on Computer Vision (ICCV)}}. \bibinfo{pages}{568--578}.
\newblock


\bibitem[Wang et~al\mbox{.}(2021a)]%
        {Wang2021HifiFace}
\bibfield{author}{\bibinfo{person}{Yuhan Wang}, \bibinfo{person}{Xu Chen}, \bibinfo{person}{Junwei Zhu}, \bibinfo{person}{Wenqing Chu}, \bibinfo{person}{Ying Tai}, \bibinfo{person}{Chengjie Wang}, \bibinfo{person}{Jilin Li}, \bibinfo{person}{Yongjian Wu}, \bibinfo{person}{Feiyue Huang}, {and} \bibinfo{person}{Rongrong Ji}.} \bibinfo{year}{2021}\natexlab{a}.
\newblock \showarticletitle{HifiFace: 3D Shape and Semantic Prior Guided High Fidelity Face Swapping}. In \bibinfo{booktitle}{\emph{Proceedings of the Thirtieth International Joint Conference on Artificial Intelligence, {IJCAI-21}}}, \bibfield{editor}{\bibinfo{person}{Zhi-Hua Zhou}} (Ed.). \bibinfo{publisher}{International Joint Conferences on Artificial Intelligence Organization}, \bibinfo{pages}{1136--1142}.
\newblock
\urldef\tempurl%
\url{https://doi.org/10.24963/ijcai.2021/157}
\showDOI{\tempurl}


\bibitem[Wang et~al\mbox{.}(2022b)]%
        {ForgeryNIR2022Wang_TIFS}
\bibfield{author}{\bibinfo{person}{Yukai Wang}, \bibinfo{person}{Chunlei Peng}, \bibinfo{person}{Decheng Liu}, \bibinfo{person}{Nannan Wang}, {and} \bibinfo{person}{Xinbo Gao}.} \bibinfo{year}{2022}\natexlab{b}.
\newblock \showarticletitle{ForgeryNIR: Deep Face Forgery and Detection in Near-Infrared Scenario}.
\newblock \bibinfo{journal}{\emph{IEEE Transactions on Information Forensics and Security}}  \bibinfo{volume}{17} (\bibinfo{year}{2022}), \bibinfo{pages}{500--515}.
\newblock
\urldef\tempurl%
\url{https://doi.org/10.1109/TIFS.2022.3146766}
\showDOI{\tempurl}


\bibitem[Westerlund(2019)]%
        {Westerlund2019Emergence}
\bibfield{author}{\bibinfo{person}{Mika Westerlund}.} \bibinfo{year}{2019}\natexlab{}.
\newblock \showarticletitle{The Emergence of Deepfake Technology: A Review}.
\newblock \bibinfo{journal}{\emph{Technology Innovation Management Review}}  \bibinfo{volume}{9} (\bibinfo{date}{11} \bibinfo{year}{2019}), \bibinfo{pages}{40--53}.
\newblock
\showISSN{1927-0321}
\urldef\tempurl%
\url{https://doi.org/10.22215/timreview/1282}
\showDOI{\tempurl}


\bibitem[Wodajo and Atnafu(2021)]%
        {Deressa2021Deepfake}
\bibfield{author}{\bibinfo{person}{Deressa Wodajo} {and} \bibinfo{person}{Solomon Atnafu}.} \bibinfo{year}{2021}\natexlab{}.
\newblock \bibinfo{title}{Deepfake Video Detection Using Convolutional Vision Transformer}.
\newblock
\newblock
\urldef\tempurl%
\url{https://arxiv.org/abs/2102.11126}
\showURL{%
\tempurl}


\bibitem[Woods et~al\mbox{.}(2019)]%
        {Woods2019Adversarial}
\bibfield{author}{\bibinfo{person}{Walt Woods}, \bibinfo{person}{Jack Chen}, {and} \bibinfo{person}{Christof Teuscher}.} \bibinfo{year}{2019}\natexlab{}.
\newblock \showarticletitle{Adversarial explanations for understanding image classification decisions and improved neural network robustness}.
\newblock \bibinfo{journal}{\emph{Nature Machine Intelligence}} \bibinfo{volume}{1}, \bibinfo{number}{11} (\bibinfo{date}{01 Nov} \bibinfo{year}{2019}), \bibinfo{pages}{508--516}.
\newblock
\showISSN{2522-5839}
\urldef\tempurl%
\url{https://doi.org/10.1038/s42256-019-0104-6}
\showDOI{\tempurl}


\bibitem[Wu et~al\mbox{.}(2022a)]%
        {Robust2022Wu_CVPR}
\bibfield{author}{\bibinfo{person}{Haiwei Wu}, \bibinfo{person}{Jiantao Zhou}, \bibinfo{person}{Jinyu Tian}, {and} \bibinfo{person}{Jun Liu}.} \bibinfo{year}{2022}\natexlab{a}.
\newblock \showarticletitle{Robust Image Forgery Detection over Online Social Network Shared Images}. In \bibinfo{booktitle}{\emph{2022 IEEE/CVF Conference on Computer Vision and Pattern Recognition (CVPR)}}. \bibinfo{pages}{13430--13439}.
\newblock
\urldef\tempurl%
\url{https://doi.org/10.1109/CVPR52688.2022.01308}
\showDOI{\tempurl}


\bibitem[Wu et~al\mbox{.}(2022b)]%
        {Robust2022Wu_TIFS}
\bibfield{author}{\bibinfo{person}{Haiwei Wu}, \bibinfo{person}{Jiantao Zhou}, \bibinfo{person}{Jinyu Tian}, \bibinfo{person}{Jun Liu}, {and} \bibinfo{person}{Yu Qiao}.} \bibinfo{year}{2022}\natexlab{b}.
\newblock \showarticletitle{Robust Image Forgery Detection Against Transmission Over Online Social Networks}.
\newblock \bibinfo{journal}{\emph{IEEE Transactions on Information Forensics and Security}}  \bibinfo{volume}{17} (\bibinfo{year}{2022}), \bibinfo{pages}{443--456}.
\newblock
\urldef\tempurl%
\url{https://doi.org/10.1109/TIFS.2022.3144878}
\showDOI{\tempurl}


\bibitem[Wu et~al\mbox{.}(2018)]%
        {wayne2018reenactgan}
\bibfield{author}{\bibinfo{person}{Wayne Wu}, \bibinfo{person}{Yunxuan Zhang}, \bibinfo{person}{Cheng Li}, \bibinfo{person}{Chen Qian}, {and} \bibinfo{person}{Chen~Change Loy}.} \bibinfo{year}{2018}\natexlab{}.
\newblock \showarticletitle{ReenactGAN: Learning to Reenact Faces via Boundary Transfer}. In \bibinfo{booktitle}{\emph{ECCV}}.
\newblock


\bibitem[Wu et~al\mbox{.}(2020)]%
        {Wu2020SSTNet}
\bibfield{author}{\bibinfo{person}{Xi Wu}, \bibinfo{person}{Zhen Xie}, \bibinfo{person}{YuTao Gao}, {and} \bibinfo{person}{Yu Xiao}.} \bibinfo{year}{2020}\natexlab{}.
\newblock \showarticletitle{SSTNet: Detecting Manipulated Faces Through Spatial, Steganalysis and Temporal Features}. In \bibinfo{booktitle}{\emph{ICASSP 2020 - 2020 IEEE International Conference on Acoustics, Speech and Signal Processing (ICASSP)}}. \bibinfo{pages}{2952--2956}.
\newblock
\urldef\tempurl%
\url{https://doi.org/10.1109/ICASSP40776.2020.9053969}
\showDOI{\tempurl}


\bibitem[Xu et~al\mbox{.}(2022)]%
        {Supervised2022Xu_WACV}
\bibfield{author}{\bibinfo{person}{Ying Xu}, \bibinfo{person}{Kiran Raja}, {and} \bibinfo{person}{Marius Pedersen}.} \bibinfo{year}{2022}\natexlab{}.
\newblock \showarticletitle{Supervised Contrastive Learning for Generalizable and Explainable DeepFakes Detection}. In \bibinfo{booktitle}{\emph{Proceedings of the IEEE/CVF Winter Conference on Applications of Computer Vision (WACV) Workshops}}. \bibinfo{pages}{379--389}.
\newblock


\bibitem[Yang et~al\mbox{.}(2019)]%
        {ExposingDeep2019Li}
\bibfield{author}{\bibinfo{person}{Xin Yang}, \bibinfo{person}{Yuezun Li}, {and} \bibinfo{person}{Siwei Lyu}.} \bibinfo{year}{2019}\natexlab{}.
\newblock \showarticletitle{Exposing Deep Fakes Using Inconsistent Head Poses}. In \bibinfo{booktitle}{\emph{ICASSP 2019 - 2019 IEEE International Conference on Acoustics, Speech and Signal Processing (ICASSP)}}. \bibinfo{pages}{8261--8265}.
\newblock
\urldef\tempurl%
\url{https://doi.org/10.1109/ICASSP.2019.8683164}
\showDOI{\tempurl}


\bibitem[Zakharov et~al\mbox{.}(2019)]%
        {nth2019zakharov}
\bibfield{author}{\bibinfo{person}{Egor Zakharov}, \bibinfo{person}{Aliaksandra Shysheya}, \bibinfo{person}{Egor Burkov}, {and} \bibinfo{person}{Victor Lempitsky}.} \bibinfo{year}{2019}\natexlab{}.
\newblock \showarticletitle{Few-Shot Adversarial Learning of Realistic Neural Talking Head Models}. In \bibinfo{booktitle}{\emph{2019 IEEE/CVF International Conference on Computer Vision (ICCV)}}. \bibinfo{pages}{9458--9467}.
\newblock
\urldef\tempurl%
\url{https://doi.org/10.1109/ICCV.2019.00955}
\showDOI{\tempurl}


\bibitem[Zhang et~al\mbox{.}(2021)]%
        {zhang2021facial}
\bibfield{author}{\bibinfo{person}{Chenxu Zhang}, \bibinfo{person}{Yifan Zhao}, \bibinfo{person}{Yifei Huang}, \bibinfo{person}{Ming Zeng}, \bibinfo{person}{Saifeng Ni}, \bibinfo{person}{Madhukar Budagavi}, {and} \bibinfo{person}{Xiaohu Guo}.} \bibinfo{year}{2021}\natexlab{}.
\newblock \showarticletitle{FACIAL: Synthesizing Dynamic Talking Face with Implicit Attribute Learning}. In \bibinfo{booktitle}{\emph{Proceedings of the IEEE/CVF International Conference on Computer Vision (ICCV)}}. \bibinfo{pages}{3867--3876}.
\newblock


\bibitem[Zhang et~al\mbox{.}(2017)]%
        {zhang2017beyond}
\bibfield{author}{\bibinfo{person}{Kai Zhang}, \bibinfo{person}{Wangmeng Zuo}, \bibinfo{person}{Yunjin Chen}, \bibinfo{person}{Deyu Meng}, {and} \bibinfo{person}{Lei Zhang}.} \bibinfo{year}{2017}\natexlab{}.
\newblock \showarticletitle{Beyond a {Gaussian} denoiser: Residual learning of deep {CNN} for image denoising}.
\newblock \bibinfo{journal}{\emph{IEEE Transactions on Image Processing}} \bibinfo{volume}{26}, \bibinfo{number}{7} (\bibinfo{year}{2017}), \bibinfo{pages}{3142--3155}.
\newblock


\bibitem[Zhang et~al\mbox{.}(2018)]%
        {blur2018zhang}
\bibfield{author}{\bibinfo{person}{Shanghang Zhang}, \bibinfo{person}{Xiaohui Shen}, \bibinfo{person}{Zhe Lin}, \bibinfo{person}{Radomír Mech}, \bibinfo{person}{João~P. Costeira}, {and} \bibinfo{person}{Jose~M.F. Moura}.} \bibinfo{year}{2018}\natexlab{}.
\newblock \showarticletitle{Learning to Understand Image Blur}. In \bibinfo{booktitle}{\emph{2018 IEEE/CVF Conference on Computer Vision and Pattern Recognition}}. \bibinfo{pages}{6586--6595}.
\newblock
\urldef\tempurl%
\url{https://doi.org/10.1109/CVPR.2018.00689}
\showDOI{\tempurl}


\bibitem[Zhang et~al\mbox{.}(2019)]%
        {OneShotFace2019}
\bibfield{author}{\bibinfo{person}{Yunxuan Zhang}, \bibinfo{person}{Siwei Zhang}, \bibinfo{person}{Yue He}, \bibinfo{person}{Cheng Li}, \bibinfo{person}{Chen~Change Loy}, {and} \bibinfo{person}{Ziwei Liu}.} \bibinfo{year}{2019}\natexlab{}.
\newblock \showarticletitle{One-shot Face Reenactment}. In \bibinfo{booktitle}{\emph{British Machine Vision Conference (BMVC)}}.
\newblock


\bibitem[Zhao et~al\mbox{.}(2021b)]%
        {Zhao2021Multi-Attentional}
\bibfield{author}{\bibinfo{person}{Hanqing Zhao}, \bibinfo{person}{Wenbo Zhou}, \bibinfo{person}{Dongdong Chen}, \bibinfo{person}{Tianyi Wei}, \bibinfo{person}{Weiming Zhang}, {and} \bibinfo{person}{Nenghai Yu}.} \bibinfo{year}{2021}\natexlab{b}.
\newblock \showarticletitle{Multi-Attentional Deepfake Detection}. In \bibinfo{booktitle}{\emph{IEEE Conference on Computer Vision and Patten Recognition (CVPR)}}. \bibinfo{pages}{2185--2194}.
\newblock


\bibitem[Zhao et~al\mbox{.}(2021a)]%
        {SelfConsis2021Zhao}
\bibfield{author}{\bibinfo{person}{T. Zhao}, \bibinfo{person}{X. Xu}, \bibinfo{person}{M. Xu}, \bibinfo{person}{H. Ding}, \bibinfo{person}{Y. Xiong}, {and} \bibinfo{person}{W. Xia}.} \bibinfo{year}{2021}\natexlab{a}.
\newblock \showarticletitle{Learning Self-Consistency for Deepfake Detection}. In \bibinfo{booktitle}{\emph{2021 IEEE/CVF International Conference on Computer Vision (ICCV)}}. \bibinfo{publisher}{IEEE Computer Society}, \bibinfo{address}{Los Alamitos, CA, USA}, \bibinfo{pages}{15003--15013}.
\newblock
\urldef\tempurl%
\url{https://doi.org/10.1109/ICCV48922.2021.01475}
\showDOI{\tempurl}


\bibitem[Zhou et~al\mbox{.}(2017)]%
        {zhou2017twostream}
\bibfield{author}{\bibinfo{person}{Peng Zhou}, \bibinfo{person}{Xintong Han}, \bibinfo{person}{Vlad~I. Morariu}, {and} \bibinfo{person}{Larry~S. Davis}.} \bibinfo{year}{2017}\natexlab{}.
\newblock \showarticletitle{Two-Stream Neural Networks for Tampered Face Detection}. In \bibinfo{booktitle}{\emph{2017 IEEE Conference on Computer Vision and Pattern Recognition Workshops (CVPRW)}}. \bibinfo{pages}{1831--1839}.
\newblock
\urldef\tempurl%
\url{https://doi.org/10.1109/CVPRW.2017.229}
\showDOI{\tempurl}


\bibitem[Zhu et~al\mbox{.}(2017)]%
        {cyclegan2017zhu}
\bibfield{author}{\bibinfo{person}{Jun-Yan Zhu}, \bibinfo{person}{Taesung Park}, \bibinfo{person}{Phillip Isola}, {and} \bibinfo{person}{Alexei~A. Efros}.} \bibinfo{year}{2017}\natexlab{}.
\newblock \showarticletitle{Unpaired Image-to-Image Translation Using Cycle-Consistent Adversarial Networks}. In \bibinfo{booktitle}{\emph{2017 IEEE International Conference on Computer Vision (ICCV)}}. \bibinfo{pages}{2242--2251}.
\newblock
\urldef\tempurl%
\url{https://doi.org/10.1109/ICCV.2017.244}
\showDOI{\tempurl}


\bibitem[Zhu et~al\mbox{.}(2021)]%
        {zhu2021megafs}
\bibfield{author}{\bibinfo{person}{Yuhao Zhu}, \bibinfo{person}{Qi Li}, \bibinfo{person}{Jian Wang}, \bibinfo{person}{Chengzhong Xu}, {and} \bibinfo{person}{Zhenan Sun}.} \bibinfo{year}{2021}\natexlab{}.
\newblock \showarticletitle{One Shot Face Swapping on Megapixels}. In \bibinfo{booktitle}{\emph{Proceedings of the IEEE conference on computer vision and pattern recognition (CVPR)}}. \bibinfo{pages}{4834--4844}.
\newblock


\bibitem[Zi et~al\mbox{.}(2020)]%
        {WildDeepfake2020}
\bibfield{author}{\bibinfo{person}{Bojia Zi}, \bibinfo{person}{Minghao Chang}, \bibinfo{person}{Jingjing Chen}, \bibinfo{person}{Xingjun Ma}, {and} \bibinfo{person}{Yu-Gang Jiang}.} \bibinfo{year}{2020}\natexlab{}.
\newblock \showarticletitle{WildDeepfake: A Challenging Real-World Dataset for Deepfake Detection}. \bibinfo{pages}{2382--2390}.
\newblock
\urldef\tempurl%
\url{https://doi.org/10.1145/3394171.3413769}
\showDOI{\tempurl}


\bibitem[Zoph et~al\mbox{.}(2018)]%
        {nas_net}
\bibfield{author}{\bibinfo{person}{Barret Zoph}, \bibinfo{person}{Vijay Vasudevan}, \bibinfo{person}{Jonathon Shlens}, {and} \bibinfo{person}{Quoc~V. Le}.} \bibinfo{year}{2018}\natexlab{}.
\newblock \showarticletitle{Learning Transferable Architectures for Scalable Image Recognition}. In \bibinfo{booktitle}{\emph{2018 IEEE/CVF Conference on Computer Vision and Pattern Recognition}}. \bibinfo{pages}{8697--8710}.
\newblock
\urldef\tempurl%
\url{https://doi.org/10.1109/CVPR.2018.00907}
\showDOI{\tempurl}


\end{thebibliography}

%\appendix

\end{document}